\documentclass[tcc,alf]{tcc-poli}  

\usepackage[utf8]{inputenc}
\usepackage{float} 
\usepackage{appendix}
\usepackage{pdflscape}
\usepackage{multirow}
\usepackage{longtable}
\usepackage{graphics}
\usepackage{times}
\usepackage{enumerate}
\usepackage[hidelinks]{hyperref}
\usepackage{epstopdf}
\usepackage{lipsum}
\usepackage{amssymb}
\usepackage{fancyvrb} 
\usepackage{etoolbox}
\usepackage{siunitx}
\usepackage{listings}
\usepackage{caption}
\usepackage{subcaption}
\usepackage[normalem]{ulem}
\useunder{\uline}{\ul}{}
\usepackage{colortbl}
\usepackage{xcolor} 
\usepackage{pgfplots}
\pgfplotsset{compat=1.17}
\usepackage{pgfplotstable}
\usepackage{caption}
\usepackage{subcaption}
\usepackage{algorithm}
\usepackage{algpseudocode}
\usepackage{svg}

\usepackage{graphicx} 
\graphicspath{ {./Figuras/} } 

\usepackage[big]{titlesec}
\titleformat{\chapter}
{\normalfont\LARGE\bfseries}{\thechapter}{1em}{}
\titleformat{\section}
{\normalfont\Large\bfseries}{\thesection}{1em}{}
\titleformat{\subsection}
{\normalfont\large\bfseries}{\thesubsection}{1em}{}
\titleformat{\subsubsection}
{\normalfont\normalsize\bfseries}{\thesubsubsection}{1em}{}
\titleformat{\paragraph}[runin]
{\normalfont\normalsize\bfseries}{\theparagraph}{1em}{}
\titleformat{\subparagraph}[runin]
{\normalfont\normalsize\bfseries}{\thesubparagraph}{1em}{}

\usepackage[intoc]{nomencl}
\makenomenclature

 \usepackage[nomain,acronym,xindy,toc,noredefwarn]{glossaries}
 \makeglossaries

\ifx\AbntTexType\StringNum
  \usepackage[num]{abntex2cite}
  \citebrackets[] 
\else
  \usepackage[alf]{abntex2cite}
\fi

\usepackage{svg}

\usepackage{graphicx} 
\graphicspath{ {./Figuras/} }

\usepackage[english]{babel}
\addto\captionsenglish{%
  
  \renewcommand{\listfigurename}{List of Figures}
  \renewcommand{\listtablename}{List of Tables}

  \renewcommand{\bibname}{References}
}
\AtBeginDocument{\selectlanguage{english}}

\begin{document}

    \title{Signature Forgery Detection: Improving Cross-Dataset Generalization}
  
  \foreigntitle{Signature Forgery Detection: Improving Cross-Dataset Generalization - POLI/UFRJ}
   \author{Matheus} {Ramos Parracho}
  \advisor{Prof.}{Eduardo}{A.B. da Silva}{Ph.D} %
  \advisor{Prof.}{Weiler}{A. Finamore}{Ph.D}  
  
  \examiner{Prof.}{Heraldo Luis Silveira de Almeida}{D.Sc.}
  \examiner{Prof.}{Sergio Lima Netto}{Ph.D.}

  \engcourse{ELT} 
  
  \date{\the\month{}}{\the\year{}}


  \maketitle
  \frontmatter
  
   \mbox{}\vfill
\begin{flushright}
\begin{minipage}{0.5\textwidth}
\textit{Dedico esta dissertação à minha mãe, Gisele, e ao meu pai, Leandro, que sempre me apoiaram e me deram todo o suporte para que eu pudesse concluir este trabalho. Amo vocês!}\par
\end{minipage}
\end{flushright}
  \begin{abstract}
\noindent
Este trabalho explora o uso de técnicas de aprendizado de características (Feature Learning) para a detecção de falsificação de assinaturas em múltiplos conjuntos de dados, tendo como objetivo principal avaliar e aprimorar a capacidade de generalização \textit{cross-dataset} do modelo, isto é, sua eficácia ao ser treinado em um conjunto e testado em outro distinto. O estudo avalia o desempenho dos modelos utilizando os conjuntos de dados CEDAR, ICDAR e GPDS Synthetic, realizando testes entre bases para avaliar a robustez. Foram propostos dois benchmarks experimentais. O primeiro utiliza as imagens brutas das assinaturas como entrada, enquanto o segundo aplica um pré-processamento denominado \textit{pré-processamento de cascas}. Diversos padrões de comportamento dos modelos puderam ser identificados e analisados, entretanto, não foi possível estabelecer uma conclusão definitiva sobre a superioridade de um benchmark em relação ao outro. Os resultados indicam que o modelo simplificado (com pré-processamento) apresenta potencial promissor, mas ainda não consegue superar o modelo baseado em imagens completas, apontando para a necessidade de refinamentos adicionais antes que possa igualar ou exceder seu desempenho.

\end{abstract}

\textit{Palavras-chave: }{Redes Neurais, Aprendizagem de Características, Redes Neurais Convolucionais, Avaliação entre Conjuntos de Dados, Detecção de Falsificação de Assinatura}

  \begin{foreignabstract}
This work explores the use of feature learning techniques for signature forgery detection across multiple datasets, with the main objective of evaluating and enhancing the model’s cross-dataset generalization capacity—that is, its effectiveness when trained on one dataset and tested on a distinct one. The study evaluates model performance using the CEDAR, ICDAR, and GPDS Synthetic datasets, conducting cross-benchmark tests to assess robustness. Two experimental benchmarks were proposed. The first employs raw signature images as input, while the second applies a preprocessing technique called shell preprocessing. Several patterns in model behavior were identified and analyzed; however, it was not possible to establish a definitive conclusion regarding the superiority of one benchmark over the other. The results indicate that the simplified model (with preprocessing) shows promising potential but has not yet surpassed the image-based model, highlighting the need for further refinements before it can match or exceed its performance.
\end{foreignabstract}

\vspace*{0.5cm}
 \par
  \tableofcontents 
  \listoffigures
  \listoftables
  \cleardoublepage
  \printglossary[title=Lista de Siglas e Abreviaturas, type=\acronymtype]
  \mainmatter
  
  \chapter{Introduction}

Automated signature verification is a critical technique in biometrics, with applications in banking, identity authentication, and legal documentation. However, given the variety of handwriting styles and the sophistication of modern forgery techniques, it is difficult to distinguish authentic and fakes signatures.

There are many studies aiming to understand and develop methods to automatically classify signatures, however, a significant limitation observed in most works on offline signature verification is their inability to generalize across datasets. As also shown in Signet \cite{signet}, different datasets show different signature characteristics, making cross-dataset generalization challenging. Papers like "Handwritten signature forgery detection using Convolutional Neural Network" by IAAC India \cite{iaac} and a similar approach by Doe and Smith \cite{ieee_paper} show strong performance within a single dataset, but do not show the same results when tested on signatures from other datasets.

The aim of this study is to develop models capable of handling signatures from diverse datasets with similar performances, therefore enhancing their robustness to variations in handwriting styles and forgery techniques. The approach uses feature learning techniques over neural networks with Siamese architecture and customized layers trained under two feature losses. By analyzing cross-dataset performance using benchmark datasets such as CEDAR, ICDAR, and GPDS Synthetic, we get a comprehensive evaluation of the effectiveness of the model.

This work examines two distinct models: one which process the raw signature data (gray level images of the signatures) and the other which feeds the verification machine with pre-processed data. It is thus divided into two parts, namely,
\begin{enumerate}
    \item \textbf{Enhancing Model Generalization:} Investigating and developing techniques of signature verification models having as input the the raw image data. Satisfactory and stable performance  when moving across different datasets with diverse handwriting styles are sought. In other words, the aim is to find techniques with acceptable performance having stable behavior (not disparaging performance) when moving datasets of diverse natures.
    \item \textbf{Designing a Shell-Based Pre-processing Pipeline:} On the other hand  techniques of signature verification models having as input the pre-processed raw image data are developed and investigated. The pre-processing strategy is based on hull extraction ("shells") and the capture of relevant features to allow improved and less complex model.
\end{enumerate}


\noindent
\textbf{PART ONE - Enhancing Model Generalization.}

In the first part of this work, it is investigated the ability of deep learning models to adapt to diverse signature datasets. Evaluation considers signature (raw image data) taken from cross-dataset. The central goal is to understand how well a model trained on a particular dataset (e.g., CEDAR) performs when tested on distinct datasets (e.g., ICDAR or GPDS Synthetic), which differ in writing style, acquisition conditions, and resolution.

A detailed comparison between two loss functions (contrastive loss and triplet loss) used in the training is presented. The discussion focus on how  distinct sample structures (pairs vs. triplets) and thhe use of the different losses affects the training dynamics and the performance. The findings serve as a foundation for understanding the challenges in offline signature verification and how proper feature learning can mitigate their disturbance.

\noindent
\textbf{PART TWO - Designing a Shell-Based Pre-processing Pipeline.}

In the second part of this work, it is presented the development of a pre-processing pipeline based on hulls extraction, which will be referred to as "shells". These shells will be used as the data input to the neural network.

The core idea is to transform the 2-D signature image into a set of 1-D functions, enabling a dimensionality reduction while preserving the structural essence of the signature.

The main target  is to analyze if the whole scheme which uses pre-processing (with smaller complexity) is a better\footnote{Provides better or equal performance.} choice as compared to the scheme which take as input data the raw image.

\textbf{All code is open-source and can be found at: \url{https://github.com/mathparracho/general_signature_forgery_detector}}

\section{Objective}
\subsection{General Objective}

Develop and evaluate feature learning techniques for the task of signature matching, focusing on the detection of forged signatures. This includes using neural network architectures, such as Siamese Networks, and exploring loss functions like contrastive and triplet losses to enhance the accuracy of signature verification systems. The goal is to create a robust model that can effectively generalize to signatures from different datasets, addressing the generalization gap present in many existing approaches.

\subsection{Specific Objectives}

Within this general objective, there are specific goals:
\begin{itemize}
    \item Design a feature extraction and comparison model using the \textit{Siamese Neural Network} paradigm;

    \item Train the model with different samples from different datasets to assure diversity;
  
    \item Compare the performance of different loss functions: Contrastive and Triplet Loss;
   
    \item Analyze the impact of forged and genuine signature distributions on model accuracy through statistical and visual analysis (e.g., ROC curves and confusion matrices);
   
    \item Validate and analyze the robustness of the proposed model using cross-dataset testing;

    \item Develop and implement the pre-processing shell algorithm and compare its performance;
\end{itemize}

\section{Methodology}

The following methodology was designed:

\subsection{Dataset Preparation}
\begin{itemize}
    \item Use multiple signature datasets, including CEDAR, ICDAR, and GPDS Synthetic, to ensure a diverse range of signature styles.
    \item Pre-process the datasets to remove noise and standardize signature dimensions.
    \item Stablish a balanced training/validation/test to avoid bias.
\end{itemize}

\subsection{Model Development and Training}
\begin{itemize}
    \item Develop a Siamese Neural Network with a ResNet-34 backbone to compare pairs of signature images.
    \item Train the model using contrastive triplet losses, enabling the network to learn embedding distances that reflect signature similarity or dissimilarity.
\end{itemize}

\subsection{Cross-Dataset Evaluation}
\begin{itemize}
    \item Cross-dataset testing to evaluate the model's ability to generalize across different datasets. This involves training the model on two datasets (e.g., CEDAR and ICDAR) and testing it on another (e.g., GPDS).
    \item Monitor the Area Under the ROC Curve (AUC) as a key metric for performance evaluation.
\end{itemize}

\subsection{Analysis and Visualization}
\begin{itemize}
    \item Analyze the results using performance metrics such as ROC curves, confusion matrices, and AUC scores.
    \item Visualize the model performance on cross-dataset tests and report findings that show the effectiveness of the approach.
\end{itemize}

\subsection{Develop and implement the Shell Algorithm}
\begin{itemize}
    \item Design the algorithm, train, and compare the results.
\end{itemize}

\newpage

\section{Structure of the Work}

In Chapter 2, a literature review is presented. This chapter introduces the fundamental concepts of neural networks, starting from the Perceptron and progressing through Multilayer Perceptrons (MLPs), Convolutional Neural Networks (CNNs), and feature learning methods. It then discusses Siamese Neural Networks and the use of contrastive and triplet loss functions, along with evaluation metrics for signature recognition.

In Chapter 3, the development of the first part of the work is detailed. This includes a description of the datasets, the analysis and preparation of the data, the design of the model architecture, and the implementation of training strategies using contrastive and triplet losses. Results of this stage are also presented and discussed, with an emphasis on cross-dataset generalization.

In Chapter 4, the second part of the work is introduced. This section focuses on the design of a shell-based pre-processing pipeline, explaining the stages of image pre-processing, shell extraction, auxiliary feature estimation (such as pressure and thickness), and data export. A pseudo-algorithm and functional representations are also provided.Also, the experimental results to the methodology described are discussed. This includes model training and evaluation using shell-based signature representations, as well as a comparative analysis with results obtained from the original 2D signature images.

Finally, Chapter 5 presents the conclusion and outlines future work. It summarizes the main findings, highlights the contributions of this study, and suggests directions for further research in offline signature verification using advanced feature learning approaches.

  \chapter{The Theory Behind Signature Recognition}
    
 \section{What is a Neural Network?}

Artificial Neural Networks (ANNs) are computational methods inspired by the structure of the human brain as shown in picture \ref{fig:neuralNode}. They consist of interconnected nodes (neurons) organized in layers, where each connection is associated with a weight. 

\begin{figure}[H]
	\centering
	\includegraphics[width=11cm]{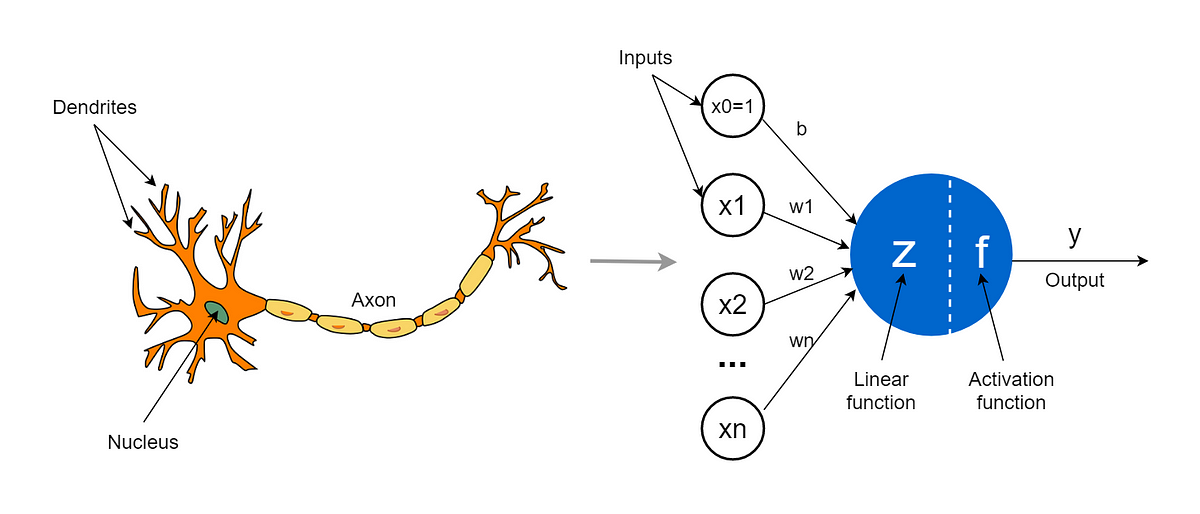}
	\caption{Similarity between neurons in the brain and neurons on neural networks}
	Fonte: Adapted from \cite{towardsdatascience_neural_networks}
    \label{fig:neuralNode}
\end{figure}

These systems are trained to model complex relationships between inputs and outputs, enabling them to perform tasks such as classification, regression, and data generation.

A single artificial neuron can be mathematically represented as:

\begin{equation} y = f\left( \sum_{i=1}^{n} w_i x_i + b \right), \end{equation}

where \( x_i \in \mathbb{R} \) represents the input features, \( w_i \in \mathbb{R} \) are the weights, and \( b \in \mathbb{R} \) is the bias term. The activation function \( f(\cdot) \) introduces non-linearity and can take different forms such as sigmoid, ReLU, or tanh.

\subsection{The Perceptron: The first and simplest Neural Network}

The perceptron is one of the first models that can be classified as an artificial neuron, and it is the basis for modern neural networks. Proposed by Frank Rosenblatt in 1958, the perceptron was designed as a linear classifier capable of mapping input features to binary outputs \cite{rosenblatt_perceptron}. It consists of just one layer of neurones, which makes it simple but crucial to understanding how artificial neural networks work.

The perceptron computes the weighted sum of its inputs, applies an activation function and adds a bias term. It can be expressed as:

\begin{equation}
    y = \text{step}\left( \sum_{i=1}^{n} w_i x_i + b \right),
\end{equation}

where:
\begin{itemize}
    \item \( x_i \in \mathbb{R} \) represents the input features.
    \item \( w_i \in \mathbb{R} \) are the weights associated with each input.
    \item \( b \in \mathbb{R} \) is the bias term.
    \item \( \text{step}(\cdot) \) is the step function, which outputs 1 if the weighted sum is positive and 0 otherwise.
\end{itemize}

It can be graphically expressed as the following image

\begin{figure}[H]
	\centering
	\includegraphics[width=11cm]{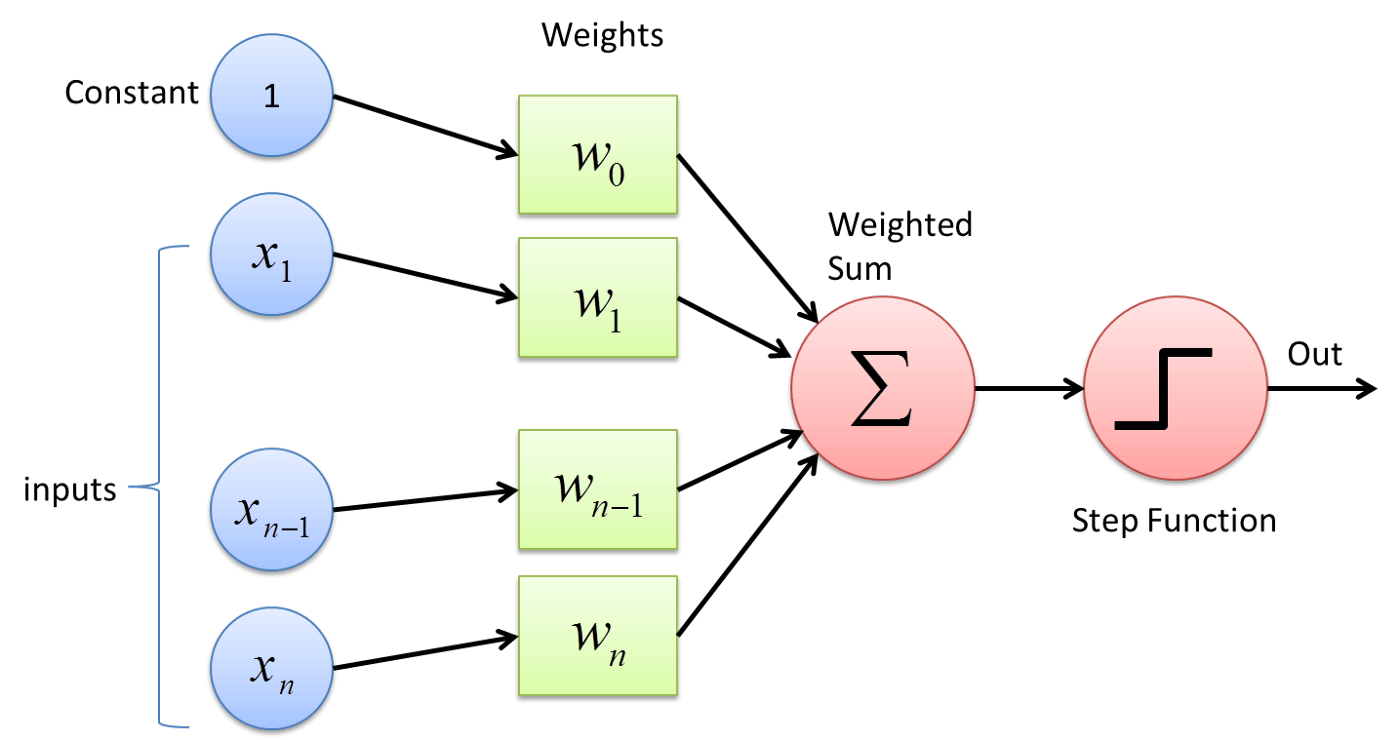}
	\caption{The Perceptron Schema}
	Fonte: Adapted from \cite{perceptron_image}
    \label{fig:neuralNode}
\end{figure}

\subsubsection{Training the Perceptron}

The perceptron is trained using a simple algorithm called the perceptron learning rule, which adjusts the weights based on the prediction error. The update rule for the weights is given by:

\begin{equation}
    w_i^{(t+1)} = w_i^{(t)} + \Delta w_i,
\end{equation}

where the weight adjustment \( \Delta w_i \) is computed as:

\begin{equation}
    \Delta w_i = \eta (y_{\text{true}} - y_{\text{pred}}) x_i.
\end{equation}

Here:
\begin{itemize}
    \item \( \eta \) is the learning rate, controlling the step size of the weight update.
    \item \( y_{\text{true}} \) is the true label of the input.
    \item \( y_{\text{pred}} \) is the predicted output of the perceptron.
\end{itemize}

This training rule ensures that the perceptron can iteratively learn to correctly classify linearly separable data by minimizing the prediction error \cite{mitchell_ml}.

\subsubsection{Observations}

It is important to note that the perceptron learning rule is not the same as gradient descent. While the goal of both methods it to minimize prediction errors, the perceptron uses a discrete update rule that adjusts the weights only when a misclassification occurs, without using derivatives or a continuous loss function. 

In contrast, gradient descent is a more general optimization technique that updates weights based on the gradient of a differentiable loss function, even for small errors. 

Therefore, although the perceptron performs a form of "learning through errors", it cannot be classified as gradient-based optimization methods.

Another important factor to consider is that even though the perceptron uses a non-linear activation function like the step function, it is still a linear classifier. This is because the decision boundary is a linear combination of the input features. Even with just one neuron and a non-linear activation like sigmoid, the output function is non-linear when it comes to mapping inputs to outputs. However, when it comes to classification, the decision boundary (the surface where output crosses the threshold, e.g. 0.5) is still linear.

The model is still a linear classifier even if the step function is replaced with a sigmoid or softmax activation. This is because the classification still depends on whether the input is on one side of a linear hyperplane. 

Non-linear decision boundaries only emerge when multiple layers are stacked with non-linear activation functions in between — such as ReLU or sigmoid — allowing the network to learn complex, non-linear patterns. Therefore, it is not just the activation function that makes a model truly non-linear, but the combination of several non-linear transformations.

This is the reason to the next subtopic:

\subsubsection{Limitations of the Perceptron}

The perceptron performs successfully for data that can be linearly separated but not for non-linear domains, like the XOR problem. This limitation led to the study of more complicated architectures, such as the multi-layer perceptron (MLP), which uses several layers and non-linear activation functions to solve more complex problems \cite{mcculloch_pitts}.

The perceptron is still an important component of the history of neural networks and is a crucial study towards understanding more advanced models \cite{goodfellow_dl}.

\subsection{From linear to non-linear: The Multilayer Perceptron (MLP)}

The Multilayer Perceptron (MLP) is an extension of the perceptron that consists of multiple layers of neurons, enabling it to learn complex patterns and solve problems that are not linearly separable. Unlike the single-layer perceptron, which can only model linear decision boundaries, MLPs use multiple layers and non-linear activation functions to learn sophisticated relationships between inputs and outputs \cite{goodfellow_dl}.

An MLP consists of at least three types of layers:

\begin{itemize}
    \item \textbf{Input Layer:} Receives the feature vector \( x \).
    \item \textbf{Hidden Layers:} One or more layers of neurons that transform the inputs using weighted connections and non-linear activation functions.
    \item \textbf{Output Layer:} Produces the final prediction.
\end{itemize}

Visually, it can be expressed by the following image:

\begin{figure}[H]
	\centering
	\includegraphics[width=11cm]{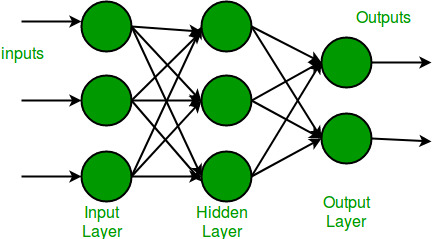}
	\caption{The Perceptron Schema}
	Fonte: Adapted from \cite{mlp_image}
    \label{fig:neuralNode}
\end{figure}

\subsubsection{Mathematical Formulation}

Each neuron in the MLP computes a weighted sum of its inputs, applies an activation function \( f(\cdot) \), and propagates the result to the next layer. The output of a neuron in a hidden layer can be expressed as:

\begin{equation}
    h_j = f\left( \sum_{i=1}^{n} w_{ji} x_i + b_j \right),
\end{equation}

where:
\begin{itemize}
    \item \( x_i \) are the input features,
    \item \( w_{ji} \) are the weights connecting input neuron \( i \) to hidden neuron \( j \),
    \item \( b_j \) is the bias term, and
    \item \( f(\cdot) \) is a non-linear activation function, such as the ReLU or sigmoid function.
\end{itemize}

The output layer applies a final transformation:

\begin{equation}
    y_k = f_{\text{out}} \left( \sum_{j=1}^{m} v_{kj} h_j + c_k \right),
\end{equation}

where:
\begin{itemize}
    \item \( v_{kj} \) are the weights between the hidden layer and output neuron \( k \),
    \item \( c_k \) is the bias term,
    \item \( f_{\text{out}}(\cdot) \) is the activation function used in the output layer (e.g., softmax for multi-class classification tasks, or sigmoid for binary classification tasks).
\end{itemize}

\subsubsection{Training the MLP}

The MLP is trained using the backpropagation algorithm, which minimizes the error between the predicted and true outputs by updating the network weights iteratively. The learning process involves:

\begin{enumerate}
    \item \textbf{Forward Propagation:} Computing activations for each layer.
    \item \textbf{Error Calculation:} Measuring the difference between predicted and actual labels using a loss function, such as cross-entropy for classification or mean squared error (MSE) for regression.
    \item \textbf{Backward Propagation:} Using the gradient of the loss function to update the weights via stochastic gradient descent (SGD) or other optimizers like Adam.
\end{enumerate}

The weight updates follow:

\begin{equation}
    w_{ji}^{(t+1)} = w_{ji}^{(t)} - \eta \frac{\partial \mathcal{L}}{\partial w_{ji}},
\end{equation}

where \( \mathcal{L} \) is the loss function, and \( \eta \) is the learning rate.

\subsubsection{Advantages over the Perceptron and Limitations of MLPs}

\textbf{Advantages:}
\begin{itemize}
    \item Can approximate any continuous function given enough hidden layers and neurons (\textit{Universal Approximation Theorem}).
    \item Handles non-linearly separable problems like XOR.
    \item Supports various activation functions to introduce non-linearity.
\end{itemize}

\textbf{Limitations:}
\begin{itemize}
    \item Computationally expensive for large networks.
    \item Vulnerable to overfitting, requiring techniques like dropout \cite{srivastava2014dropout}.
    \item Training can be slow and requires careful selection of hyperparameters.
\end{itemize}

MLPs form the basis of deep learning models, where deeper architectures allow for higher levels of abstraction and generalization. More advanced networks, such as Convolutional Neural Networks (CNNs) and Recurrent Neural Networks (RNNs), build upon the MLP framework to specialize in tasks such as image and sequence processing.

\subsection{Improving training performance: Optimizers}

\begin{figure}[H]
	\centering
	\includegraphics[width=11cm]{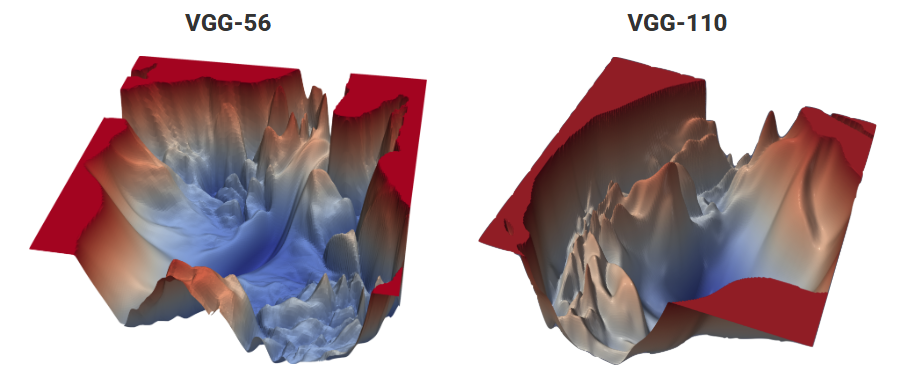}
	\caption{Loss landscapes without skip connections. It shows the loss function of a 56-layer and 110-layer net using the CIFAR-10 dataset, without residual connections. \cite{li2018visualizing}}
    \label{fig:loss_landscape}
\end{figure}

Some models have to work in very complex loss landscapes, and it is not easy to determine and follow a good gradient direction. The image \ref{fig:loss_landscape} shows an example of this fact training in the CIFAR-10 dataset.

Therefore, optimizers were studied to update the weights of a neural network to minimize the loss function during training in an efficient way. Different optimizers have different strategies for adjusting the weights to improve the convergence speed and generalization.

Among the various optimization algorithms available, this section will focus on two optimizers used in this project:

\subsubsection{Stochastic Gradient Descent (SGD)}

One of the most fundamental and common optimization techniques for neural network training is the Stochastic Gradient Descent (SGD). SGD updates the model parameters using only one training example (or a small mini-batch) at each iteration, as opposed to traditional \textit{batch gradient descent}, which computes the gradient of the loss function over the entire dataset. Although noise is added to the optimisation process, there are also major computational and generalization advantages.

The update rule is expressed as:

\begin{equation}
    w^{(t+1)} = w^{(t)} - \eta \nabla \mathcal{L}(w),
\end{equation}

where \( w^{(t)} \) are the model parameters at step \( t \), \( \eta \) is the learning rate, and \( \nabla \mathcal{L}(w) \) is the gradient of the loss with respect to the parameters. In practice, this gradient is often computed over a mini-batch of data points, which balances stability and efficiency.

\begin{table}[H]
\centering
\caption{Advantages and disadvantages of Stochastic Gradient Descent (SGD).}
\arrayrulecolor[gray]{0.7}
\begin{tabular}{|p{0.45\linewidth}|p{0.45\linewidth}|}
\hline
\textbf{Pros} & \textbf{Cons} \\
\hline
Simple, easy to implement, and computationally efficient. & Susceptible to oscillations around minima, especially with noisy gradients. \\
\arrayrulecolor[gray]{0.9}\hline
Scales well to large datasets, since updates do not require processing all data at once. & Requires careful tuning of the learning rate and may converge slowly. \\
\arrayrulecolor[gray]{0.9}\hline
Introduces stochasticity that can help escape narrow local minima, improving generalization. & Sensitive to the choice of hyperparameters (batch size, learning rate schedule). \\
\arrayrulecolor[gray]{0.7}\hline
\end{tabular}
\arrayrulecolor{black}
\end{table}

\subsubsection{Momentum}

Momentum is an extension of SGD that helps accelerate convergence by accumulating a moving average of past gradients. This smooths the parameter updates, reducing oscillations and making it easier to traverse narrow valleys in the loss surface.

The update rules are given by:

\begin{equation}
    v_t = \beta v_{t-1} + (1 - \beta) \nabla \mathcal{L}(w),
\end{equation}

\begin{equation}
    w^{(t+1)} = w^{(t)} - \eta v_t,
\end{equation}

where \( \beta \in [0,1) \) is the momentum coefficient that controls the contribution of past gradients.

\subsubsection{Adam (Adaptive Moment Estimation)}

Adam is one of the most popular optimizers, as it combines the benefits of momentum and adaptive learning rates. It maintains two moving averages: the first moment (\(m_t\)) corresponding to the mean of gradients, and the second moment (\(v_t\)) corresponding to the uncentered variance of gradients.

\begin{equation}
    m_t = \beta_1 m_{t-1} + (1 - \beta_1) \nabla \mathcal{L}(w),
\end{equation}

\begin{equation}
    v_t = \beta_2 v_{t-1} + (1 - \beta_2) \nabla \mathcal{L}(w)^2,
\end{equation}

where \( \beta_1 \) and \( \beta_2 \) are decay rates for the moving averages.  
The parameter update rule is:

\begin{equation}
    w^{(t+1)} = w^{(t)} - \eta \frac{m_t}{\sqrt{v_t} + \epsilon},
\end{equation}

where \( \epsilon \) is a small constant added for numerical stability.

\begin{table}[H]
\centering
\caption{Advantages and disadvantages of Adam optimizer.}
\arrayrulecolor[gray]{0.7}
\begin{tabular}{|p{0.45\linewidth}|p{0.45\linewidth}|}
\hline
\textbf{Pros} & \textbf{Cons} \\
\hline
Combines momentum and adaptive learning rates, leading to faster convergence than SGD. & May not generalize as well as plain SGD in some cases. \\
\arrayrulecolor[gray]{0.9}\hline
Automatically adapts the learning rate for each parameter. & Requires tuning of multiple hyperparameters (\(\beta_1, \beta_2, \eta\)). \\
\arrayrulecolor[gray]{0.9}\hline
Well-suited for problems with sparse or noisy gradients. & Can converge to suboptimal solutions if hyperparameters are not well chosen. \\
\arrayrulecolor[gray]{0.7}\hline
\end{tabular}
\arrayrulecolor{black}
\end{table}

The strengths and weaknesses of different optimizers are complementary. The dataset's properties, the problem being solved, and the intended trade-off between generalization and convergence speed all influence the optimal option.

\subsection{Artificially increasing the amount of data: Data Augmentations}

Typically, training Deep Learning models requires a large amount of data. These data may not be so abundant. Therefore, using data augmentations is a very common method to artificially increase the amount of data.

Data augmentation is the process of applying various transformations to the original data in order to artificially increase the size and diversity of a dataset. This is interesting because it provides ``new'' data by simply modifying the already existing data. By generating new variations, data augmentation helps prevent overfitting, improve generalization, and enhance model performance \cite{shorten2019survey}.

\subsubsection{Types of Data Augmentation}

Different types of data augmentation techniques are used depending on the nature of the dataset. Common augmentation methods include:

\begin{itemize}
    \item \textbf{Geometric Transformations:} Modifying images by applying operations such as rotation, scaling, translation, flipping, and cropping.
    
    \item \textbf{Color and Intensity Transformations:} Adjusting brightness, contrast, saturation, hue, or converting to grayscale.
    
    \item \textbf{Noise Injection:} Adding random noise (Gaussian, salt-and-pepper, speckle noise) to increase robustness against variations in input data.
    
    \item \textbf{Cutout and Mixup:} Techniques like Cutout remove random portions of an image, while Mixup blends multiple images and their labels to create new training examples \cite{zhang2017mixup}.
    
    \item \textbf{Adversarial Augmentation:} Generates perturbed inputs designed to mislead the model, therefore increasing robustness against adversarial attacks \cite{goodfellow2015explaining}.
    
    \item \textbf{Feature Space Augmentation:} Instead of modifying the raw data, augmentations can be applied in the feature space of a neural network to generate diverse embeddings.
\end{itemize}

\begin{figure}[H]
	\centering
	\includegraphics[width=11cm]{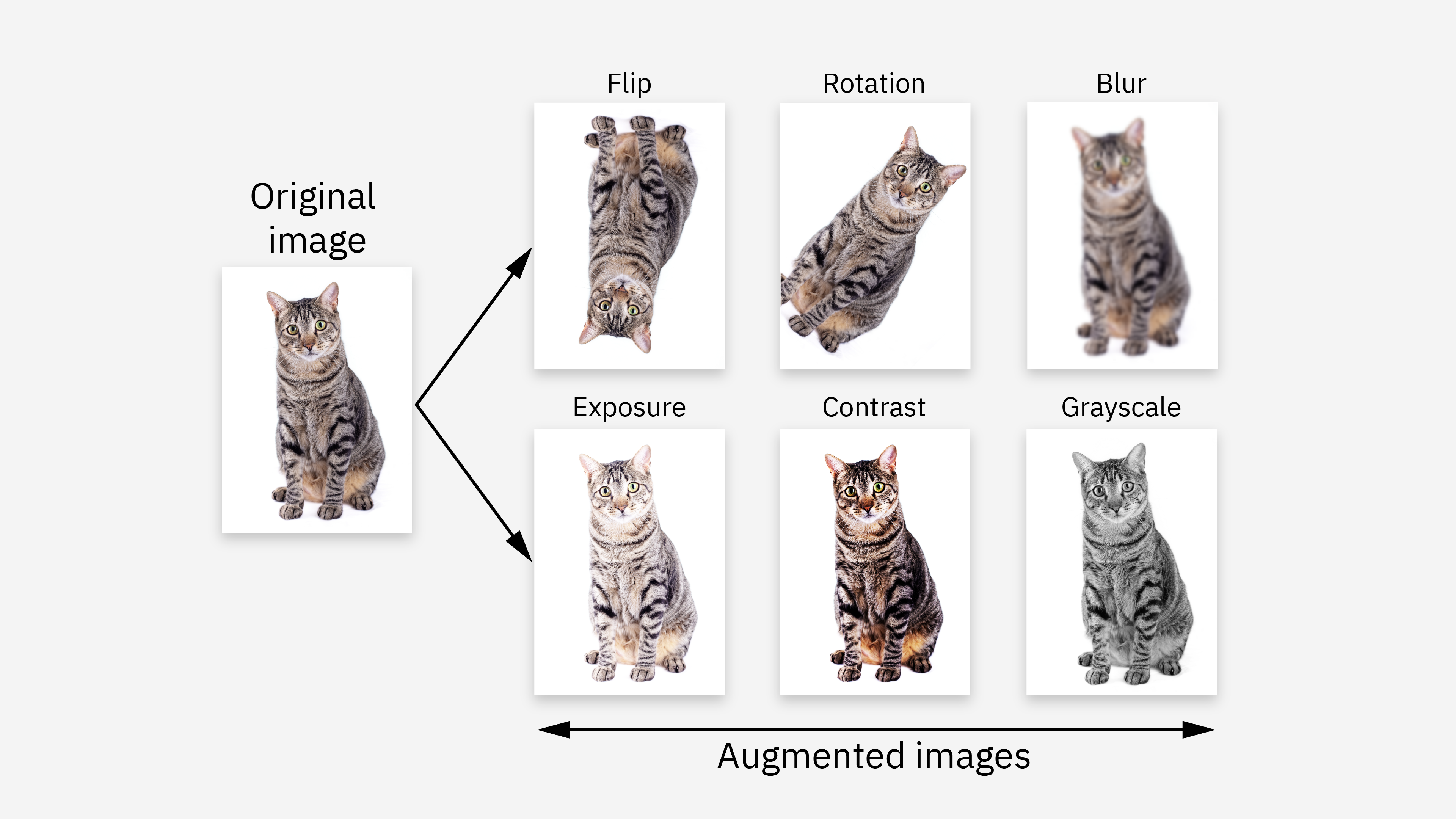}
	\caption{Data Augmentation Examples}
	Fonte: Adapted from \cite{DA_IBM}
    \label{fig:neuralNode}
\end{figure}

\section{Dealing with images: Convolutional Networks}

Before the introduction of Convolutional Neural Networks, it was common practice to process images by flattening them into one-dimensional vectors before feeding them into fully connected neural networks. However this method results in a lot of parameters, especially with high-resolution images, making the models computationally costly and susceptible to overfitting.

Additionally, the spatial relationships between pixels—which are crucial for comprehending visual patterns—are eliminated by flattening. Convolutional neural networks were created to address these constraints by utilizing weight sharing and local connections to reduce the number of parameters and better capture spatial hierarchies.

It is a specialized neural network made to process data that has a grid-like structure, like pictures \cite{LeCun1998}. CNNs are very powerful at image-related tasks like segmentation, object detection, and classification.

The convolution operation is mathematically expressed as:

\[
S(i, j) = \sum_{m=1}^{M} \sum_{n=1}^{N} X(i+m, j+n) \cdot K(m, n),
\]

where:
\begin{itemize}
    \item \( S(i, j) \in \mathbb{R} \) represents the output feature map,
    \item \( X(i, j) \in \mathbb{R} \) is the input data,
    \item \( K(m, n) \in \mathbb{R} \) is the kernel or filter, and
    \item \( M, N \) are the dimensions of the kernel.
\end{itemize}

Below is an example of a convolution operation. The matrix \( I \) represents the input data to be processed. A filter \( K \) (also called a kernel) is applied to \( I \) using the convolution operation to produce the output matrix \( I * K \), which corresponds to the resulting feature map. Convolution involves sliding the filter over the input matrix and computing the dot product at each position.

\vspace{0.5cm}

Let the input matrix be:

\[
I = \begin{pmatrix}
2 & 3 & 0 & 1 \\
1 & 5 & 2 & 3 \\
4 & 0 & 1 & 2 \\
3 & 2 & 4 & 1
\end{pmatrix}
\]

And the kernel (filter):

\[
K = \begin{pmatrix}
0 & 1 \\
-1 & 0
\end{pmatrix}
\]

Applying the convolution operation \( I * K \) (valid padding, stride 1), we obtain the output matrix:

\[
I * K = \begin{pmatrix}
2 & -5 & -1 \\
1 & 2 & 2 \\
-3 & -1 & -2
\end{pmatrix}
\]  

Pooling layers are often applied after convolution to downsample the spatial dimensions of the feature maps, reducing computational costs and mitigating overfitting \cite{Krizhevsky2012}. For max pooling, the output is defined as:

\[
P(i, j) = \max_{m, n \in R} S(i+m, j+n),
\]

where \( R \) defines the pooling region. Pooling layers improve computational efficiency by reducing the number of parameters and make the model invariant to small translations in the input data.

For example, given the following feature map:

\[
S = 
\begin{pmatrix}
2 & -5 & -1 & 3 \\
1 & 2 & 2 & 0 \\
-3 & -1 & -2 & 4 \\
0 & 1 & 3 & -2
\end{pmatrix}
\]

Applying a max pooling operation with a $2 \times 2$ window and stride 2, we divide the feature map into non-overlapping $2 \times 2$ regions:

\[
\begin{pmatrix}
\color{red}{\boxed{2}} & \color{red}{\boxed{-5}} & \color{green}\boxed{-1} & \color{green}\boxed{3} \\
\color{red}{\boxed{1}} & \color{red}{\boxed{2}} & \color{green}\boxed{2} & \color{green}\boxed{0} \\
\boxed{-3} & \boxed{-1} & \color{blue}{\boxed{-2}} & \color{blue}{\boxed{4}} \\
\boxed{0} & \boxed{1} & \color{blue}{\boxed{3}} & \color{blue}{\boxed{-2}} \\
\end{pmatrix}
\Rightarrow
\begin{pmatrix}
\max(2, -5, 1, 2) & \max(-1, 3, 2, 0) \\
\max(-3, -1, 0, 1) & \max(-2, 4, 3, -2)
\end{pmatrix}
=
\begin{pmatrix}
2 & 3 \\
1 & 4
\end{pmatrix}
\]

Thus, the result of the max pooling operation is:

\[
P = 
\begin{pmatrix}
2 & 3 \\
1 & 4
\end{pmatrix}
\]

The combination of convolutional layers and pooling layers allows CNNs to hierarchically extract features, where lower layers capture basic patterns such as edges and textures, and higher layers learn more complex features like shapes and objects. This hierarchical feature learning is a key reason CNNs have become the foundation for most modern deep learning applications in computer vision \cite{LeCun2015}.

\subsection{ResNet-34}

The ResNet-34 \cite{He2016} was created to address difficulties with training very deep networks, such as the vanishing gradient problem. Its main innovation is its use of residual connections, which let the network learn residual mappings rather than the intended transformation directly. This enhances deep architecture performance and simplifies optimization.

The residual block, which is the fundamental building block of ResNet-34, can be mathematically expressed as:

\[
y = F(x, \{W_i\}) + x,
\]

where:
\begin{itemize}
    \item \( x \) is the input to the block,
    \item \( F(x, \{W_i\}) \) represents the output of the convolutional layers within the block,
    \item \( \{W_i\} \) are the weights of the convolutional layers, and
    \item \( y \) is the output of the block.
\end{itemize}

The addition of \( x \) (known as the skip connection) ensures that the network retains the input information, enabling the gradient to flow directly through the network during backpropagation. This design alleviates the degradation problem observed in very deep networks \cite{He2016}.

Skip connections not only speed up training and enhance information flow, but they also help in backpropagation by smoothing the gradient map. They reduce the possibility of vanishing or exploding gradients in deep networks by offering alternate routes for the gradient to travel. Weight updates become more consistent as a result, making training more reliable and effective. This fact can be observed in figure \ref{fig:landscape_resnet}:

\begin{figure}[H]
	\centering
	\includegraphics[width=11cm]{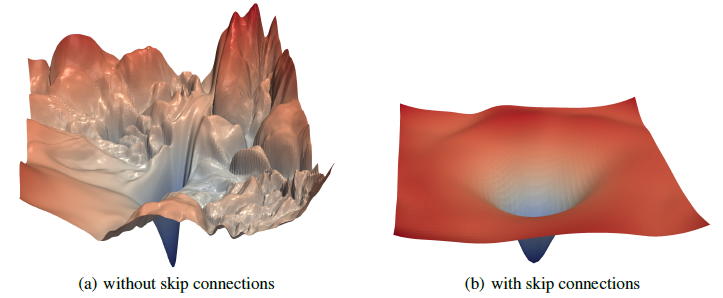}
	\caption{Effect of the skip connection}
	Fonte: Adapted from \cite{li2018visualizing}
    \label{fig:landscape_resnet}
\end{figure}

ResNet-34 consists of 34 layers, including convolutional layers, batch normalization layers, and ReLU activations. The architecture processes input images hierarchically, with earlier layers capturing low-level features such as edges and textures, while deeper layers learn high-level semantic information. This hierarchical feature extraction makes ResNet-34 an effective backbone for many deep learning tasks.

\begin{figure}[H]
	\centering
	\includegraphics[width=14cm]{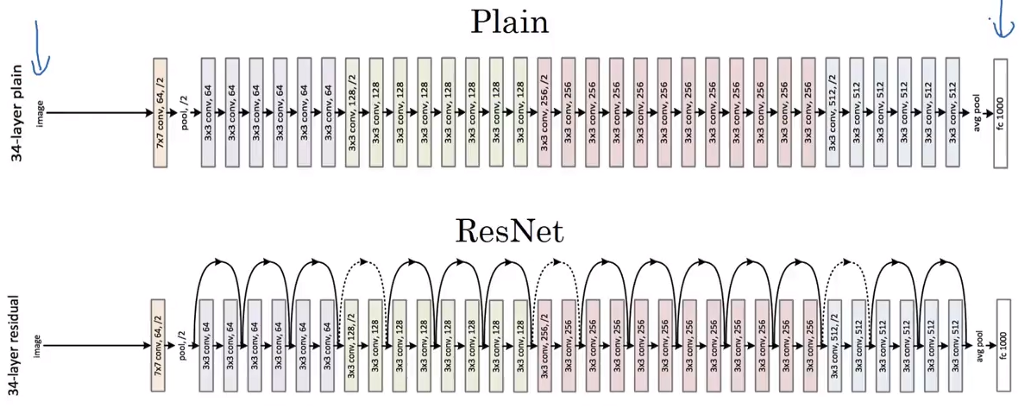}
	\caption{ResNet model in comparison to a plain model without skip connections}
    \label{fig:resnet_model}
\end{figure}

ResNet-34 is often used as the feature extractor and backbone of different neural network architectures, such as Siamese Networks. ResNet-34 creates an embedding \( h \) in a high-dimensional feature space given an input image \( x \):

\[h = \text{ResNet34}(x).\]

In these configurations, a smaller embedding layer usually takes the place of the last fully connected layer of ResNet-34 in order to address a particular task. For tasks like object recognition, metric learning, and face verification, its strong feature extraction capability makes it ideal \cite{Huang2017}.

\section{Generating Embeddings: Feature Learning}

Neural networks can automatically extract meaningful representations from unprocessed input data thanks to feature learning. While the deeper layers of a CNN represent high-level semantic information, like objects or shapes, the early layers capture low-level features, like edges and textures.

Instead of being applied directly to classification and/or regression, feature-learning losses are made to create embeddings. Following training with these losses, a simple threshold, KNN, or another classifier must be used to discretize these embeddings in the case of signature matching (a classification task).

\subsection{Siamese Networks}

Siamese Networks are a type of neural network architecture designed to learn a similarity metric between two inputs. The two backbones of the network are identical and have the same architecture and weights. These sub-networks create embeddings in a common feature space after processing two input samples separately. The objective is to maximize the distance for dissimilar pairs and minimize the distance or a similarity function between embeddings of similar pairs.

The operation of a Siamese Network can be described as:

\[
h_1 = f_{\theta}(x_1), \quad h_2 = f_{\theta}(x_2),
\]

where:
\( x_1, x_2 \in \mathbb{R}^d \) are the input samples,
\( f_{\theta} \) is the shared subnetwork with parameters \( \theta \), and
\( h_1, h_2 \in \mathbb{R}^k \) are the embeddings produced for the inputs.

The similarity or dissimilarity between the embeddings can be quantified using various distance metrics or similarity measures. One commonly used metric is the Euclidean distance:

\[
D_{\text{Euclidean}} = \sqrt{\sum_{k=1}^{K} (h_{1,k} - h_{2,k})^2},
\]

where \( K \) is the dimensionality of the embedding space.

Other examples of distance metrics include:
\begin{itemize}
    \item \textbf{Manhattan Distance (L1 norm):}
    \[
    D_{\text{Manhattan}} = \sum_{k=1}^{K} |h_{1,k} - h_{2,k}|,
    \]
    which measures the absolute difference between embeddings along each dimension.

    \item \textbf{Cosine Similarity:}
    \[
    S_{\text{Cosine}} = \frac{\sum_{k=1}^{K} h_{1,k} \cdot h_{2,k}}{\|h_1\|_2 \cdot \|h_2\|_2},
    \]
    where \( \|h_1\|_2 \) and \( \|h_2\|_2 \) are the Euclidean norms of the embeddings. Cosine similarity measures the angular similarity between vectors, ranging from \(-1\) (opposite) to \(1\) (identical).
\end{itemize}

Below, at figure \ref{fig:contrastive_net} we can visually understand the architecture of the Siamese network with some distance metric and a loss funtction to determine the map:

\begin{figure}[H]
	\centering
	\includegraphics[width=15cm]{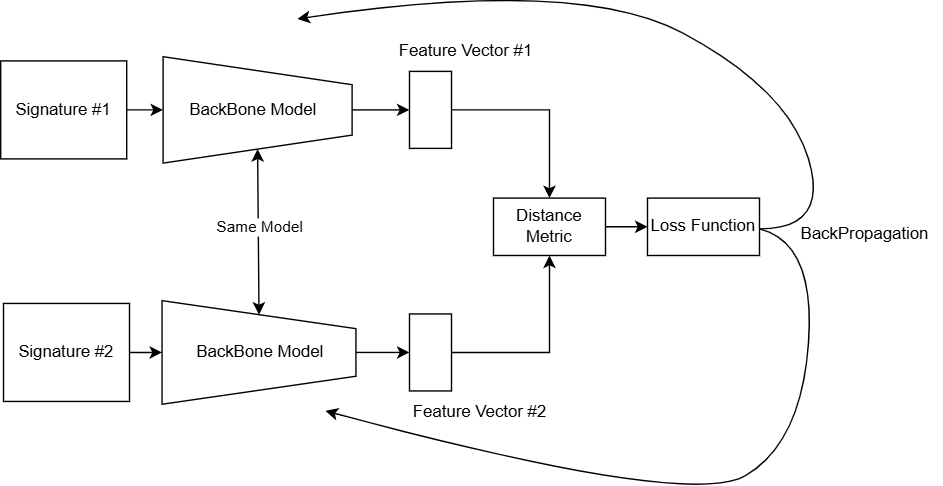}
	\caption{High-level Siamese Network architecture using some Distance metric}
	Source: Author
    \label{fig:contrastive_net}
\end{figure}

Siamese Networks are commonly trained using loss functions such as Contrastive Loss \cite{Hadsell2006} or Triplet Loss \cite{Schroff2015}.

Therefore, the signature forgery detection can be formalized as follows: given two input signatures \( x_1 \) and \( x_2 \), the Siamese network aims to learn an embedding function \( f_\theta(\cdot) \) such that the distance between their embeddings is minimized if the signatures are from the same person (genuine pair), and maximized if they are from different people (forged pair).

\[
z_1 = f_\theta(x_1), \quad z_2 = f_\theta(x_2)
\]

Then, the network is trained so that:

\[
\text{distance}(z_1, z_2) =
\begin{cases}
\text{small}, & \text{if } x_1 \text{ and } x_2 \text{ are genuine pair} \\
\text{large}, & \text{if } x_1 \text{ and } x_2 \text{ are forged pair}
\end{cases}
\]

In many applications, ResNet-34 is employed as the backbone of Siamese Networks to extract robust and meaningful features from the input data. And this work will also use it.

\subsection{Contrastive Loss}

Contrastive Loss is widely used in metric learning tasks, particularly for Siamese Networks \cite{Bromley1993}. Its objective is to learn a feature space where embeddings of similar pairs are close, and embeddings of dissimilar pairs are separated by at least a margin. The loss function is defined as:

\[
L = \frac{1}{N} \sum_{i=1}^{N} \left[ (1 - y) \cdot D^2 + y \cdot \max(0, m - D)^2 \right],
\]

where:
\begin{itemize}
    \item \( y \in \{0, 1\} \) is a binary label indicating whether the pair is similar (\( y = 0 \)) or dissimilar (\( y = 1 \)),
    \item \( D \in \mathbb{R}_{\geq 0} \) is the Euclidean distance between the embeddings, and
    \item \( m > 0 \) is the margin that separates dissimilar pairs.
\end{itemize}

The Euclidean distance, which quantifies the similarity between two embeddings, is calculated as:

\[
D = \sqrt{\sum_{k=1}^{d} (h_{i, k} - h_{j, k})^2},
\]

where:
\begin{itemize}
    \item \( h_i \in \mathbb{R}^d \) and \( h_j \in \mathbb{R}^d \) are the embeddings of the two samples in a \( d \)-dimensional feature space,
    \item \( d \) is the dimensionality of the embeddings.
\end{itemize}

The margin \( m \) enforces a minimum distance between dissimilar pairs, enhancing the feature space's discriminative capability. This loss function was first introduced in the context of time delay neural networks by \cite{Bromley1993}, and later formalized and extended in metric learning by \cite{Hadsell2006}.

Face verification, signature verification, and general metric learning problems are just a few of the tasks where Contrastive Loss has been successfully used. It is a popular option for training models to learn similarity metrics because of its interpretability and simplicity \cite{Hadsell2006}.

\subsection{Triplet Loss}

Triplet Loss is a fundamental objective function in metric learning, commonly applied in tasks such as face verification and person re-identification \cite{Schroff2015}. Its objective is to ensure that, within a specified margin \( m \), the embeddings of a sample and a similar sample (positive) are closer than those of a sample and a dissimilar sample (negative) by a defined margin \( m \).  

\begin{figure}[H]
	\centering
	\includegraphics[width=12cm]{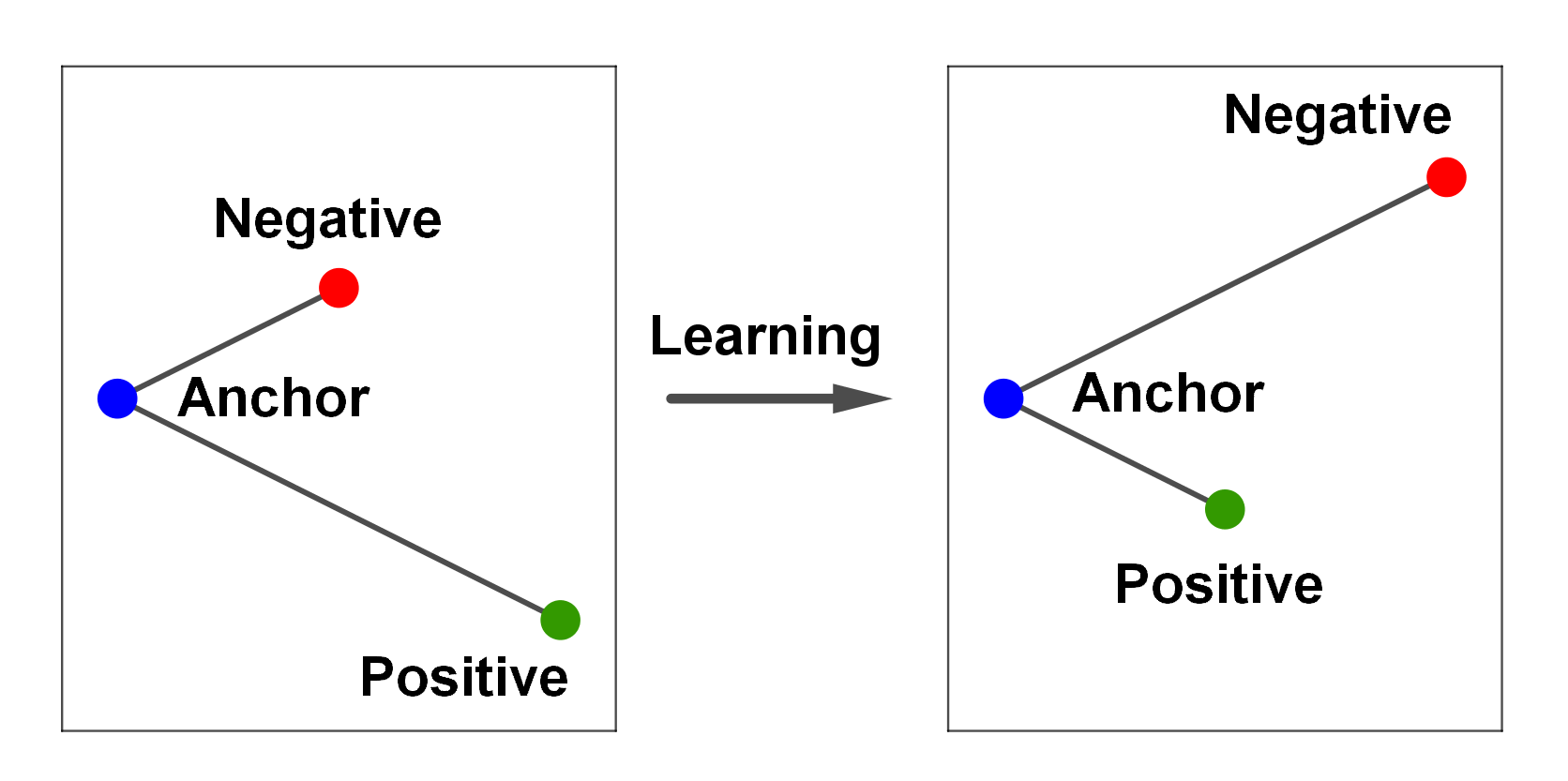}
	\caption{High-level Siamese Network architecture using some Distance metric}
	Source: \cite{wikipedia_triplet_loss}
    \label{fig:contrastive_net}
\end{figure}

The loss is formulated as:

\[
L = \frac{1}{N} \sum_{i=1}^{N} \left[ \max(0, D(h_i, h_i^+) - D(h_i, h_i^-) + m) \right],
\]

where:
\begin{itemize}
    \item \( h_i \) is the embedding of the anchor sample,
    \item \( h_i^+ \) is the embedding of the positive (similar) sample,
    \item \( h_i^- \) is the embedding of the negative (dissimilar) sample,
    \item \( D(\cdot, \cdot) \) is a distance function (commonly Euclidean distance),
    \item \( m > 0 \) is the margin that enforces a separation between the positive and negative pairs.
\end{itemize}

By minimizing the distance between the anchor-positive pair and maximizing the distance between the anchor-negative pair, the loss ensures a margin \( m \) between the two, therefore optimizing the feature space.

While both Triplet Loss and Contrastive Loss are used in metric learning, they differ in how they define and optimize the embedding space:
\begin{itemize}
    \item \textbf{Contrastive Loss} focuses on pairs of samples, classifying them as similar or dissimilar. It explicitly enforces a margin \( m \) for dissimilar pairs to be separated in the embedding space.
    \item \textbf{Triplet Loss} extends this concept by comparing three samples (anchor, positive, and negative). It enforces relative distances between these samples, ensuring that the anchor-positive distance is smaller than the anchor-negative distance by at least \( m \).
\end{itemize}

In practice, Triplet Loss often requires more sophisticated sampling strategies (e.g., mining hard triplets) to achieve optimal performance. Contrastive Loss, on the other hand, can be easier to implement but may lack the "fine-tuned" relational learning achieved by Triplet Loss.

Triplet Loss is expected to improve performance by generating a more discriminative feature representation as it places stricter constraints on the embedding space. This idea will be explored throughout the work.

\section{Evaluation Methods}

Several evaluation techniques, such as the Area Under the Curve (AUC), the confusion matrix, and the Receiver Operating Characteristic (ROC) curve, are used to evaluate the performance of the proposed approach. These metrics provide a comprehensive understanding of the model's classification abilities.

\subsection{Confusion Matrix}

The confusion matrix is a tabular representation of a classifier’s predictions, comparing the predicted and actual classes. For binary classification, it is structured as:

\[
\begin{bmatrix}
    \text{TP} & \text{FP} \\
    \text{FN} & \text{TN}
\end{bmatrix}
\]

where:
\begin{itemize}
    \item \textbf{TP (True Positives)}: The number of positive samples correctly classified.
    \item \textbf{FP (False Positives)}: The number of negative samples incorrectly classified as positive.
    \item \textbf{FN (False Negatives)}: The number of positive samples incorrectly classified as negative.
    \item \textbf{TN (True Negatives)}: The number of negative samples correctly classified.
\end{itemize}

The figure below \ref{fig:confusion_matrix_example} shows an example of the confusion matrix graph:

\begin{figure}[H]
	\centering
	\includegraphics[width=11cm]{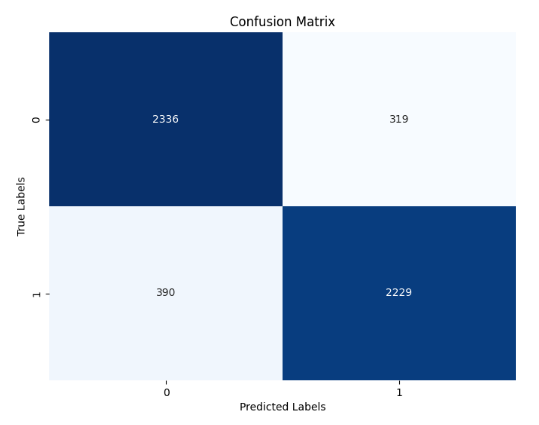}
	\caption{Confusion Matrix Example}
	Source: Author
    \label{fig:confusion_matrix_example}
\end{figure}

The diagonal elements (2336 and 2229) represent the correctly classified instances, while the off-diagonal elements (319 and 390) represent the misclassified instances.

From the confusion matrix, several evaluation metrics can be derived:

\begin{itemize}
    \item \textbf{Accuracy:} Measures the overall correctness of the model’s predictions.
    \begin{equation}
        \text{Accuracy} = \frac{\text{TP} + \text{TN}}{\text{TP} + \text{FP} + \text{FN} + \text{TN}}
    \end{equation}

    \item \textbf{Precision:} Also known as Positive Predictive Value, it measures the proportion of true positive predictions out of all positive predictions.
    \begin{equation}
        \text{Precision} = \frac{\text{TP}}{\text{TP} + \text{FP}}
    \end{equation}

    \item \textbf{Recall (Sensitivity or True Positive Rate):} Measures the proportion of actual positive samples correctly identified by the model.
    \begin{equation}
        \text{Recall} = \frac{\text{TP}}{\text{TP} + \text{FN}}
    \end{equation}

    \item \textbf{F1-Score:} The harmonic mean of Precision and Recall, providing a single measure that balances these two metrics.
    \begin{equation}
        \text{F1} = 2 \cdot \frac{\text{Precision} \cdot \text{Recall}}{\text{Precision} + \text{Recall}}
    \end{equation}
\end{itemize}

\subsection{Receiver Operating Characteristic (ROC) Curve}

The ROC curve is a graphical representation of a classifier’s performance across various threshold values. It plots the True Positive Rate (TPR) against the False Positive Rate (FPR). These metrics are defined as:

\begin{equation}
    \text{TPR} = \frac{\text{TP}}{\text{TP} + \text{FN}}
\end{equation}

\begin{equation}
    \text{FPR} = \frac{\text{FP}}{\text{FP} + \text{TN}}
\end{equation}

An ideal ROC curve passes close to the top-left corner, indicating a high TPR and low FPR for most thresholds.

\begin{figure}[H]
	\centering
	\includegraphics[width=11cm]{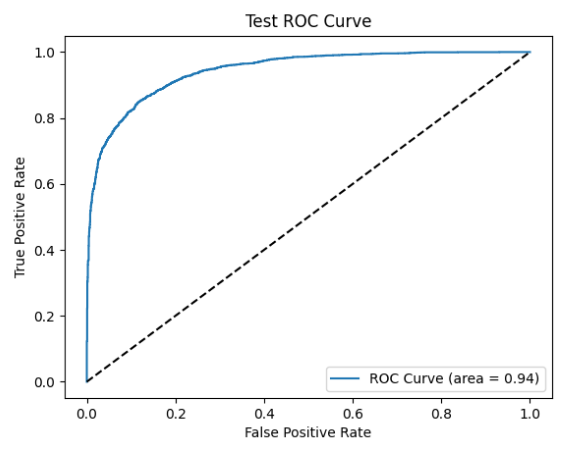}
	\caption{ROC Curve Example}
	Source: Author
    \label{fig:roc_curve_example}
\end{figure}

Figure \ref{fig:roc_curve_example} shows an example of the Receiver Operating Characteristic (ROC) curve, which illustrates the trade-off between the True Positive Rate (TPR) and the False Positive Rate (FPR) across various decision thresholds. The model's capacity to distinguish between classes is illustrated by the ROC curve.

A random classifier (AUC = 0.5) is represented by the dashed black diagonal line, whereas the solid blue line shows the classifier's performance. The model performs better the closer the curve gets to the graph's upper-left corner.

\subsection{Area Under the Curve (AUC)}

The Area Under the Curve (AUC) is a numerical summary of the ROC curve, measuring the overall ability of the model to distinguish between classes. It is calculated as the area under the ROC curve:

\begin{equation}
    \text{AUC} = \int_{0}^{1} \text{TPR}(\text{FPR}) \, d(\text{FPR})
\end{equation}

The AUC value ranges from 0 to 1:
\begin{itemize}
    \item AUC = 1: Indicates perfect classification.
    \item AUC = 0.5: Indicates random guessing or no discriminative power.
    \item AUC $<$ 0.5: Indicates performance worse than random guessing.
\end{itemize}

In the example of figure \ref{fig:roc_curve_example}, the Area Under the Curve (AUC) is 0.94, indicating a strong ability of the model to differentiate between the positive and negative classes. AUC values close to 1 suggest excellent classification performance, whereas values near 0.5 indicate no discriminative power.
  \chapter{Part I – Enhancing Model Generalization: Development}

\section{Dataset Details}

Firstly, the most widely used datasets for signature forgery detection were explored. In this work, it is described three datasets commonly employed in research: CEDAR, ICDAR, and GPDS Synthetic.

\subsection{CEDAR}

The CEDAR dataset is one of the earliest datasets developed for signature verification and forgery detection. It consists of both genuine and forged signatures collected from 55 individuals, with 24 genuine and 24 forged signatures per subject \cite{cedar_dataset}. 

The dataset includes:
\begin{itemize}
    \item Genuine signatures: Handwritten by the actual individual.
    \item Forged signatures: Simulated by a forger attempting to replicate the genuine signatures as closely as possible.
\end{itemize}

In the image below \ref{fig:CEDAR_examples} shows some examples of the CEDAR dataset:

\begin{figure}[H]
	\centering
	\includegraphics[width=11cm]{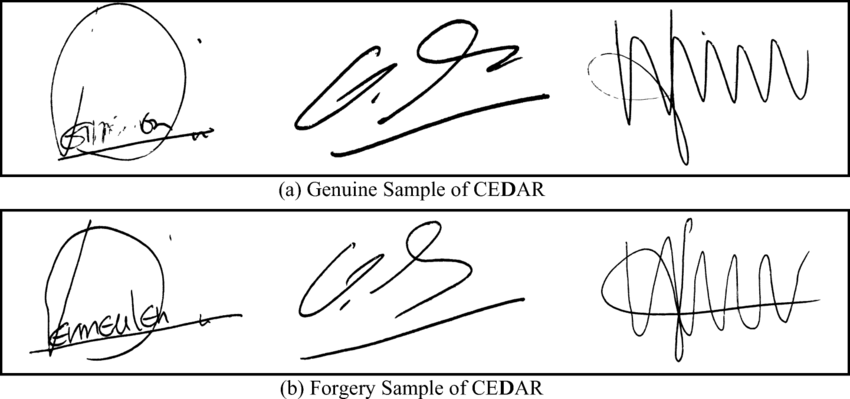}
	\caption{CEDAR Signature Examples}
	Source: \cite{cedar_dataset_reference}
    \label{fig:CEDAR_examples}
\end{figure}

Because it offers high-quality scanned signatures and includes both skilled and random forgeries, CEDAR is frequently used to test different signature verification algorithms. Many signature verification models have been benchmarked using this dataset.

\subsection{ICDAR}

The ICDAR dataset is part of the International Conference on Document Analysis and Recognition (ICDAR) competitions on signature verification \cite{icdar_dataset}. This dataset is divided into two main tracks:
\begin{itemize}
    \item Offline signature verification: Provides scanned images of signatures written on paper.
    \item Online signature verification: Contains dynamic signature data, including pen trajectory, pressure, and speed.
\end{itemize}

ICDAR datasets have been critical for evaluating models in both offline and online signature verification tasks. They include a variety of forgery types:
\begin{itemize}
    \item Random forgeries: Signatures unrelated to the genuine individual.
    \item Skilled forgeries: Carefully created by forgers with access to genuine signature samples.
\end{itemize}

In this study, only offline signatures are explored. 

The image below \ref{fig:ICDAR_examples} shows some examples of the ICDAR dataset:

\begin{figure}[H]
	\centering
	\includegraphics[width=11cm]{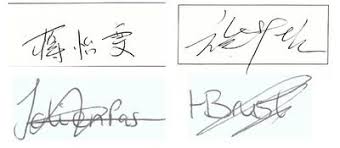}
	\caption{ICDAR Signature Examples}
	Source: ICDAR Dataset
    \label{fig:ICDAR_examples}
\end{figure}

The ICDAR datasets are widely used in competitions to benchmark and compare the performance of signature verification systems across multiple research teams worldwide.

\subsection{GPDS Synthetic}

The GPDS Synthetic Dataset is a large-scale signature dataset generated synthetically to simulate genuine and forged signatures. This dataset is part of the GPDS (Grupo de Procesado Digital de Señales) signature dataset family developed by the Universidad de Las Palmas de Gran Canaria \cite{gpds_dataset, Ferrer2015, Ferrer2013}.

Key features of the GPDS Synthetic dataset include:
\begin{itemize}
    \item A large number of subjects (4.000 subjects).
    \item Each subject includes 24 genuine signatures and 30 forged signatures.
    \item High variability to simulate different writing styles and conditions.
\end{itemize}

In the image below \ref{fig:GPDS_examples} shows some examples of the GPDS dataset:

\begin{figure}[H]
	\centering
	\includegraphics[width=11cm]{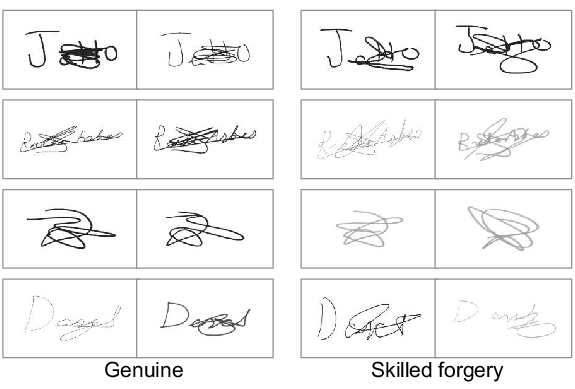}
	\caption{GPDS Signature Examples}
	Source: GPDS Dataset
    \label{fig:GPDS_examples}
\end{figure}

GPDS has some noticeably higher signature variability than CEDAR and ICDAR.

\section{Dataset Preparation and Analysis}

Studying the features of the dataset and promoting balanced batches during training are essential. The datasets used in this study are described in detail in this section, along with information on the number of samples, data splits, and pair generation strategy.

\subsection{Dataset Overview}

The key characteristics of the CEDAR, ICDAR, and GPDS datasets are compiled in Table \ref{tab:DatasetsAnalysis}. There are a set number of samples of both real and fake signatures in each dataset. The creation of strong training pairs and the model's ability to detect both real and fake signatures depend on this balanced distribution.

\begin{table}[H]
\centering
\caption{Datasets}
{%
\begin{tabular}{|l|c||c||c|}
\hline
\textbf{} & \textbf{CEDAR} & \textbf{ICDAR} & \textbf{GPDS} \\ \hline
Number of People &  55 & 69 & 4000\\ \hline
Genuine/person& 24 &24 & 24 \\ \hline
Forged/person& 24 & 24 & 30 \\ \hline
\end{tabular}%
}
\label{tab:DatasetsAnalysis}
\end{table}

\subsection{Generating the Pairs}

When designing the training and testing splits, certain strategic decisions were made to enhance the model’s ability to generalize to unseen data. These were identified empirically through experimentation and analysis of performance metrics across different split strategies:

\begin{enumerate}

    \item \textbf{Balance in terms of forged and original signatures:} Each split maintains a near-equal proportion of forged and genuine samples. This prevents the model from becoming biased toward one class and ensures more stable and reliable learning.

    \item \textbf{Balance in terms of writers:} Care was taken to ensure that the distribution of signature samples from different writers is homogeneous across the training, validation, and test sets. This reduces the risk of overfitting to the peculiarity of specific writers.

    \item \textbf{Pairing strategy to encode intra- and inter-writer comparisons:}
    \begin{itemize}
        \item \textbf{Genuine-Genuine from the same writer (label 0):}
        \begin{itemize}
            \item e.g., original\_idWriter1\_signature1 \& original\_idWriter1\_signature4 $\rightarrow$ 0
        \end{itemize}

        \item \textbf{Genuine-Forged from the same writer (label 1):}
        \begin{itemize}
            \item e.g., original\_idWriter1\_signature3 \& forged\_idWriter1\_signature6 $\rightarrow$ 1
        \end{itemize}
        
        \item \textbf{Genuine-Genuine from different writers (label 1):}
        \begin{itemize}
            \item e.g., original\_idWriter1\_signature1 \& original\_idWriter2\_signature4 $\rightarrow$ 1
        \end{itemize}
    \end{itemize}

    This approach enforces the model to not only distinguish between genuine and forged signatures, but also to learn the discriminative features that separate one writer from another. In particular, including genuine pairs from different writers (assigned label 1) teaches the model that signatures can be original and yet belong to distinct identities—a key insight for avoiding overfitting to dataset-specific or writer-specific patterns.

    Incorporating these types of inter-writer positive pairs proved essential to improving generalization across datasets. Traditional approaches often exclude such combinations, limiting the model’s capacity to capture between-author variability. The experiments showed that their inclusion increases robustness and performance, especially in cross-dataset evaluations.
\end{enumerate}

\subsubsection{Pair Formation Strategy}

\begin{figure}[H]
	\centering
	\includegraphics[width=15cm]{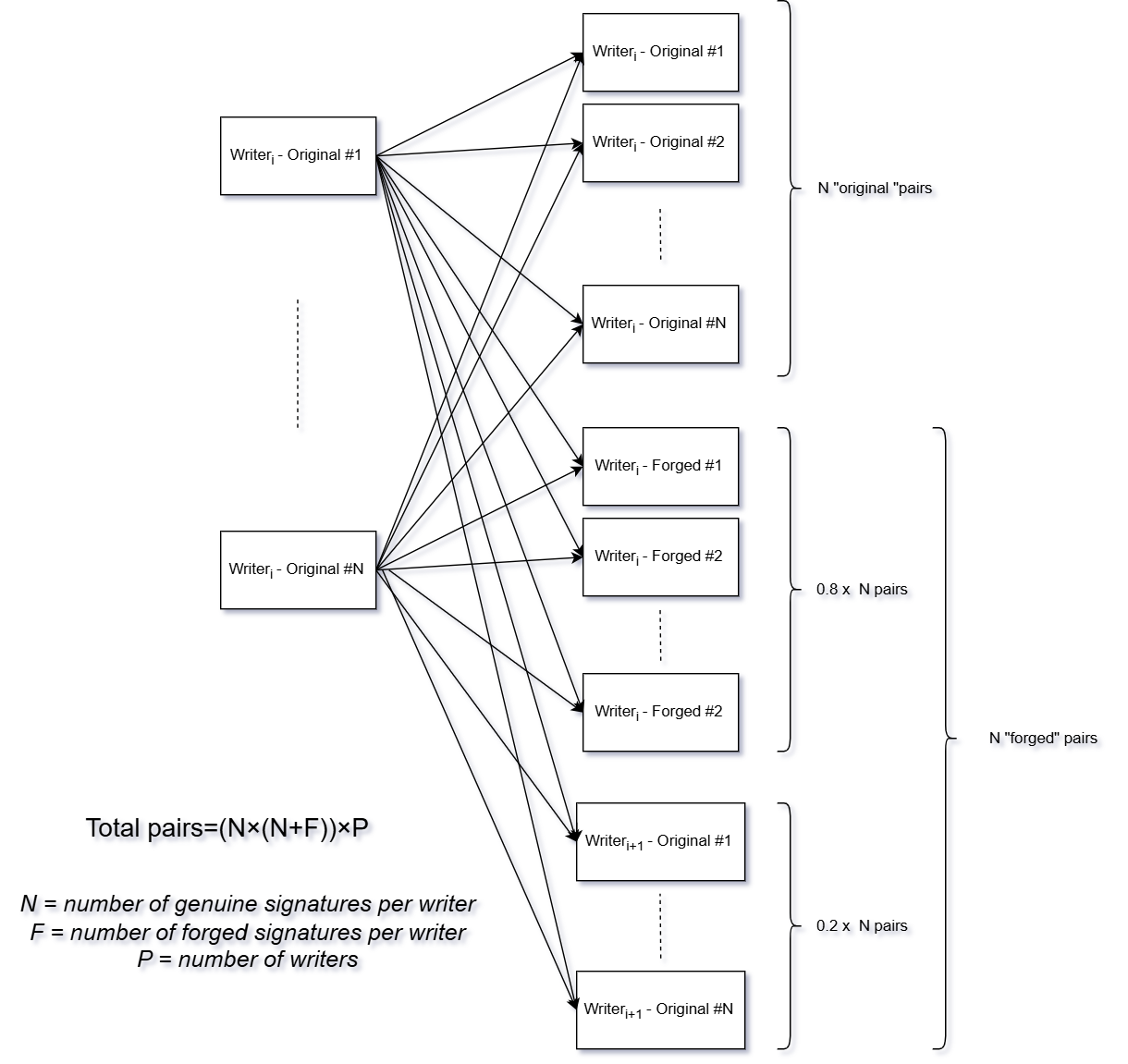}
	\caption{Diagram describing the formation of the samples (pairs of images).}
    \label{fig:pairs formation}
\end{figure}

Therefore, using this idea, we can diagrammatically illustrate the pair formation process in Figure~\ref{fig:pairs formation}.

Each dataset containing $N$ genuine signatures from a given writer is used to construct positive and negative pairs according to the following formulation:

\[
\mathcal{P} = \Big\{ (x_i, x_j, y) \;\big|\; 
y = \begin{cases} 
0, & \mathrm{writer}(x_i) = \mathrm{writer}(x_j)\text{ and (} x_i \text{ genuine,} x_j \text{ genuine)}, \\[3pt]
1, & (x_i \text{ genuine},\, x_j \text{ forged}) \;\text{ or }\; \mathrm{writer}(x_i) \neq \mathrm{writer}(x_j).
\end{cases}
\Big\}
\]

where $y$ denotes the label associated with the pair: $y=0$ for genuine pairs and $y=1$ for forged pairs.

\vspace{0.3cm}

\noindent\textbf{Observations:}  
As stated earlier, it was discovered that the model's generalization was improved by using signatures from multiple authors and labeling them as "forged" (since they essentially reflect multiple identities).  However, since the main objective of the system is to detect genuine skilled forgeries, only $20\%$ of the forged split was dedicated to this type of cross-writer pairing.  Since genuine-forged pairs are the most challenging situations in signature verification, the remaining $80\%$ section was dedicated to them.

Additionally, pairs of genuine signatures $(x_i, x_j)$ and $(x_j, x_i)$ were considered as distinct samples. The main reason of this choice is that it allows the utilization of all available forged data while preserving a balanced split among classes.  Also, these reversed pairs are not just duplicates; instead, they provide the network slightly different inputs because each sample is subjected to independent data augmentation.  This decision makes the splits easier to construct, and it did not lead to significant issues, as the raw image results were already quite satisfactory (as will be demonstrated later).

\subsection{ICDAR Dataset Details}

\begin{figure}[H]
	\centering
	\includegraphics[width=\linewidth]{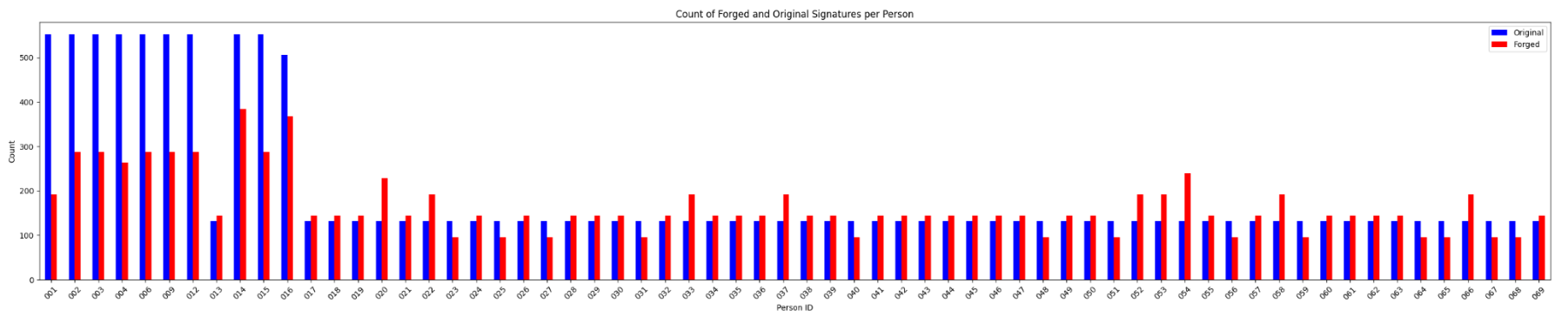}
	\caption{ICDAR Dataset Pair distribution}
    \label{fig:ICDAR_split_image}
\end{figure}

The ICDAR dataset theoretically contains 69 individuals. However, IDs 05, 07, 08, 10, and 11 are missing from the original dataset. Also, the original dataset presented some imbalance among the first writers. Both of these details are illustrated in Figure~\ref{fig:ICDAR_split_image}.

\begin{figure}[H]
	\centering
	\includegraphics[width=\linewidth]{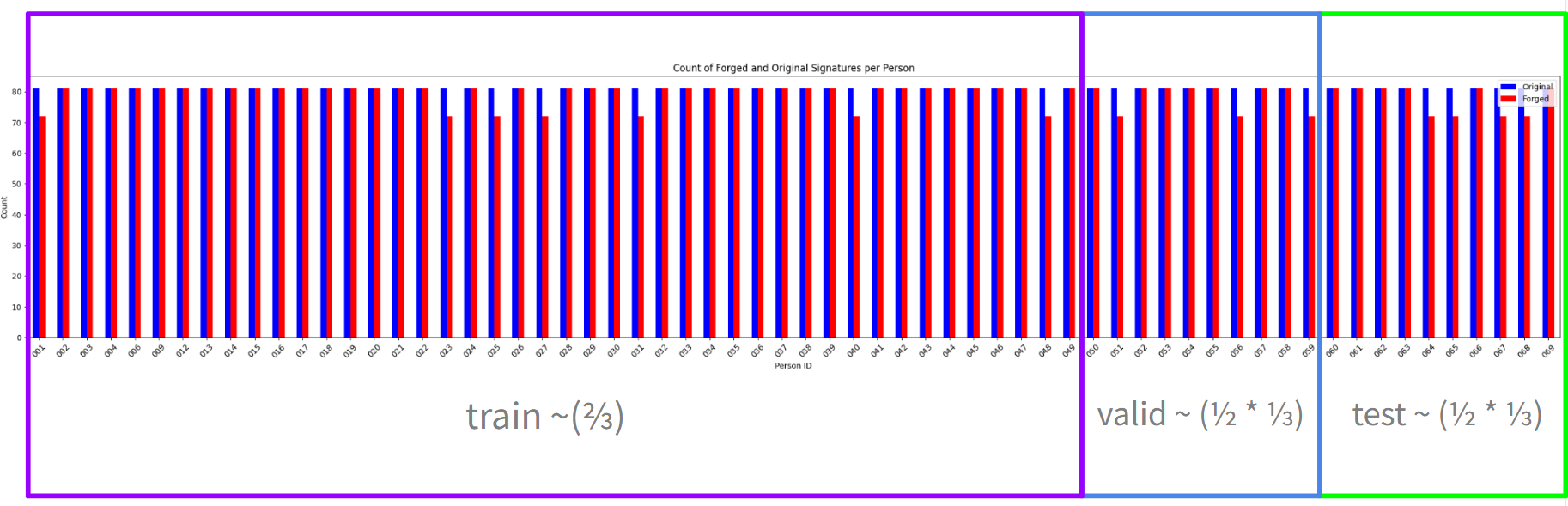}
	\caption{ICDAR Dataset Pair distribution after downsampling}
    \label{fig:ICDAR_split_image_down}
\end{figure}

To correct the issue of data imbalance across writers and ensure a uniform distribution between them, a downsampling procedure was applied, resulting in the balanced dataset shown in Figure~\ref{fig:ICDAR_split_image_down}.

Each individual contributes approximately 81 genuine and 81 forged signature pairs (9 original and 9 forged signature images per writer). Table \ref{tab:ICDARDatasetDetails} outlines the data splits and pair generation, showing that the training set contains samples from 49 individuals, while validation and testing each involve 10 individuals. The approximate nature of the pair counts is noted due to variations in available signatures.

\begin{table}[H]
\centering
\caption{ICDAR Split file Details}
\resizebox{\textwidth}{!}{%
\begin{tabular}{|l|c|c|c|c|}
\hline
\textbf{ICDAR}       & \textbf{Train} & \textbf{Valid} & \textbf{Test} & \textbf{Total} \\ \hline
Number of People     & 44 (01–49) *    & 10 (50–59)     & 10 (60–69)    & 69             \\ \hline
PAIRS Genuine/person & $\sim$81       & $\sim$81       & $\sim$81      &                \\ \hline
PAIRS Forged/person  & $\sim$81       & $\sim$81       & $\sim$81      &                \\ \hline
Total                & 7065 (7128)    & 1593 (1620)    & 1584 (1620)   &                \\ \hline
\end{tabular}%
}
\label{tab:ICDARDatasetDetails}
\end{table}

\textbf{* Observation}: As IDs 05, 07, 08, 10, and 11 are missing from the original dataset, only 44 individuals are present within the range of IDs 01 to 49

\subsection{CEDAR Dataset Details}

\begin{figure}[H]
	\centering
	\includegraphics[width=\linewidth]{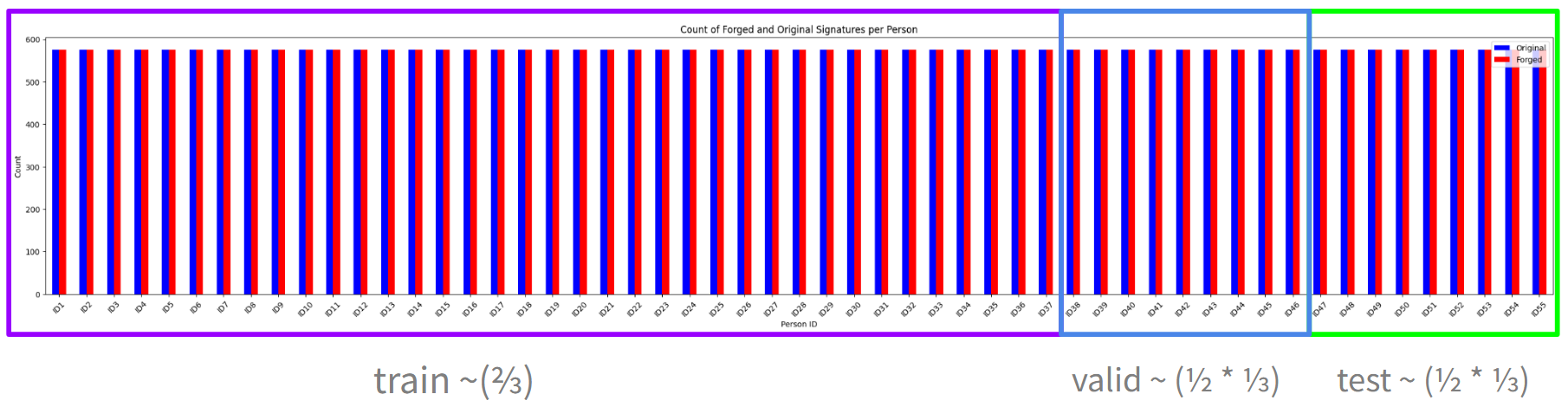}
	\caption{CEDAR Dataset Pair distribution}
    \label{fig:CEDAR_split_image}
\end{figure}

The CEDAR dataset consists of 55 individuals, each providing 24 genuine and 24 forged signatures. Figure~\ref{fig:CEDAR_split_image} provides a visual illustration of the pair formation and the splitting strategy adopted to ensure a carefully balanced dataset.

To preserve balance across the datasets, a downsampling procedure was applied, ensuring that the distribution of pairs remains consistent among the training, validation, and testing sets between ICDAR and CEDAR.

Table \ref{tab:CEDARDatasetDetails} details the data splits for training, validation, and testing. The training set includes signatures from 37 individuals, while the remaining samples are divided equally for validation and testing. For each individual, 105 pairs of genuine and forged signatures are generated, ensuring balanced training and evaluation sets.

\begin{table}[H]
\centering
\caption{CEDAR Split file Details}
\resizebox{\textwidth}{!}{%
\begin{tabular}{|l|c|c|c|c|}
\hline
\textbf{CEDAR}       & \textbf{Train} & \textbf{Valid} & \textbf{Test} & \textbf{Total} \\ \hline
Number of People     & 37 (01–37)     & 09 (38–46)     & 09 (47–55)    & 55             \\ \hline
PAIRS Genuine/person & 105            & 105            & 105           &                \\ \hline
PAIRS Forged/person  & 105            & 105            & 105           &                \\ \hline
Total                & 7770           & 1890           & 1890          &                \\ \hline
\end{tabular}%
}
\label{tab:CEDARDatasetDetails}
\end{table}

\subsection{GPDS Dataset Details}

The GPDS dataset is significantly larger, with 4000 individuals contributing to the dataset. Each individual provides 100 pairs of genuine and forged signatures. 

Again, to maintain consistency and balance with the other datasets, a subset of 58 random individuals was selected to form the GPDS split file.

Table \ref{tab:GPDSDatasetDetails} details the data splits, which include 40 individuals for training and 9 individuals each for validation and testing.

\begin{table}[H]
\centering
\caption{GPDS Split file Details}
\resizebox{\textwidth}{!}{%
\begin{tabular}{|l|c|c|c|c|}
\hline
\textbf{GPDS}        & \textbf{Train}  & \textbf{Valid}  & \textbf{Test}   & \textbf{Total} \\ \hline
Number of People     & 40 (random)     & 09 (random)     & 09 (random)     & 4000           \\ \hline
PAIRS Genuine/person & 100             & 100             & 100             &                \\ \hline
PAIRS Forged/person  & 100             & 100             & 100             &                \\ \hline
Total                & 8000            & 1800            & 1800            & 11600          \\ \hline
\end{tabular}%
}
\label{tab:GPDSDatasetDetails}
\end{table}

\section{Model Architecture and Training Details}

For this work, two models were developed: one Siamese with ResNet34 backbone trained with Constrastive Loss, and other Siamese with ResNet34 backbone trained with Triplet Loss.

Both models were trained with the following parameter details:

\begin{table}[H]
\centering
\caption{Training Details}
{%
\begin{tabular}{|l|c|}
\hline
\textbf{Details} & \textbf{Value} \\ \hline
Optimizer        & SGD \\ \hline
Learning Rate    & 0.001 \\ \hline
Batch Size       & 32 \\ \hline
Epochs           & 20 \\ \hline
Loss Function    & Contrastive/Triplet Loss (margin = 1) \\ \hline
Dataset Split    & 2/3 Train - 1/6 Validation - 1/6 Test \\ \hline
Early Stopping    & Validation Loss \\ \hline
\end{tabular}%
}
\label{tab:trainingDetails}
\end{table}

These hyperparameters were chosen based on a series of empirical tests detailed on the appendix session.

And regarding data augmentation, the models were trained with the data augmentation structures presented in table 3.6:

\begin{table}[H]
\centering
\caption{Image Transformations for Training and Validation}
\resizebox{\textwidth}{!}{%
\begin{tabular}{|l|l|c|}
\hline
\textbf{Transformation}            & \textbf{Description}                                                              & \textbf{Training / Validation} \\ \hline
Resize                             & Resizes images to 512x512 pixels                                                  & Both                           \\ \hline
HorizontalFlip                     & Random horizontal flip with probability 0.5                                       & Training                       \\ \hline
Affine                             & \begin{tabular}[c]{@{}l@{}}
    Translates up to 5\%, rotates with elliptical method, \\
    shears between -2° and 2°, scales from 70\% to 130\%, \\
    and maintains aspect ratio (p=0.5)
    \end{tabular}                                                       & Training                       \\ \hline
Sharpen                            & Randomly sharpens the image with probability 0.5                                  & Training                       \\ \hline
RandomBrightnessContrast           & Adjusts brightness and contrast randomly within limits (±0.01, p=0.5)             & Training                       \\ \hline
Normalize                          & Normalizes the image to mean 0 and standard deviation 1                           & Both                           \\ \hline
ToTensorV2                         & Converts the image to PyTorch tensor format                                       & Both                           \\ \hline
\end{tabular}%
}
\label{tab:ImageTransformations}
\end{table}

\subsection{Siamese with Contrastive Loss}

\begin{figure}[H]
	\centering
	\includegraphics[width=15cm]{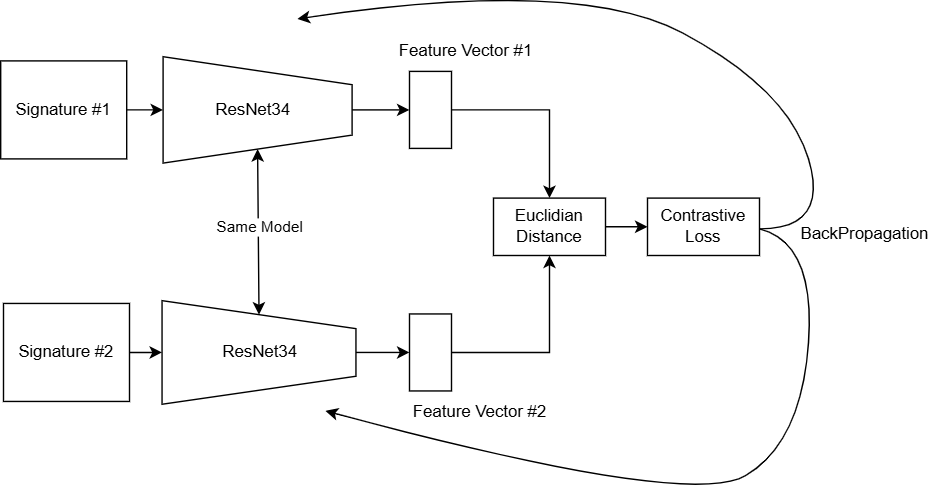}
	\caption{High-level Siamese Network architecture used in Training}
	Source: Author
    \label{fig:finalArchitectureContrastiveTrain}
\end{figure}

The architecture developed for the task includes training a Siamese Network (ResNet34) with Contrastive Loss with margin 1, detailed in picture \ref{fig:finalArchitectureContrastiveTrain}

In this architecture, two input signatures are processed by the same ResNet34 backbone with shared weights, producing feature vectors in a common embedding space. The Euclidean distance between the embeddings is then computed and passed to the Contrastive Loss function, which guides the backpropagation process. This setup enforces that genuine pairs are mapped closer together, while forged pairs are pushed farther apart in the embedding space.

The model was trained using a feature learning loss function with a distance margin of 1 to create the embeddings. During classification, a threshold of 0.5 is applied to the Euclidean distance between embeddings: distances less than 0.5 are classified as a true (genuine) signature, while distances greater than 0.5 are classified as a false (forged) signature.

And for validation and test, the following paradigm was used:

\begin{figure}[H]
	\centering
	\includegraphics[width=15cm]{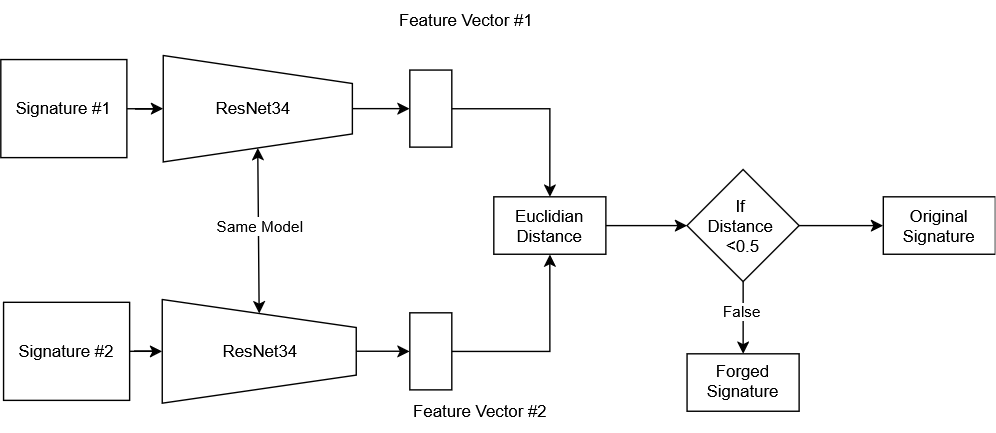}
	\caption{High-level Siamese Network architecture used in Validation and Test}
	Source: Author
    \label{fig:finalArchitectureContrastiveValid}
\end{figure}


\subsection{Siamese with Triplet Loss}

\begin{figure}[H]
	\centering
	\includegraphics[width=15cm]{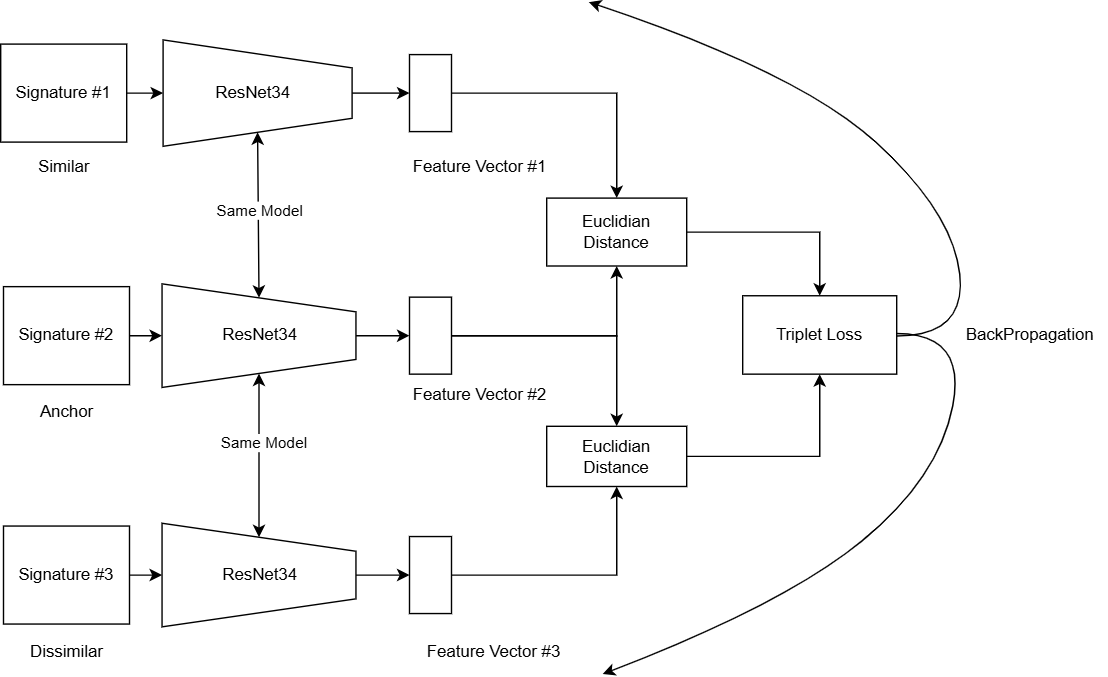}
	\caption{High-level Siamese Network architecture with Triplet Loss used in training}
	Source: Author
    \label{fig:finalArchitectureTripletTrain}
\end{figure}

The architecture developed for the task includes training a Siamese Network (ResNet34) with Triplet Loss with margin 1, detailed in picture \ref{fig:finalArchitectureTripletTrain}

The Triplet Loss function receives three inputs: an anchor sample $x_a$, a positive sample $x_p$ (from the same class as the anchor), and a negative sample $x_n$ (from a different class). All three samples are mapped to a shared embedding space through the same backbone network (ResNet34). The goal is to minimize the distance between the anchor and the positive, while maximizing the distance between the anchor and the negative, enforcing a margin $m$ between them:

\[
L = \max\left(0, \; d(f(x_a), f(x_p)) - d(f(x_a), f(x_n)) + m \right),
\]

where $f(\cdot)$ denotes the embedding function learned by the network, $d(\cdot,\cdot)$ is the Euclidean distance, and $m = 1$ is the margin applied in the experiments. In practice, this ensures that the embeddings of genuine signatures remain close to each other, while forged signatures are pushed farther away from the anchor in the embedding space.

For validation and test, the pipeline is exactly the same as used in the Contrastive Loss, demonstrated in \ref{fig:finalArchitectureContrastiveValid}

\section{Results and Discussion}

Combinations of distinct sections from different datasets (CEDAR, ICDAR, GPDS) are represented by the dataset groups (CI, CG, IG, and CIG). In particular, the training, validation, and testing splits from each dataset are combined to create each group. It is defined in this approach:

\begin{itemize}
    \item \textbf{CI (CEDAR + ICDAR)} is formed by combining \textit{trainingCEDAR} with \textit{trainingICDAR}, \textit{validationCEDAR} with \textit{validationICDAR}, and \textit{testingCEDAR} with \textit{testingICDAR}.
    \item \textbf{CG (CEDAR + GPDS)} combines \textit{trainingCEDAR} and \textit{trainingGPDS}, \textit{validationCEDAR} and \textit{validationGPDS}, and similarly for the testing portions.
    \item \textbf{IG (ICDAR + GPDS)} follows the same approach, merging the respective splits of \textit{ICDAR} and \textit{GPDS}.
    \item \textbf{CIG (CEDAR + ICDAR + GPDS)} is the union of all splits from \textit{CEDAR}, \textit{ICDAR}, and \textit{GPDS}, combining \textit{trainingCEDAR + trainingICDAR + trainingGPDS}, and similarly for validation and testing.
\end{itemize}

This configuration aims to evaluate the model's \textbf{cross-dataset testing} capabilities and determine whether it can accurately classify signatures from datasets that were not used for training.

The results exposed in the following tables represent the AUC (Area Under the Curve) of the ROC graph.

\subsection{Contrastive Loss Results}

\begin{table}[H]
\centering
\caption{Model Performance Across Datasets using Contrastive Loss (AUC values)}
\resizebox{\textwidth}{!}{%
\begin{tabular}{|l|l|c|c|c|c|}
\hline
\textbf{Group / Training} & \textbf{Model} & \multicolumn{4}{c|}{\textbf{Dataset Test}} \\ \hline
                         &                & \textbf{CEDAR} & \textbf{ICDAR} & \textbf{GPDS} & \textbf{CIG} \\ \hline
1 (CI) & model\_2025-01-05\_16:57:36 & 1.0            & 0.99           & \textbf{0.53} & 0.84          \\ \hline
2 (CG) & model\_2025-01-05\_20:51:16 & 1.0            & \textbf{0.83} & 0.81          & 0.92          \\ \hline
3 (IG) & model\_2025-01-05\_22:22:18 & \textbf{0.76} & 0.99           & 0.75          & 0.85          \\ \hline
4 (CIG) & model\_2025-01-06\_18:51:03 & 1.0            & 0.99           & 0.73          & 0.96          \\ \hline
\end{tabular}%
}
\label{tab:ModelPerformance}
\end{table}

The results indicate that the proposed structure can generalize enough to categorize signatures from different datasets. The values that are bolded indicate the results of cross-testing, in which the model was evaluated using datasets that were not used for training.

All models outperformed the stochastic baseline, indicating their robustness. In particular, the model trained with the CI group did the worst on the GPDS dataset. This is most likely due to the GPDS dataset's complexity and variety of synthetic samples.

\subsection{Triplet Loss Results}

\begin{table}[H]
\centering
\caption{Model Performance Across Datasets using Triplet Loss (AUC values)}
\resizebox{\textwidth}{!}{%
\begin{tabular}{|l|l|c|c|c|c|}
\hline
\textbf{Group / Training} & \textbf{Model} & \multicolumn{4}{c|}{\textbf{Dataset Test}} \\ \hline
                          &                & \textbf{CEDAR} & \textbf{ICDAR} & \textbf{GPDS} & \textbf{CIG} \\ \hline
1 (CI) & model\_2025-01-23\_20:53:07 & 1.0            & 0.86           & \textbf{0.67} & 0.78          \\ \hline
2 (CG) & model\_2025-01-23\_21:27:33 & 1.0            & \textbf{0.83} & 0.71          & 0.82          \\ \hline
3 (IG) & model\_2025-01-23\_21:55:16 & \textbf{0.95} & 0.85           & 0.79          & 0.85          \\ \hline
4 (CIG) & model\_2025-01-24\_14:58:08 & 0.98           & 0.90           & 0.61          & 0.81          \\ \hline
\end{tabular}%
}
\label{tab:ModelPerformance}
\end{table}

Training a Siamese Network with Triplet Loss is more challenging compared to Contrastive Loss. This problem occurs because the anchor, positive, and negative sample selection during training has a significant impact on the model's performance. As a result, in comparison to Contrastive Loss, it is noticeable a lower overall performance metrics.

However, Triplet Loss demonstrates a significant advantage in its ability to generalize across datasets, as seen in the cross-test results. The models trained with Triplet Loss exhibit superior performance when evaluated on unseen datasets, achieving better results compared to those trained with Contrastive Loss. This indicates that while Triplet Loss is harder to train, it offers enhanced generalization capabilities, making it a valuable approach for tasks involving diverse data distributions.

Another notable observation during training with Triplet Loss was that high numbers of epochs were not required to achieve competitive performance. This is because, in each backpropagation step, the model processes three samples (anchor, positive, and negative) instead of two as in Contrastive Loss. While this increases the computational time per iteration, it reduces the number of epochs needed to reach similar or superior performance, resulting in more efficient training overall.

In the table below it is clear that, in terms of generalization between datasets (cross-dataset test), Triplet outperforms the Contrastive loss:

\begin{table}[H]
\centering
\caption{Comparative AUC of Models using Contrastive Loss and Triplet Loss}
\resizebox{\textwidth}{!}{%
\begin{tabular}{|l|c|c|c|c|c|c|c|c|}
\hline
\multirow{2}{*}{\textbf{Dataset}} & \multicolumn{2}{c|}{\textbf{CI}} & \multicolumn{2}{c|}{\textbf{CG}} & \multicolumn{2}{c|}{\textbf{IG}} & \multicolumn{2}{c|}{\textbf{CIG}} \\ \cline{2-9}
 & \textbf{Contrastive} & \textbf{Triplet} & \textbf{Contrastive} & \textbf{Triplet} & \textbf{Contrastive} & \textbf{Triplet} & \textbf{Contrastive} & \textbf{Triplet} \\ \hline
CEDAR & 1.00 & 1.00 & 1.00 & 1.00 & 0.76 & \textbf{0.95} & 1.00 & 0.98 \\ \hline
ICDAR & 0.99 & 0.86 & \textbf{0.83} & \textbf{0.83} & 0.99 & 0.85 & 0.99 & 0.90 \\ \hline
GPDS & 0.53 & \textbf{0.67} & 0.81 & 0.71 & 0.75 & 0.79 & 0.73 & 0.61 \\ \hline
CIG & 0.84 & 0.78 & 0.92 & 0.82 & 0.85 & 0.85 & 0.96 & 0.81 \\ \hline
\end{tabular}%
}
\label{tab:comparison_loss}
\end{table}

  \chapter{Part II - Designing a Shell-Based Pre-processing Pipeline - Development}

In this chapter, it is described the development of a shell-based pre-processing pipeline to be used as an input to train the model. The primary objective of this part of the study is to deterministically reduce the dimensionality of the input data, while trying to lose the less information as possible, and to study the model performance.

\section{Image Pre-processing}

In the image below \ref{fig:cascas_processamento} shows the high-level diagram for the pre-processing pipeline:

\begin{figure}[H]
	\centering
	\includegraphics[width=13cm]{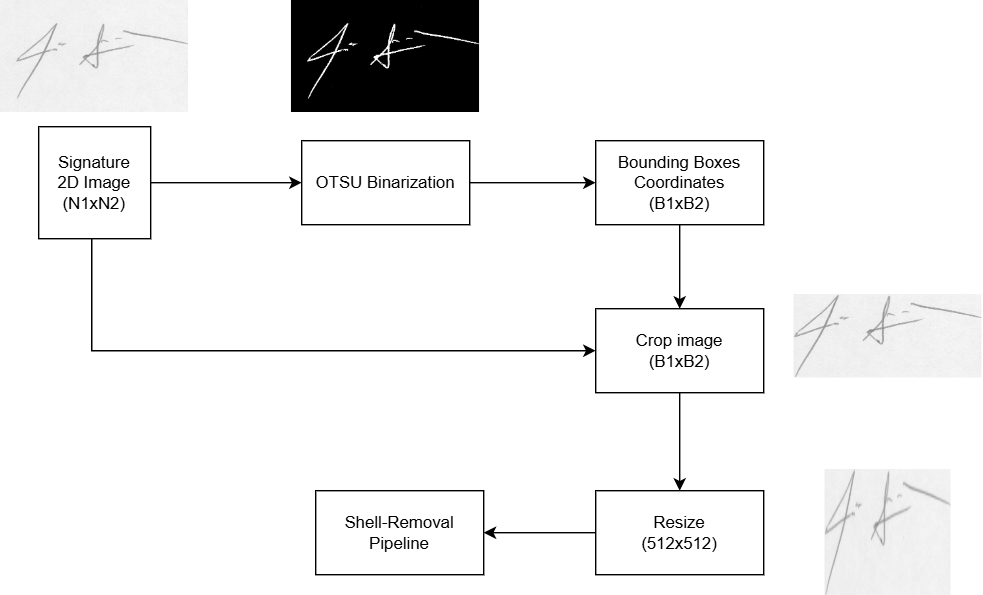}
	\caption{Pre-processing pipeline}
    \label{fig:cascas_processamento}
\end{figure}

\subsection{Input and OTSU Binarization}

The pipeline starts with a raw grayscale signature image, denoted as a 2D array of dimensions \(N_1 \times N_2\). The first operation applied is OTSU's binarization, a global thresholding method that automatically selects an optimal threshold to separate foreground (signature strokes) from the background. This results in a binary image where pixels are either 0 (background) or 1 (ink trace), effectively isolating the signature structure.

\subsection{Bounding Box Extraction}

Once binarized, the coordinates of the bounding box surrounding the signature strokes are computed. This step identifies the minimum and maximum row and column indices that contain ink, producing a bounding box of shape \(B_1 \times B_2\). This bounding box localizes the meaningful region of the image, excluding unnecessary blank space.

\subsection{Cropping}

With the bounding box coordinates obtained, the original input image (prior to binarization) is then cropped accordingly. This ensures that the original gray-level information of the ink is kept but reduce the spatial dimensions by focusing only on the active region. The result is a cropped image of shape \(B_1 \times B_2\).

\subsection{Resize Operation}

After cropping, the image is resized to a fixed resolution of \(512 \times 512\). This standardization is essential to ensure that all input images have the same size, regardless of their original dimensions, which allows for consistent batch processing and model input.

\section{Shell-Removal Pipeline}

\subsection{Definition of Shells}

Shells are defined as successive horizontal contours external from the central ink regions of the signature. Each shell is constructed from a binary signature image by separating and isolating the upper and lower boundaries of the ink trace along each column. These are referred to as the \textit{superior} and \textit{inferior} shells, respectively.

The extraction is performed using a custom morphological scan through the binary image that identifies the first (top) and last (bottom) foreground pixels in each column. This results in one-dimensional curves:

\[
\text{Shell}_\text{superior}(j) = \max \{ i \mid \text{shell}[i, j] = 1 \}
\]
\[
\text{Shell}_\text{inferior}(j) = \min \{ i \mid \text{shell}[i, j] = 1 \}
\]

for each column \( j \) in the image. Variants of these shells are also obtained through specific morphological operations (e.g., removing overlaps or combining them into residual shells).

\subsection{Shell Extraction Pseudo-Algorithm}

The shell extraction procedure receives a preprocessed binary image (cropped and resized to $512 \times 512$) and outputs the superior, inferior, and residual shell representations. The steps are summarized as follows:

\begin{algorithm}
\caption{Shell Extraction Procedure}
\begin{algorithmic}[1]
\State \textbf{Input:} Binary image $img \in \{0,1\}^{512 \times 512}$
\State \textbf{Output:} Shell matrices: $shellS1$, $shellI1$, $shellS2$, $shellI2$, $resS$, $resI$
\Statex
\State \textbf{(1)} Apply morphological pruning:
\State \hspace{1em} $skel, mask \gets pruning(img)$
\Statex
\State \textbf{(2)} Extract primary shells:
\State \hspace{1em} $sup\_bin \gets shellS\_binary(mask)$
\State \hspace{1em} $inf\_bin \gets shellI\_binary(mask)$
\Statex
\State \textbf{(3)} Extract residual shell:
\State \hspace{1em} $res\_bin \gets res\_binarized(mask - (sup\_bin \lor inf\_bin))$
\Statex
\State \textbf{(4)} Map shells to 1D functional form:
\State \hspace{1em} $(shellS1, shellI1) \gets img\_to\_shell\_func(sup\_bin)$
\State \hspace{1em} $(shellS2, shellI2) \gets img\_to\_shell\_func(inf\_bin)$
\State \hspace{1em} $(resS, resI) \gets img\_to\_shell\_func(res\_bin)$
\Statex
\State \Return All six shell vectors.
\end{algorithmic}
\end{algorithm}

The pseudo-algorithms for \textit{pruning(img)}, \textit{shellS\_binary(img)}, \textit{shellI\_binary(img)}, \textit{res\_binarized(img)}, and \textit{img\_to\_shell\_func(img)} are described on the appendix A.

The algorithm can be intuitively visualized as a scanning arrow that moves from left to right, progressively capturing the coordinates of the outermost (superior) representation of the signature. At each column, once a discontinuity is detected, the arrow is reset downwards to the bottom of the image (row index 0), resuming the search for the next visible stroke.

\begin{figure}[H]
\centering
\includegraphics[width=\linewidth]{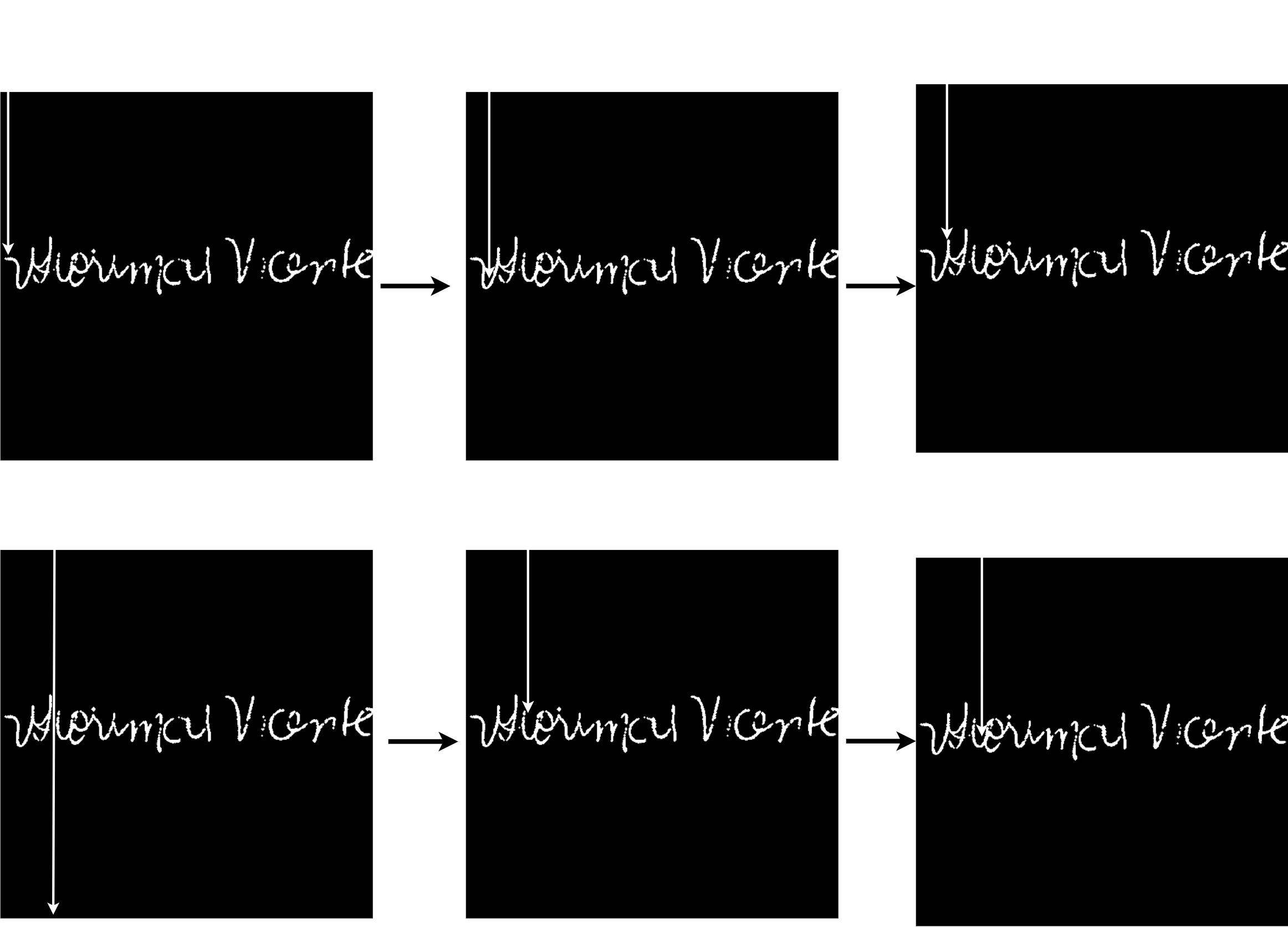}
\caption{Illustrative pipeline of the superior shell extraction. The arrow traverses each column from left to right, capturing the upper boundary of the signature. When a discontinuity is encountered, the arrow resets to the bottom of the image and continues scanning.}
\label{fig:pipeline}
\end{figure}

Each shell is a 1D vector of length 512, representing the row index (vertical position) of the detected contour along each image column. The outputs can be normalized, padded, or used to compute pressure and thickness values from the original grayscale image.

\newpage

\subsection{Visualizing the Pipeline's outputs}

\noindent To better illustrate this process, an example is provided below. First, the original signature image is shown:

\begin{figure}[H]
	\centering
	\includegraphics[width=8cm]{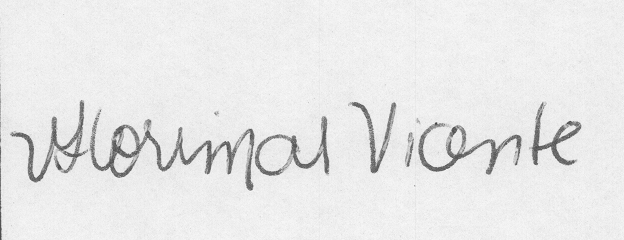}
	\caption{Original Signature Image}
    \label{fig:part1_cascas}
\end{figure}

Following pre-processing, the image is binarized, cropped, and normalized before going through the shell-removal pipeline. Consequently, the mask is the outcome of this step:

\begin{figure}[H]
	\centering
	\includegraphics[width=8cm]{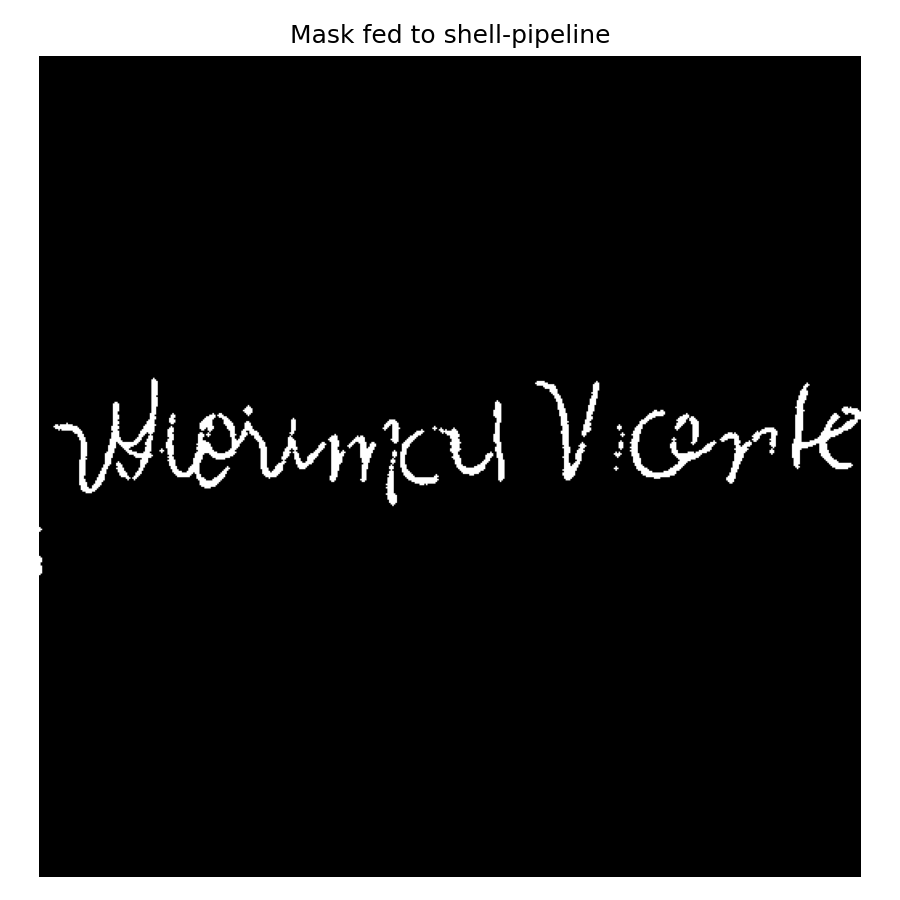}
	\caption{Mask used as the input to the shell-removal pipeline}
    \label{fig:part2_cascas}
\end{figure}

Note that the mask has a fixed resolution of 512 $\times$ 512 pixels (has a square shape).

\noindent Next, the six shells extracted by the pipeline (superior and inferior contours, along with residuals) are presented:

\begin{figure}[H]
	\centering
	\includegraphics[width=14cm]{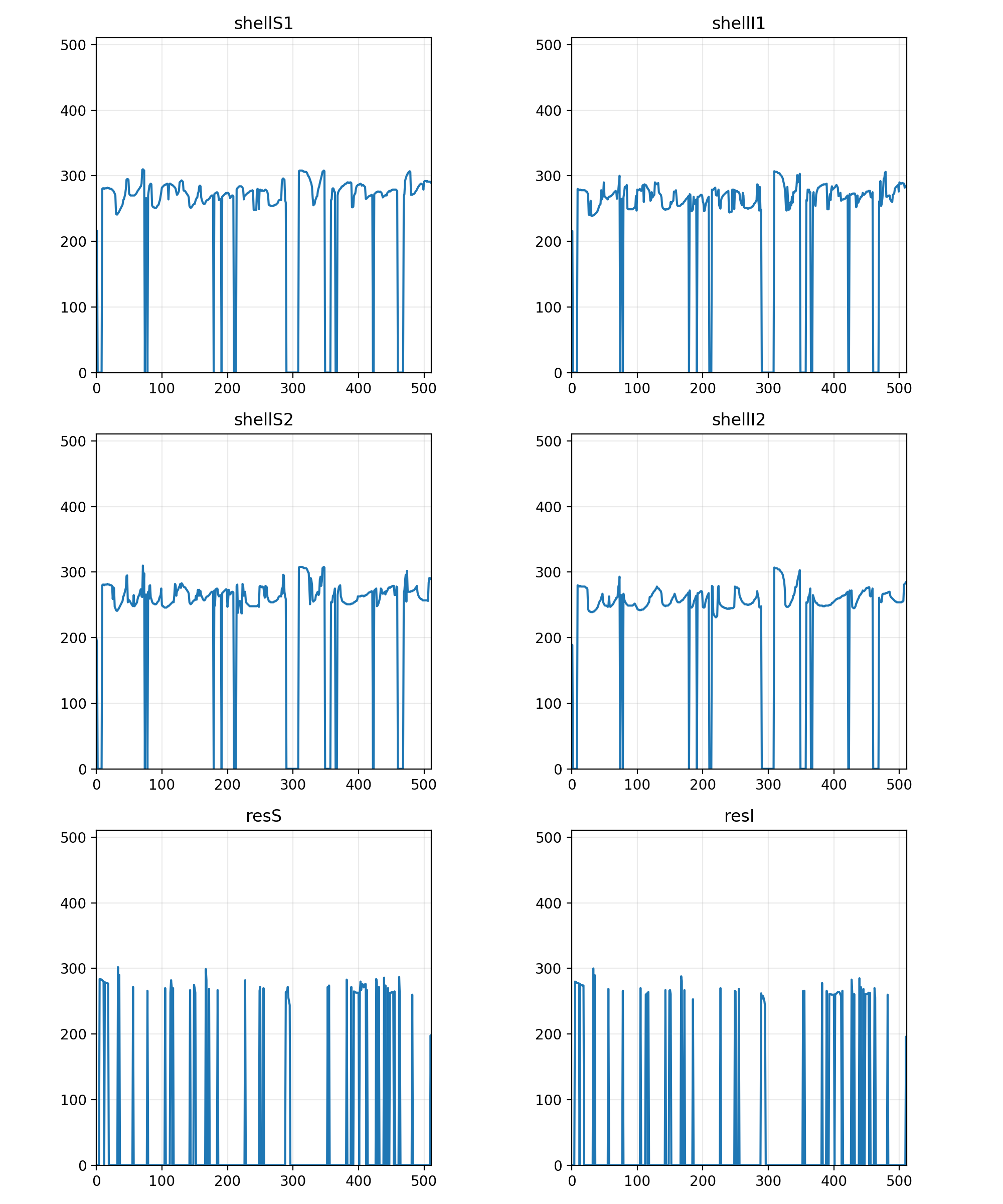}
	\caption{All six shells generated from the mask.}
    \label{fig:part3_cascas}
\end{figure}

\noindent Notably, the significant horizontal lines shown in the shell vectors represent discontinuity places in the signature, which are areas where the signer stops writing and then continues.

Also note that, since the residual shells are obtained as the difference between the first and second shells, they capture only fine details, often generating multiple discontinuities, and may not necessarily provide relevant information.

To facilitate interpretation, overlays between superior and inferior shells with the original mask are shown below:

\begin{figure}[H]
	\centering
	\includegraphics[width=8cm]{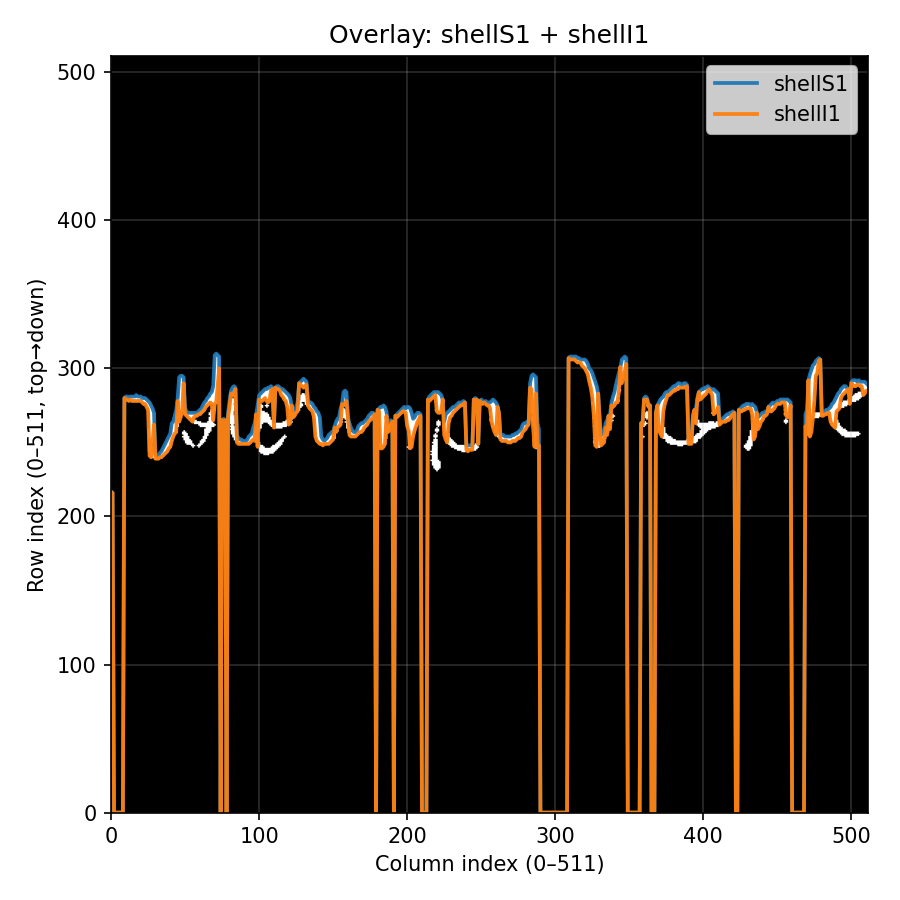}
	\caption{Overlay of the first superior shell (S1) with the first inferior shell (I1) on top of the mask.}
    \label{fig:part4_cascas}
\end{figure}

\begin{figure}[H]
	\centering
	\includegraphics[width=8cm]{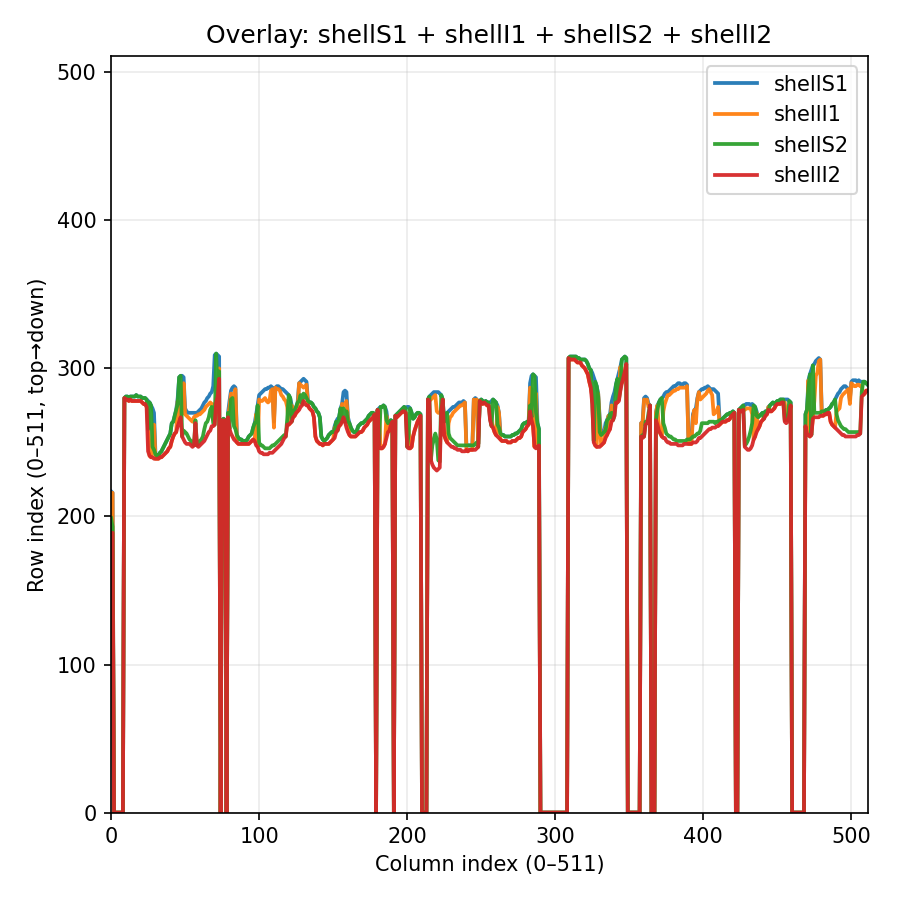}
	\caption{Overlay of the second superior shell (S2) with the second inferior shell (I2), displayed without the mask to highlight the reconstruction.}
    \label{fig:part5_cascas}
\end{figure}

\begin{figure}[H]
	\centering
	\includegraphics[width=8cm]{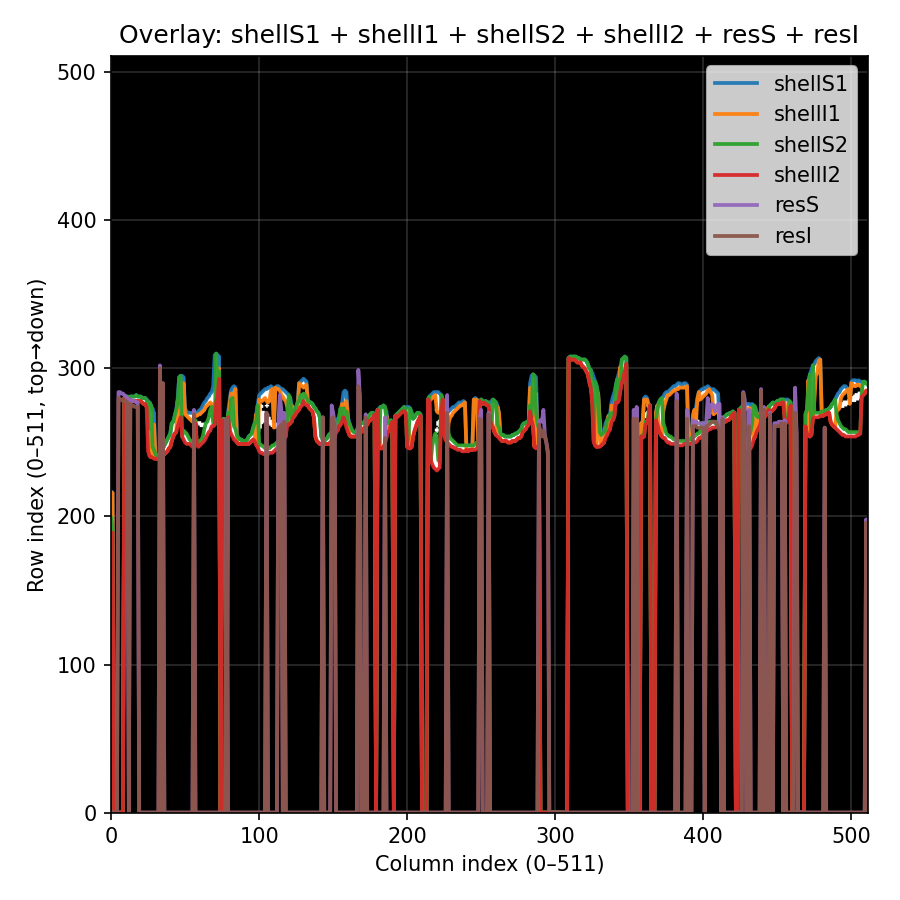}
	\caption{Overlay of all superior and inferior shells, including the residual shells, on top of the mask.}
    \label{fig:part6_cascas}
\end{figure}

\subsection{Functional Representation}

Each shell is flattened into a one-dimensional vector by directly using the column-wise height values (row indices) of the superior and inferior contours. These vectors represent the structural profile of the signature across the horizontal axis.

All resulting shell vectors are stored as rows in a CSV, forming a matrix where:

\begin{itemize}
    \item Each row corresponds to one shell (e.g., superior shell 1, inferior shell 1, residuals, etc.)
    \item Each column corresponds to a column position in the image (fixed width of 512 pixels)
\end{itemize}

\section{Auxiliary Feature Extraction}

\subsection{Pressure Estimation}

As offline signature images lack real pen pressure data, an approximation is computed based on grayscale intensity in the neighborhood of each shell point. For each pixel along a shell, a vertical sampling window is applied, and grayscale values are extracted from the original gray-level image. This simulates a pseudo-pressure measure based on ink darkness and thickness.

The pseudo-pressure matrix has shape \(11 \times 512\) and is computed as:

\[
P(i, j) = \text{gray\_level}(y + i - 5, j), \quad i \in [0, 10]
\]

where \( y \) is the row index of the shell pixel and \( j \) is the column index. These values are saved as CSV files, with one row per offset from the central shell (total of 11 rows).

\subsection{Thickness Calculation}

The thickness is computed using two custom binary scans:

\begin{itemize}
    \item The \textbf{superior thickness} is defined as the number of continuous foreground pixels starting from the topmost ink pixel in a column and going downward.
    \item The \textbf{inferior thickness} is defined similarly, but starting from the bottommost ink pixel and going upward.
\end{itemize}

These are stored as two 1D arrays (length 512), where each entry represents the estimated thickness at a column.

Although this provides a valid representation of local stroke width, in practice the shell coordinates themselves already encode thickness information: the difference between the superior and inferior shells at each column directly reflects the vertical span of the ink. For this reason, the explicit thickness vectors were not used in this work.

\section{Output and Export}

Once all features are computed, they are saved into structured Excel or CSV files for later use. The output consists of:

\begin{itemize}
    \item \textbf{Shells:} Matrix with 6 rows (superior/inferior of two layers + residuals), each of length 512.
    \item \textbf{Pressure:} Matrix with shape \(6 \times 11 \times 512\) stored as flattened rows (or multiple sheets).
    \item \textbf{Thickness:} Two 1D arrays, one for the difference between shells superior and inferior n°1 and one for the difference between shells superior and inferior n°2.
\end{itemize}

In addition, the skeletonized and processed signature images are saved as PNG files for visual inspection and validation.

\section{Model Architecture and Analysis}

The first step was to design a simpler architecture that could handle this structure, as shell-based data representation is unconventional and does not follow a standardized format like 2D images or common sequential signals. The idea was to construct a Siamese embedding network capable of handling multiple shell signals simultaneously, while also dedicating a specific branch for the pressure vectors. This structure was selected to enable the extraction of discriminative features from both the shape-based and dynamic aspects of the signature.

After several experiments, the proposed \textit{PS Layer} (PressureShell Layer) was introduced as the initial component of the network, specifically designed to process the unconventional shell-based representation of signatures. Its purpose is to integrate the multiple shell signals with the dynamic features of pressure before passing them to deeper layers, ensuring that both aspects of the data are preserved from the beginning of the pipeline.

\begin{figure}[H]
	\centering
	\includegraphics[width=15cm]{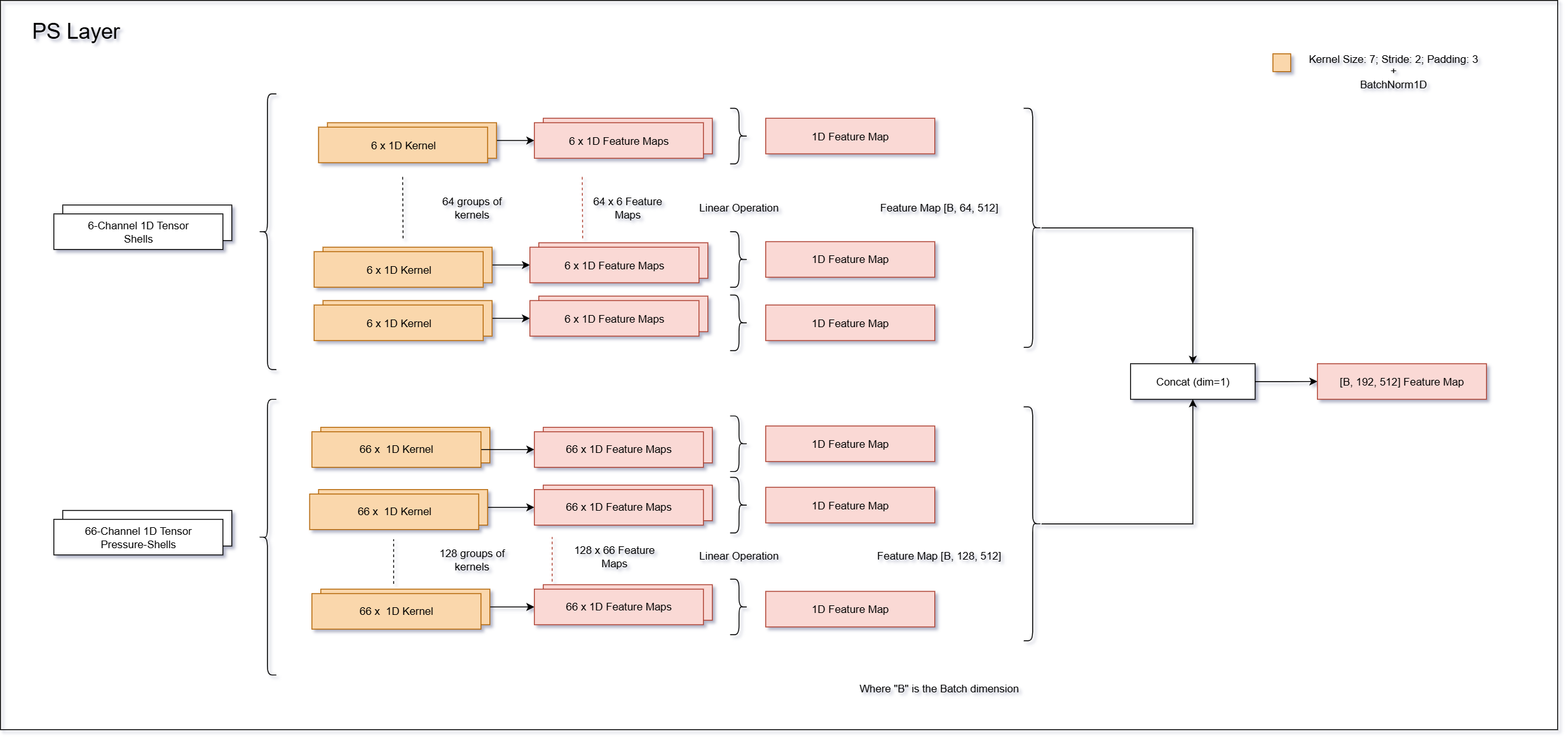}
	\caption{Detailed PS Layer (PressureShell Layer)}
    \label{fig:PS_layer}
\end{figure}

The \textit{PS Layer} architecture (Figure~\ref{fig:PS_layer}) begins with separate convolutional layers: one branch for the six shell-based representations, and another specialized branch for the pressure features. 

For these initial convolutions, a kernel size of $7$, stride of $2$, and padding of $3$ were chosen. This configuration was selected to preserve the overall size of the signal after the first convolution while simultaneously providing a larger receptive field at the entry stage of the network. Although smaller kernels such as $3 \times 3$ could have been used to emphasize more localized patterns, the choice of a $7$-wide kernel ensures that the filters have a broader perception window at the beginning of training, allowing the model to capture more global structures of the signature before progressing to deeper layers. In addition, a \textit{BatchNorm1d} layer was applied after each convolution in order to normalize the feature maps, stabilize training, and reduce internal covariate shift, which contributes to faster convergence and improved generalization.

The outputs of these branches are concatenated to generate the final \textit{PS Feature Map}.

\begin{figure}[H]
	\centering
	\includegraphics[width=13cm]{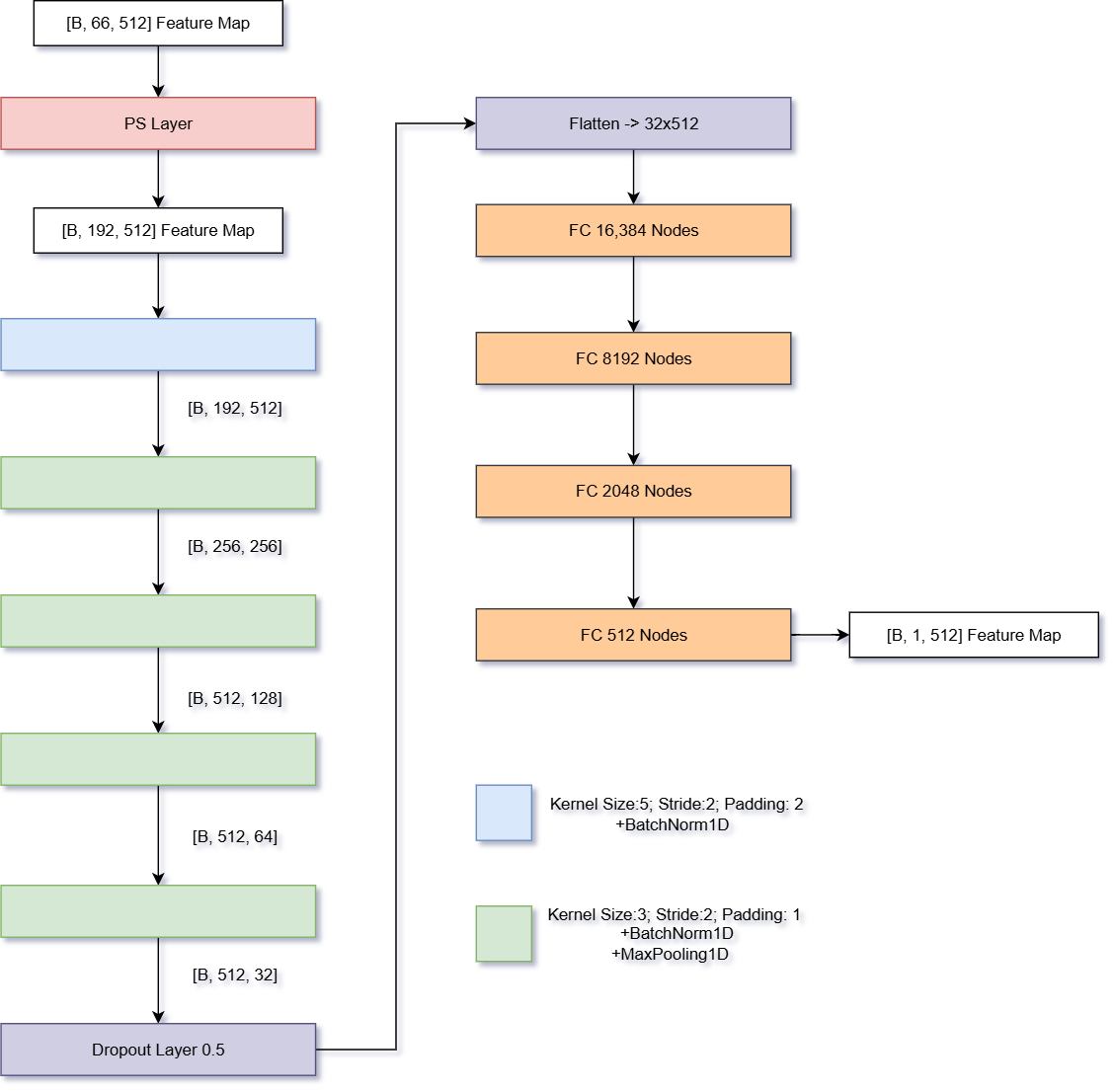}
	\caption{Model Detailed Architecture}
    \label{fig:cascas_architecture}
\end{figure}

From this point, the representation is passed through a sequence of convolutional blocks with progressively increasing depth (Figure~\ref{fig:cascas_architecture}). The first block applies a convolution with kernel size $5$, stride $2$, and padding $2$, followed by a \textit{BatchNorm1d} and ReLU activation. This configuration reduces the temporal resolution while increasing the channel dimension to better capture higher-level abstractions.

Subsequent blocks use convolutions with smaller kernels of size $3$, stride $2$, and padding $1$, each followed by \textit{BatchNorm1d}, ReLU activations, and in some cases \textit{MaxPooling1d}. These progressively reduce the feature map resolution from $[B, 192, 512]$ to $[B, 512, 32]$, effectively compressing the temporal dimension while expanding the channel capacity to 512. This gradual compression ensures that the network retains both local and global structural information from the shell and pressure signals. A dropout layer with rate $0.5$ is applied at the end of this stage to prevent overfitting and improve generalization.

The resulting feature map is then flattened to a vector of size $[B, 1, 16.384]$ (i.e., $ 32 \times 512 = 16.384)$, which is passed through a series of fully connected layers. The first dense layer expands the representation to $16{.}384$ nodes, followed by layers of $8{.}192$, $2{.}048$, and finally $512$ nodes. Each fully connected layer is accompanied by batch normalization and ReLU activations to stabilize learning and increase non-linear expressivity. The last layer produces a compact embedding of size $[B, 1, 512]$, which represents the final signature feature vector.

This embedding is then used within the Siamese training framework. By applying a contrastive loss, the network learns a similarity function that minimizes the distance between embeddings of genuine signature pairs while maximizing the distance for forged pairs. This structure ensures that both the shape-based aspects (captured by the shell signals) and the dynamic aspects (captured by the pressure vectors) are jointly encoded into a discriminative representation.

\subsubsection{Discussion and Design Challenges during Experimentation}

Note that during training using the architecture presented in \ref{fig:cascas_architecture}, the final embeddings are not normalized.

Observing this effect on the contrastive loss function:

\[
L = \frac{1}{N} \sum_{i=1}^{N} \left[ (1 - y) \cdot D^2 + y \cdot \max(0, m - D)^2 \right],
\]

For this loss to be minimum, the distance \textit{D} should be equals to \textit{0} when the target is \textit{0} (original pairs), and the distance should be \textbf{greater than the margin} when the target is \textit{1} (forged pairs):

\[
L_{\min} =
\begin{cases}
D = 0, & \text{if } y = 0, \\[6pt]
D > m, & \text{if } y = 1.
\end{cases}
\]

If the embedding can have any value and it is not normalized, the ``forged distance'' could achieve any value beyond the margin without restrictions.

At first, this seemed to be a potential issue. Intuitively, unconstrained spaces tend to cause instability, so in the initial attempts a normalization step was introduced in the final layer, projecting the embeddings onto a hypersphere via euclidean normalization. Each feature vector $v \in \mathbb{R}^d$ was projected onto the unit hypersphere as

\[
\hat{v} = \frac{v}{\|v\|_2} \quad \text{with} \quad 
\|v\|_2 = \sqrt{\sum_{i=1}^d v_i^2}.
\]

\begin{figure}[H]
	\centering
	\includegraphics[width=\linewidth]{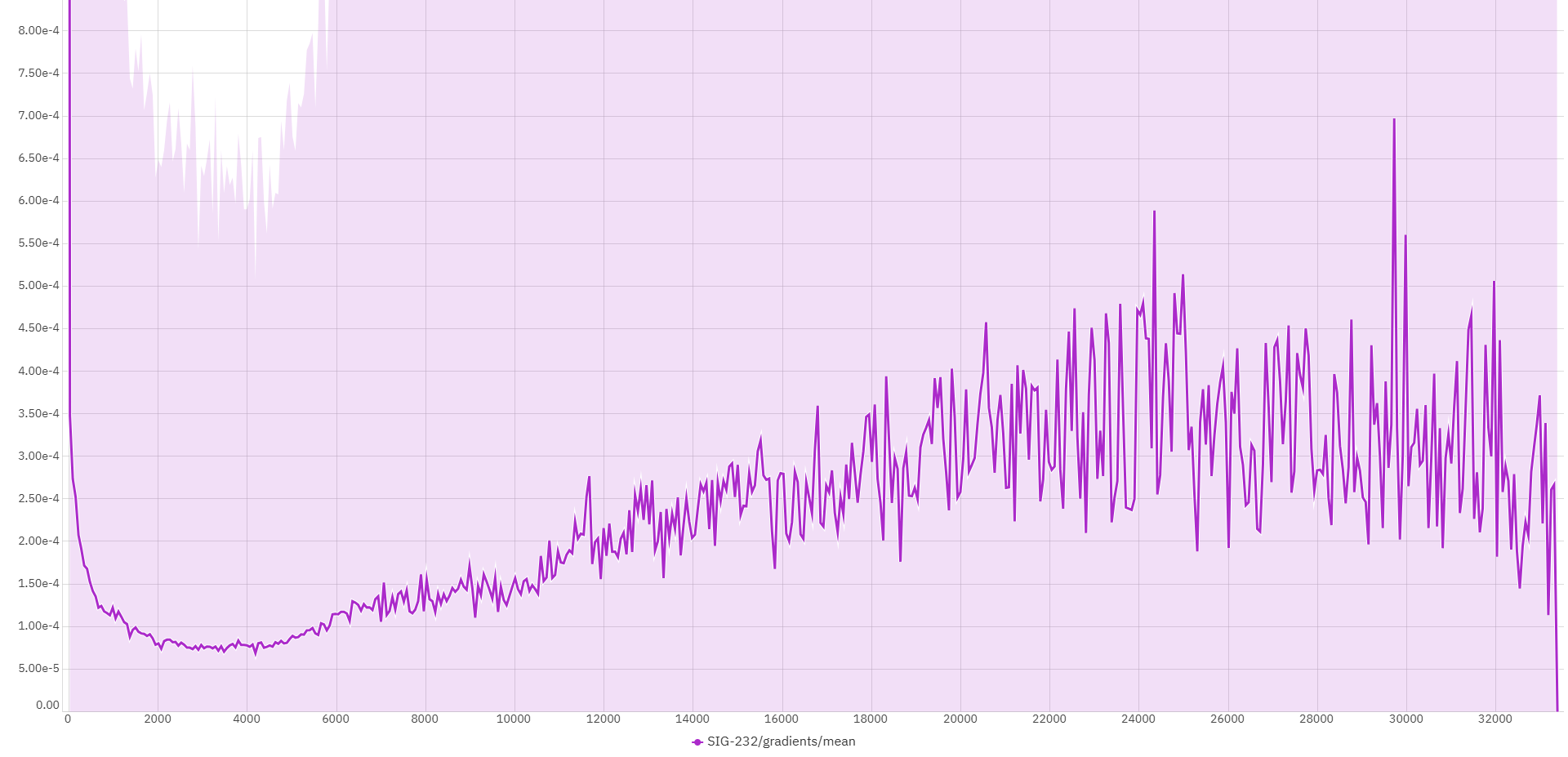}
	\caption{Average (in terms of batches) of the gradients for each iteration for the model using L2 normalization}
    \label{fig:gradients_mean}
\end{figure}

During the experiments, however, the model showed strong instability in the training process when a hypersphere normalization layer was applied at the end of the embedding as shown in (Figure~\ref{fig:gradients_mean}) as an example of trained model.

Although this normalization enforces unit norm embeddings ($\|\hat{v}\|_2 = 1$), 
the gradient updates became highly sensitive to noise. 
This sensitivity led to large oscillations in the loss function. 

This oscillation led to lack of generalization and poor convergence. This can be seen in the loss function behavior:

\begin{figure}[H]
	\centering
	\includegraphics[width=\linewidth]{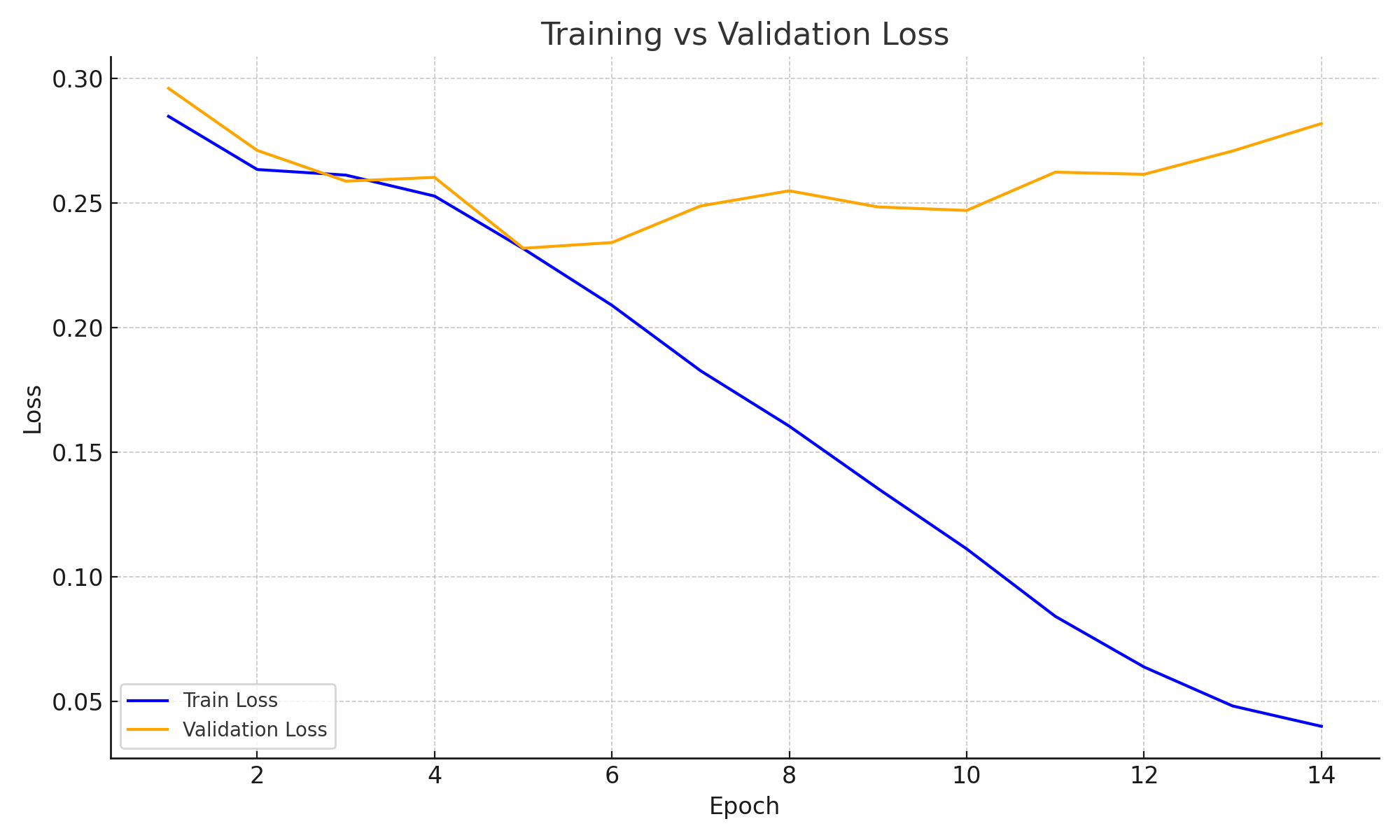}
	\caption{Train X Loss function behavior using the L2 normalization}
    \label{fig:train_val_loss}
\end{figure}

Once this normalization step was removed, the optimization process stabilized: 

\begin{figure}[H]
	\centering
	\includegraphics[width=\linewidth]{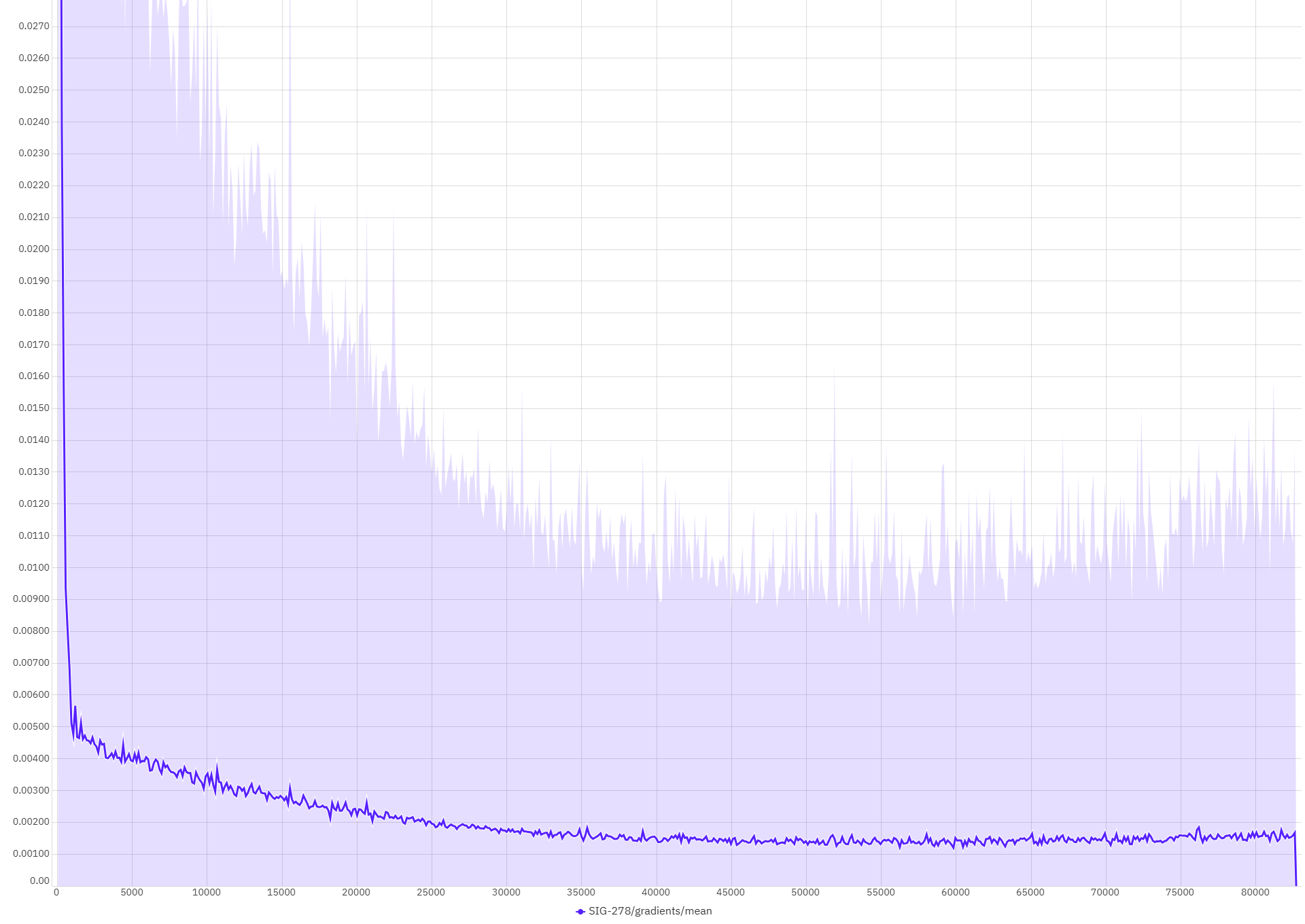}
	\caption{Average (in terms of batches) of the gradients for each iteration for the model without L2 normalization}
    \label{fig:gradients_mean_right}
\end{figure}

The gradients became more stable, resulting in best loss decay:

\begin{figure}[H]
	\centering
	\includegraphics[width=\linewidth]{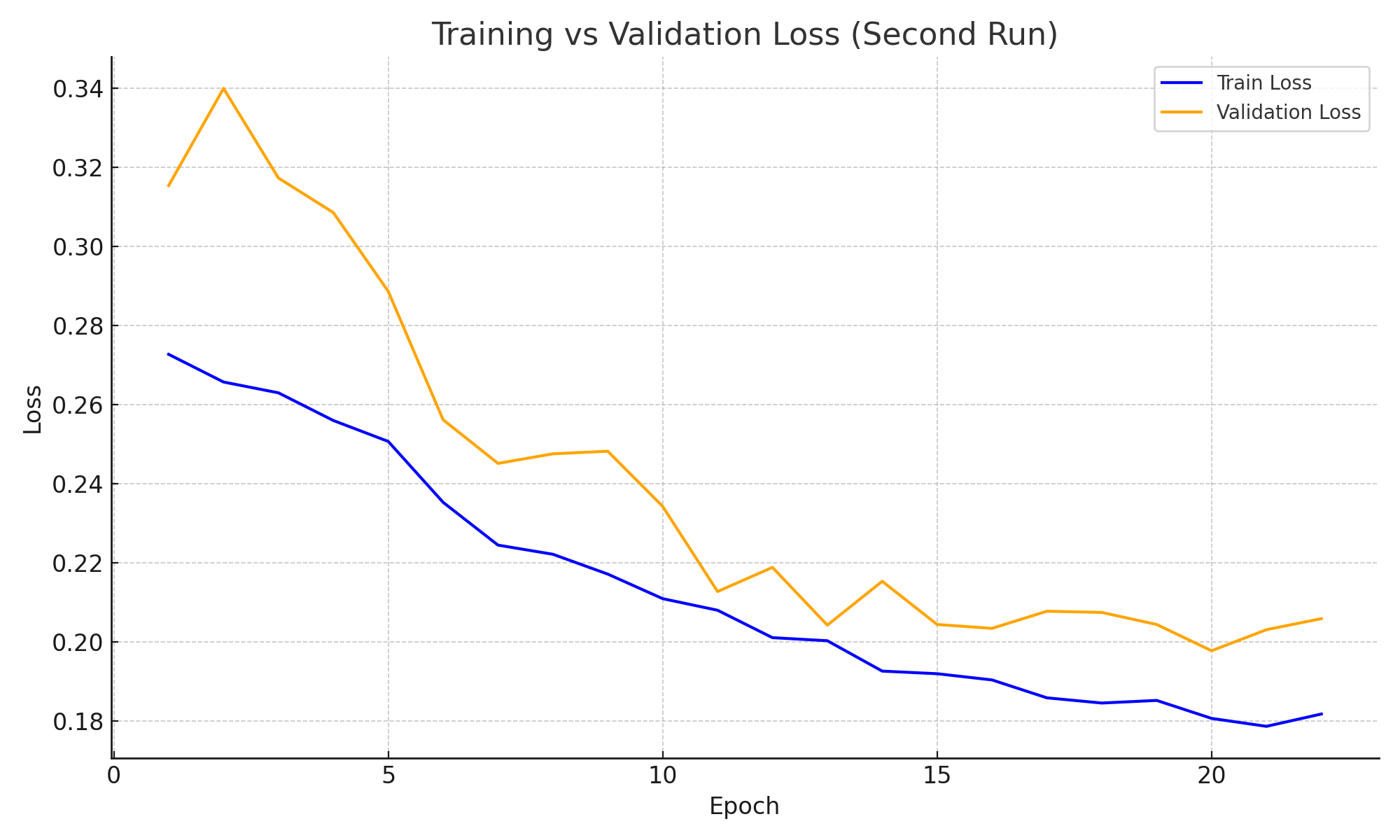}
	\caption{Train X Loss function behavior without the L2 normalization}
    \label{fig:train_val_loss_2}
\end{figure}

The instability observed when applying hypersphere normalization at the final embedding can be explained by several factors. 

First, the normalization constrains all feature vectors to have unit norm, which amplifies the effect of small perturbations during training: even minor noise in the gradient updates can cause large angular shifts in the representation space. Since contrastive loss depends heavily on the relative distances between pairs of embeddings, these abrupt angular changes result in strong oscillations of the loss curve. 

Moreover, the magnitude of the embeddings carries discriminative information itself: it reflects the “energy” or amplitude of the representation. By forcing all embeddings onto the unit hypersphere, this attribute is discarded, eliminating a potentially valuable source of separation between genuine and forged signatures.

Once this normalization step was removed, the embeddings were allowed to evolve more smoothly in Euclidean space, which led to a more monotonic decay in loss values.

\subsubsection{Observing the Distribution}

After several experiments, the loss, in average, converged to 0.18 for training and 0.20 for validation. This can be interpreted by observing the distances generated by the trained model:

\begin{figure}[H]
	\centering
	\includegraphics[width=12cm]{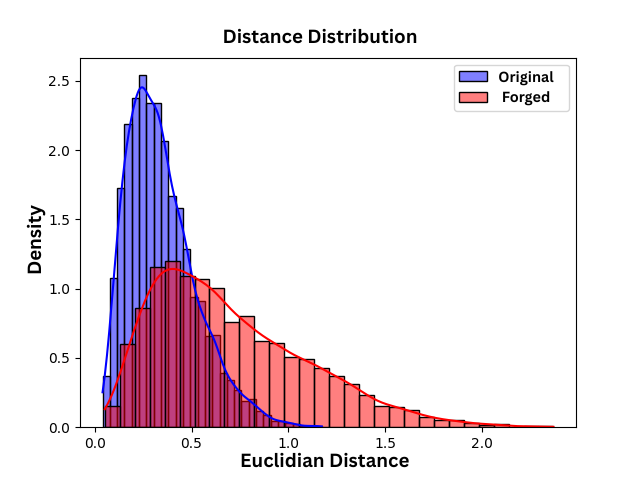}
	\caption{Distribution generated by a model with a 0.18 training loss in the train set}
    \label{fig:train_val_loss_2}
\end{figure}

\begin{figure}[H]
	\centering
	\includegraphics[width=12cm]{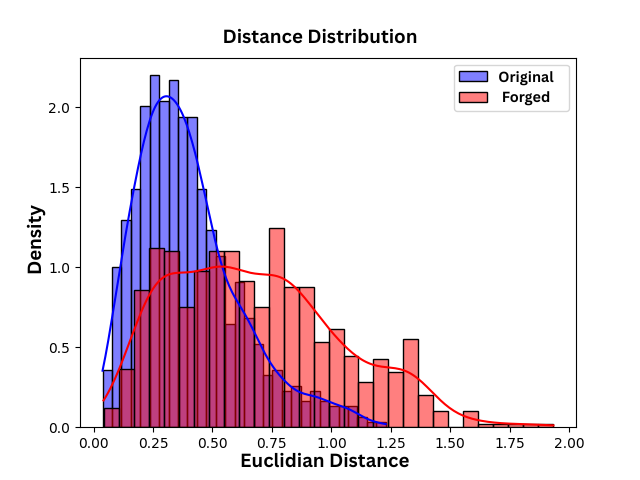}
	\caption{Distribution generated by a model with a 0.20 validation loss in the validation set}
    \label{fig:train_val_loss_2}
\end{figure}

It can observe that this model is able to establish a certain degree of separation between genuine and forged signatures. However, the overlap between the two distributions remains significant, indicating that while the embeddings capture some discriminative patterns, the distinction is not strong enough to ensure a reliable classification.

\begin{figure}[H]
	\centering
	\includegraphics[width=12cm]{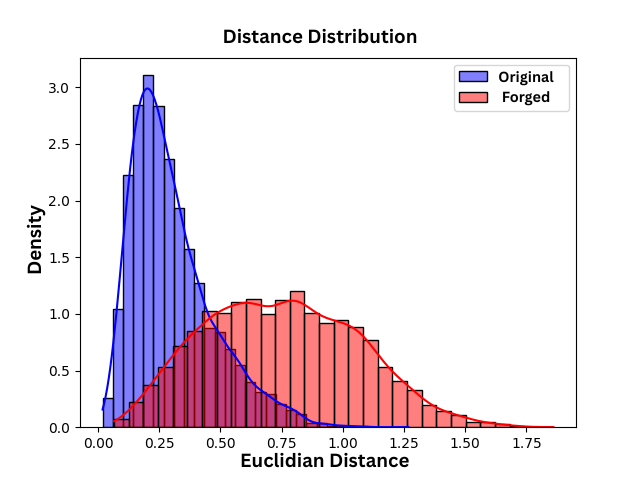}
	\caption{Distribution generated by a model with a 0.13 training loss in the training set}
    \label{fig:dist_ci_resnet_train}
\end{figure}

\begin{figure}[H]
	\centering
	\includegraphics[width=12cm]{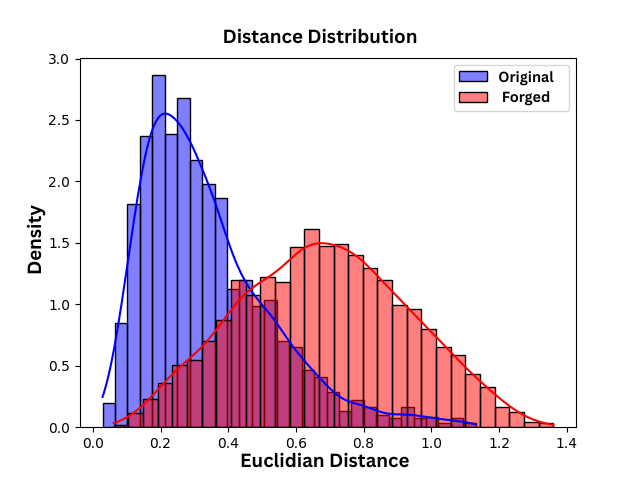}
	\caption{Distribution generated by a model with a 0.14 validation loss in the validation set}
    \label{fig:dist_ci_resnet_val}
\end{figure}

Just for comparison purposes, Figures \ref{fig:dist_ci_resnet_train} and \ref{fig:dist_ci_resnet_val} illustrate the distributions generated by another model that achieved lower losses of 0.13 in training and 0.14 on average in validation. Compared to the previous case, this model exhibits a clearer separation between genuine and forged samples. Although some overlap still exists, the discriminative boundary is noticeably sharper, suggesting that the embeddings learned by this model are more consistent and generalizable.

The model above was trained using the shell-based pre-processing pipeline in conjunction with the ResNet-1D backbone, which is detailed in the following section.

\subsubsection{The ResNet-1D}

So, in order to determine if the it can improve this general performance (check if the "issue" was the lack of data information or the model itself) and to create a fairer comparison to a way deeper model like the ResNet34 used in the raw signature images training, an additional model was trained based on a one-dimensional adaptation of the ResNet34 architecture. This version was capable of directly processing 1D tensors, making it suitable for shell-based representations. The code for this adapted ResNet34 was inspired from an open-source project\footnote{\url{https://github.com/hsd1503/resnet1d}}, which provided a baseline for evaluating the performance against the raw images.

The final architecture used to generate the benchmark exposed in the following section was built in this manner:

\begin{figure}[H]
	\centering
	\includegraphics[width=5cm]{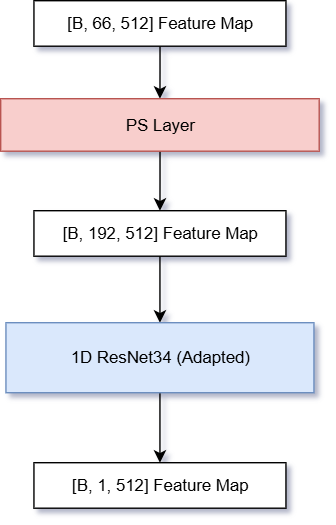}
	\caption{Final ResNet34 1D adaptation}
    \label{fig:resnet34_1d}
\end{figure}

Therefore, for the next section of results and benchmark comparison, the model trained using the shell data uses the ResNet1D as backbone.

\section{Results and Discussion}

Following the same protocol of the first part, the training set is divided into the same dataset groups used previously: CI, CG, IG, and CIG. This consistency allows for a fair and controlled comparison between traditional input representations and the shell-based representation introduced in this part.

\begin{itemize}
    \item \textbf{CI (CEDAR + ICDAR)} combines the shell representations extracted from \textit{trainingCEDAR} and \textit{trainingICDAR}, and similarly for validation and testing sets.
    \item \textbf{CG (CEDAR + GPDS)} includes data from CEDAR and GPDS using the same methodology, now with shell-processed features.
    \item \textbf{IG (ICDAR + GPDS)} is formed by combining the filtered shell-based data from ICDAR and GPDS.
    \item \textbf{CIG (CEDAR + ICDAR + GPDS)} is the full combination of shell-based features from all datasets.
\end{itemize}

The aim of this second part is to analyze whether the morphological shell-based representation enhances model robustness and cross-dataset generalization. 

The same evaluation strategy is adopted: models are trained with contrastive and triplet losses and tested across datasets using identical groupings. Performance metrics such as AUC and ROC curves are then compared with the results from the first part. This setup allows us to verify if the shell-based approach leads to performance gains and improved invariance to dataset-specific styles.

\subsection{Contrastive Loss Results}

\begin{table}[H]
\centering
\caption{"Shell "Model Performance Across Datasets using Contrastive Loss (AUC values)}
\resizebox{\textwidth}{!}{%
\begin{tabular}{|l|l|c|c|c|c|}
\hline
\textbf{Group / Training} & \textbf{Model} & \multicolumn{4}{c|}{\textbf{Dataset Test}} \\ \hline
                          &                & \textbf{CEDAR} & \textbf{ICDAR} & \textbf{GPDS} & \textbf{CIG} \\ \hline
1 (CI) & model\_2025-08-04\_16:12:16 & 0.81 & 0.89 & \textbf{0.63} & 0.82 \\ \hline
2 (CG) &model\_2025-08-05:13:26:19  &   0.76          & \textbf{0.73}           & 0.71 &   0.72       \\ \hline
3 (IG) & model\_2025-08-05\_16:47:18 & \textbf{0.73}            &  0.81         & 0.71 &     0.76      \\ \hline

\end{tabular}%
}
\label{tab:ModelPerformance}
\end{table}

After analyzing these values, it is noticed that there are no discrepant per-dataset outliers. This suggests that the shell-based preprocessing reduced the dataset-specific biases. 

\subsection{Triplet Loss Results}

\begin{table}[H]
\centering
\caption{"Shell "Model Performance Across Datasets using Triplet Loss (AUC values)}
\resizebox{\textwidth}{!}{%
\begin{tabular}{|l|l|c|c|c|c|}
\hline
\textbf{Group / Training} & \textbf{Model} & \multicolumn{4}{c|}{\textbf{Dataset Test}} \\ \hline
                          &                & \textbf{CEDAR} & \textbf{ICDAR} & \textbf{GPDS} & \textbf{CIG} \\ \hline
1 (CI) &model\_2025-08-16\_17:07:27&0.76&0.79& \textbf{0.66} & 0.74 \\ \hline
2 (CG) & model\_2025-08-16\_17:12:33 &  0.72       &    \textbf{0.72}      & 0.67 &   0.71     \\ \hline
3 (IG) & model\_2025-08-16\_18:15:18 &    \textbf{0.72}      &   0.73      & 0.69 &   0.71     \\ \hline

\end{tabular}%
}
\label{tab:ModelPerformance}
\end{table}

In contrast, when comparing the two training strategies (contrastive and triplet losses), the shell-based representation does not exhibit the same improvements observed with raw images. This indicates that important discriminative details were lost during preprocessing, since triplet loss—designed to benefit from richer feature relationships—was unable to achieve gains over contrastive loss in this representation. Furthermore, when comparing directly with the overall performance achieved using raw images (which will be illustrated in the graphs presented later), it becomes clear that the average performance with shells is consistently lower.

To demonstrate these effects, a model was trained on a single dataset and then tested on the others.

\begin{table}[H]
\centering
\caption{Model Performance Across Datasets using Contrastive Loss (AUC values)}
\resizebox{\textwidth}{!}{%
\begin{tabular}{|l|l|c|c|c|c|}
\hline
\textbf{Group / Training} & \textbf{Model} & \multicolumn{4}{c|}{\textbf{Dataset Test}} \\ \hline
                          &                & \textbf{CEDAR} & \textbf{ICDAR} & \textbf{GPDS} & \textbf{CIG} \\ \hline
(CEDAR - SHELLS) &model\_2025-08-16\_19:27:19& 0.74 & \textbf{0.72}  & \textbf{0.64} & 0.70 \\ \hline
(CEDAR - IMAGES) &model\_2025-08-17\_23:09:33& 0.89 & \textbf{0.91}  & \textbf{0.59} & 0.80 \\ \hline

\end{tabular}%
}
\label{tab:ModelPerformance}
\end{table}

From these results, it can be observed that the use of shell-based preprocessing led to a form of \textit{performance equalization} across datasets. While the raw image training achieved higher scores on CEDAR and ICDAR, its performance dropped considerably when evaluated on GPDS. With preprocessing, however, the performance became more balanced, mitigating large discrepancies between datasets. As expected, GPDS remains the most challenging dataset. At the same time, it also becomes evident that CEDAR and ICDAR share similar characteristics, which explains their consistently higher scores when compared to GPDS.

\begin{figure}[H]
	\centering
	\includegraphics[width=13cm]{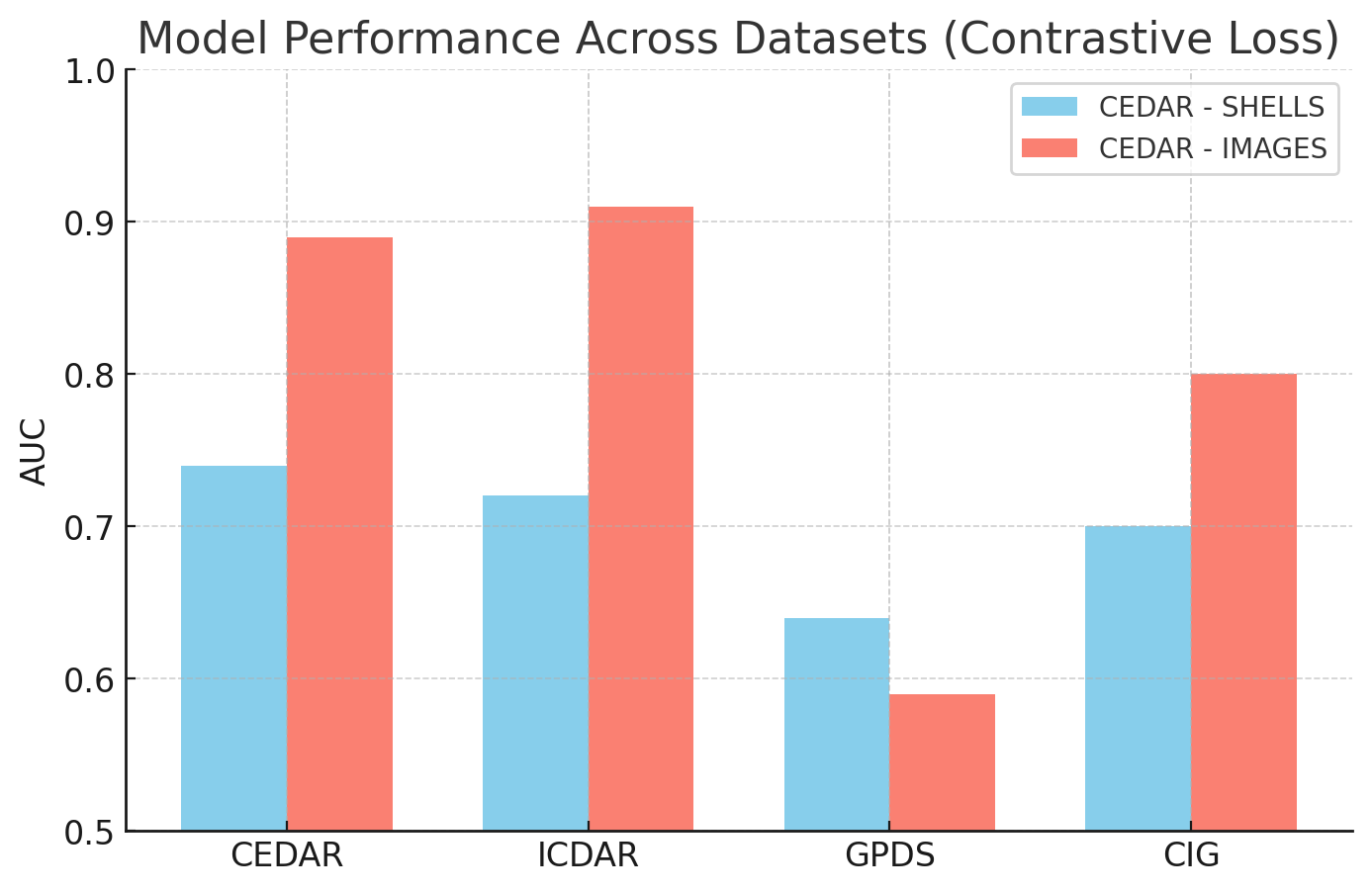}
	\caption{CEDAR Training Comparison}
    \label{fig:CEDAR_comparison}
\end{figure}

In Figure~\ref{fig:CEDAR_comparison}, these aspects become visually clear. The shell-based preprocessing narrows the performance disparity across datasets, while the image-based approach, although stronger on CEDAR and ICDAR, shows a strong decline on GPDS. This graphical comparison supports the idea of normalization brought by the preprocessing pipeline.

\paragraph{Raw vs. Shell-Preprocessed Models.}
Table~\ref{tab:RawVsShellPerformance} summarizes the cross-dataset AUC for models trained with raw images and with the proposed shell-based preprocessing (all using Contrastive Loss). 

Overall, the shell pipeline shows a \emph{performance equalization} across datasets: it reduces the gap between the best and worst cases and notably improves transfer to GPDS, the most challenging dataset. At the same time, CEDAR and ICDAR remain the most similar pair—both approaches reach their highest scores on these datasets—indicating shared characteristics between them. The price of the \textit{performance equalization} is a reduction of peak performance on the in-domain datasets (e.g., CEDAR/ICDAR), but with a consistent gain in generalization, especially toward GPDS.

\begin{table}[H]
\centering
\caption{Comparison of Raw-Image and Shell-Preprocessed Models using Contrastive Loss (AUC values). Bold indicates the cross-test performance and bold and blue indicates the best performance per cross-test.}
\resizebox{\textwidth}{!}{%
\begin{tabular}{|l|l|c|c|c|c|c|}
\hline
\textbf{Group / Training} & \textbf{Approach} & \textbf{Model} & \textbf{CEDAR} & \textbf{ICDAR} & \textbf{GPDS} & \textbf{CIG} \\ \hline
\multirow{2}{*}{1 (CI)} 
 & Raw Images & model\_2025-01-05\_16:57:36 & 1.00 & 0.99 & \textbf{0.53} & 0.84 \\ \cline{2-7}
 & Shells     & model\_2025-08-04\_16:12:16 & 0.81 & 0.89 & \textbf{\textcolor{blue}{0.63}} & 0.82 \\ \hline
\multirow{2}{*}{2 (CG)} 
 & Raw Images & model\_2025-01-05\_20:51:16 & 1.00 & \textbf{\textcolor{blue}{0.83}} & 0.81 & 0.92 \\ \cline{2-7}
 & Shells     & model\_2025-08-05\_13:26:19 & 0.76 & \textbf{0.73} & 0.71 & 0.72 \\ \hline
\multirow{2}{*}{3 (IG)} 
 & Raw Images & model\_2025-01-05\_22:22:18 & \textbf{\textcolor{blue}{0.76}} & 0.99 & 0.75 & 0.85 \\ \cline{2-7}
 & Shells     & model\_2025-08-05\_16:47:18 & \textbf{0.73} & 0.81 & 0.71 & 0.76 \\ \hline
\end{tabular}%
}
\label{tab:RawVsShellPerformance}
\end{table}

\subsection*{Per-Group Graphical Comparisons}

The graphical comparisons reinforce the tendencies already summarized in Table~\ref{tab:RawVsShellPerformance}. 

\subsubsection{CI training (CEDAR+ICDAR)}
\begin{figure}[H]
    \centering
    \includegraphics[width=13cm]{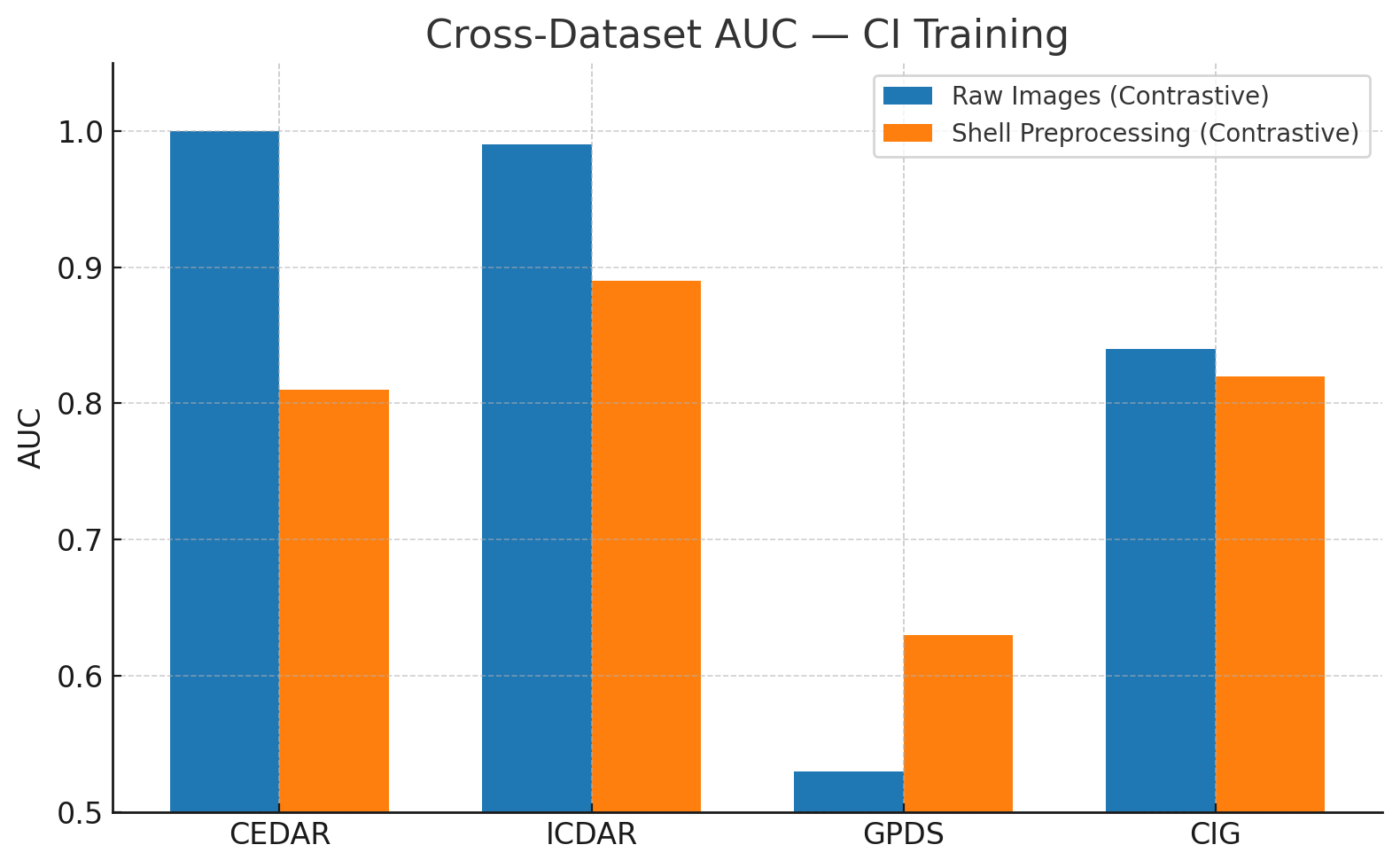}
    \caption{Cross-dataset AUC for models trained on CI. Shell preprocessing narrows the spread and lifts GPDS notably (from 0.53 to 0.63), evidencing the normalization effect.}
    \label{fig:auc_bar_CI}
\end{figure}

For CI training (Figure~\ref{fig:auc_bar_CI}), the raw image model is superior in terms of the own dataset, reaching almost perfect discrimination in CEDAR and ICDAR. However, its performance worsens abruptly when tested on GPDS, which drops to 0.53 AUC. By contrast, the shell-preprocessed model achieves more balanced results, reducing peak scores on CEDAR/ICDAR but raising the GPDS AUC to 0.63, thus illustrating the \textit{equalization effect} induced by preprocessing.

\subsubsection{CG training (CEDAR+GPDS)}
\begin{figure}[H]
    \centering
    \includegraphics[width=13cm]{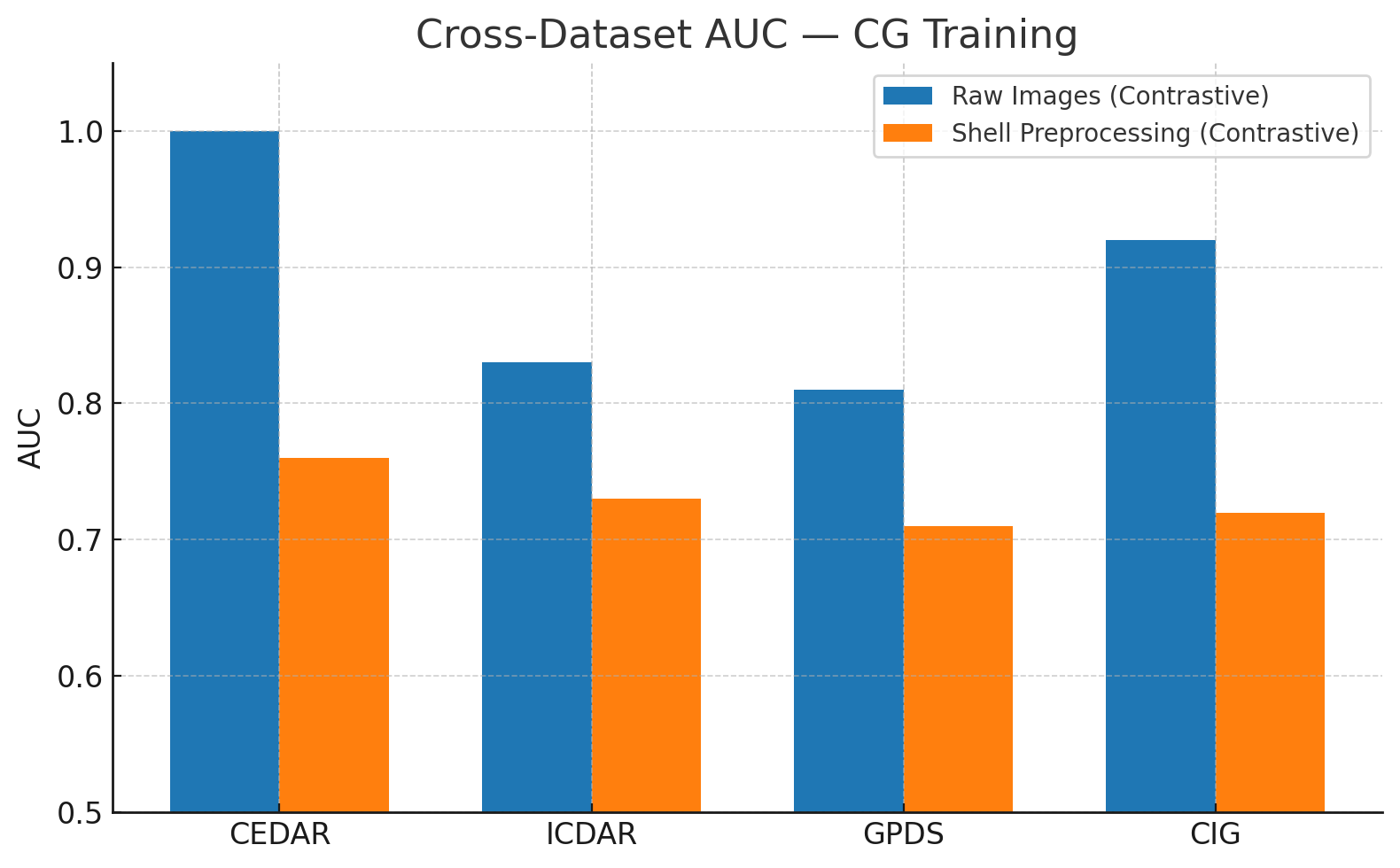}
    \caption{Cross-dataset AUC for models trained on CG. The shell pipeline yields more balanced outcomes across datasets, at the cost of slightly lower peaks on CEDAR/GPDS, reinforcing the robustness–peak trade-off.}
    \label{fig:auc_bar_CG}
\end{figure}

In CG training (Figure~\ref{fig:auc_bar_CG}), a similar trade-off is evident. Raw images produce very high results on CEDAR and a considerably high results on GPDS, but the shell-based variant compresses the overall range, lowering the top values while reducing the drop on the more difficult datasets. Still, the raw-image model was superior in all cases.

\subsubsection{IG training (ICDAR+GPDS)}
\begin{figure}[H]
    \centering
    \includegraphics[width=13cm]{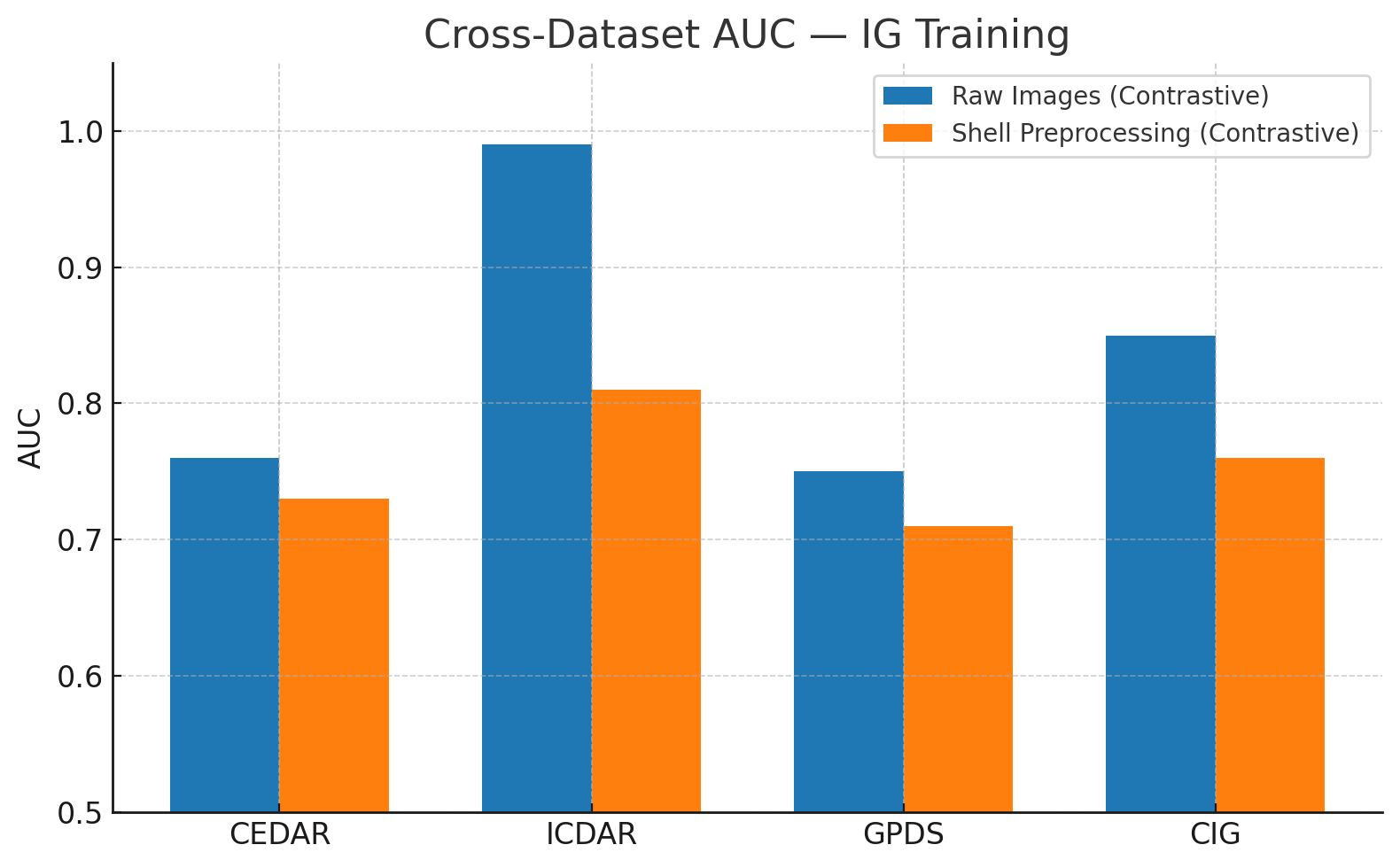}
    \caption{Cross-dataset AUC for models trained on IG. The shell preprocessing reduces variance across tests and sustains competitive results on GPDS relative to the raw-image baseline.}
    \label{fig:auc_bar_IG}
\end{figure}

The IG training setup (Figure~\ref{fig:auc_bar_IG}) provides further evidence of this trend. Raw-image training produces high values for ICDAR, yet the performance drops inconsistently across CEDAR and GPDS. The shell-preprocessed model, however, reduces this fluctuation, maintaining a steadier profile across all datasets and again reinforcing the normalization interpretation.

\begin{figure}[H]
    \centering
    \includegraphics[width=13cm]{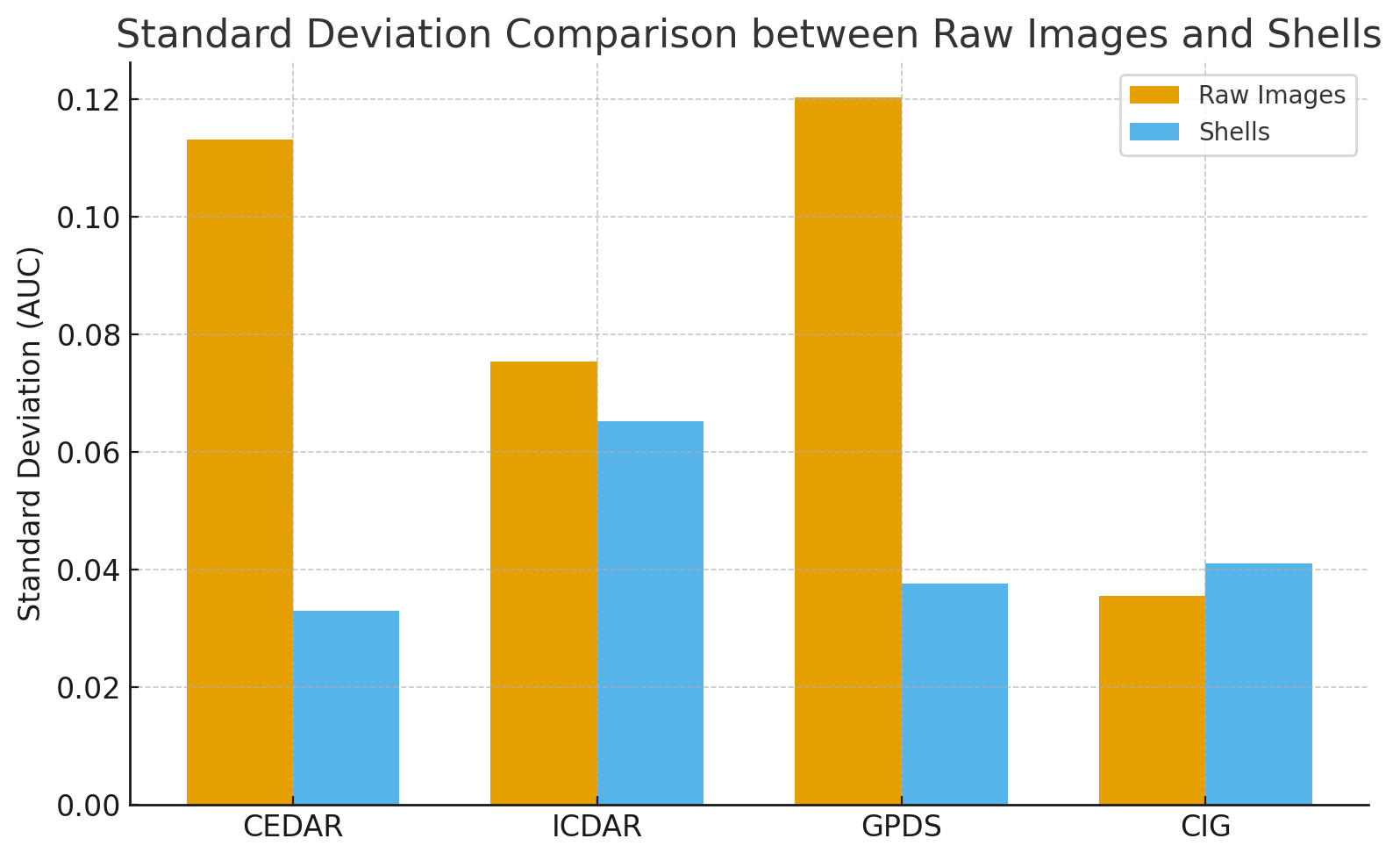}
    \caption{Standard deviation of the performance metrics across datasets. Lower variance indicates more consistent generalization behavior when applying the shell-based preprocessing.}
    \label{fig:std_models}
\end{figure}

To make this performance normalization more evident, a standard deviation graph is presented in Figure~\ref{fig:std_models}, highlighting how the variance across datasets decreases when applying the proposed preprocessing strategy. This visualization reinforces the observation that, despite differences in absolute performance, the shell-based approach provides more stable generalization, as shown by the lower spread of results.

  \chapter{Conclusion and Future Work}

This work investigated the use of Siamese neural networks for offline signature verification, focusing on their ability to generalize across different datasets. A complete training pipeline was implemented, including dataset preparation with writer-disjoint splits, balanced pair generation, and data augmentation strategies. These steps helped the model avoid overfitting and improved the reliability of the evaluation.

For the fist part of this work, the architecture developed, based on a ResNet-34 backbone, achieved consistent results in cross-dataset experiments using CEDAR, ICDAR, and GPDS Synthetic. This showed that the proposed pipeline is effective in learning features that transfer beyond the training dataset, which is one of the main challenges in signature verification using neural networks.

In addition to the image-based approach, a shell-based preprocessing method was introduced. This method extracts successive one-dimensional contours from the ink trace, providing a compact representation of structural handwriting features. While the shell inputs reduced dataset-specific biases and improved stability between datasets, they also led to lower absolute performance compared to raw images, indicating that some discriminative details were lost.

In summary, the main contributions of this work are:
\begin{enumerate}
    \item The implementation of a training pipeline for Siamese networks with emphasis on cross-dataset evaluation, including writer-disjoint tests, generation of dissimilar pairs from different writers, and controlled balancing between positive and negative pairs.
    \item The design and testing of a ResNet-based Siamese architecture for signature verification.
    \item The investigation of a shell-based preprocessing pipeline as an alternative representation of signatures.
\end{enumerate}

For future work, it is possible to:
\begin{itemize}
    \item Explore more advanced loss optimization strategies, such as the quadruplet loss \cite{chen2017beyond}.
    \item Implement smarter sampling techniques during training (e.g., hard negative mining), which may enhance the effectiveness of triplet loss.
    \item Investigate meta-learning or few-shot learning techniques, enabling better adaptation to new users with limited genuine samples.
    \item Further refine the shell-based pipeline to recover fine-grained discriminative features while maintaining its generalization benefits.
    \item Due to the significant downsampling that the ICDAR dataset requires in order to balance all three datasets, it would be interesting to study the impact of only considering CEDAR and GPDS (which would still require some downsampling to balance, but to a lesser extent) and, in the end, consider only the GPDS for study using all the dataset.
\end{itemize}

This dual investigation on both robust architectures and shell-based representations creates a new path toward a new generation of signature verification systems that combine accuracy, generalization, and adaptability.

  \backmatter

  \ifx\AbntTexType\StringNum
    \bibliographystyle{abntex2-num}
  \else
    \bibliographystyle{abntex2-alf}
  \fi

  \bibliography{referencias}

  \appendix
  \chapter{Pseudo-Algorithms}

This appendix presents the detailed pseudo-algorithms corresponding to each core function used in the shell-based signature pre-processing pipeline. Each algorithm is described step-by-step to facilitate reproducibility and to provide a clear understanding of the internal logic behind the pipeline components, including pruning, shell isolation, and functional transformation.

\begin{algorithm}
\caption{Pruning Procedure}
\begin{algorithmic}[1]
\State \textbf{Input:} Binary image $img \in \{0,1\}^{H \times W}$
\State \textbf{Output:} Skeleton image $skel$, filtered binary image $mask$
\Statex
\State Apply binary morphological opening on $img$
\State Remove small holes in the result
\State Compute the skeleton of the cleaned binary image: $skel \gets skeletonize(img)$
\State Set $mask \gets img$ as a cleaned binary mask
\State \Return $skel$, $mask$
\end{algorithmic}
\end{algorithm}

\begin{algorithm}
\caption{Superior Shell Extraction: \texttt{shellS\_binary}}
\begin{algorithmic}[1]
\State \textbf{Input:} Binary image $img$
\State \textbf{Output:} Binary image $sup\_bin$ with only superior shell
\Statex
\For{each column $j$ in $img$}
    \State Find the first foreground pixel index $i$
    \While{$i <$ image height \textbf{and} $img[i, j] = 1$}
        \State $sup\_bin[i, j] \gets 1$
        \State $i \gets i + 1$
    \EndWhile
\EndFor
\State \Return $sup\_bin$
\end{algorithmic}
\end{algorithm}

\begin{algorithm}
\caption{Inferior Shell Extraction: \texttt{shellI\_binary}}
\begin{algorithmic}[1]
\State \textbf{Input:} Binary image $img$
\State \textbf{Output:} Binary image $inf\_bin$ with only inferior shell
\Statex
\For{each column $j$ in $img$ (from right to left)}
    \State Find the last foreground pixel index $i$
    \While{$i \geq 0$ \textbf{and} $img[i, j] = 1$}
        \State $inf\_bin[i, j] \gets 1$
        \State $i \gets i - 1$
    \EndWhile
\EndFor
\State \Return $inf\_bin$
\end{algorithmic}
\end{algorithm}

\begin{algorithm}
\caption{Residual Shell Extraction: \texttt{res\_binarized}}
\begin{algorithmic}[1]
\State \textbf{Input:} Binary image $img$
\State \textbf{Output:} Residual binary mask $res\_bin$
\Statex
\For{each column $j$ in $img$}
    \For{each row $i$ from top to bottom}
        \While{$img[i, j] = 1$}
            \State $res\_bin[i, j] \gets 1$
            \State $i \gets i + 1$
        \EndWhile
    \EndFor
\EndFor
\State \Return $res\_bin$
\end{algorithmic}
\end{algorithm}

\begin{algorithm}
\caption{Shell to Function Mapping: \texttt{img\_to\_shell\_func}}
\begin{algorithmic}[1]
\State \textbf{Input:} Binary shell image $img$
\State \textbf{Output:} Vectors $shell_s$, $shell_i$
\Statex
\State Flip the image vertically
\For{each column $j$}
    \State Identify foreground rows in column $j$
    \If{any foreground pixel exists}
        \State $shell_i[j] \gets$ top-most pixel index
        \State $shell_s[j] \gets$ bottom-most pixel index
    \EndIf
\EndFor
\State \Return $shell_s$, $shell_i$
\end{algorithmic}
\end{algorithm}
  \chapter{Training Details}

This section presents a series of figures that illustrate the performance of the trained models across various dataset configurations. The evaluation includes confusion matrices and ROC curves, which help analyze classification accuracy and the model's ability to distinguish between genuine and forged signatures.

\section{PART 1}

\subsection{Contrastive Loss Training}

\subsubsection{CEDAR + ICDAR Training}
\begin{figure}[H]
    \centering
    \begin{subfigure}{0.49\textwidth}
        \centering
        \includegraphics[width=\linewidth]{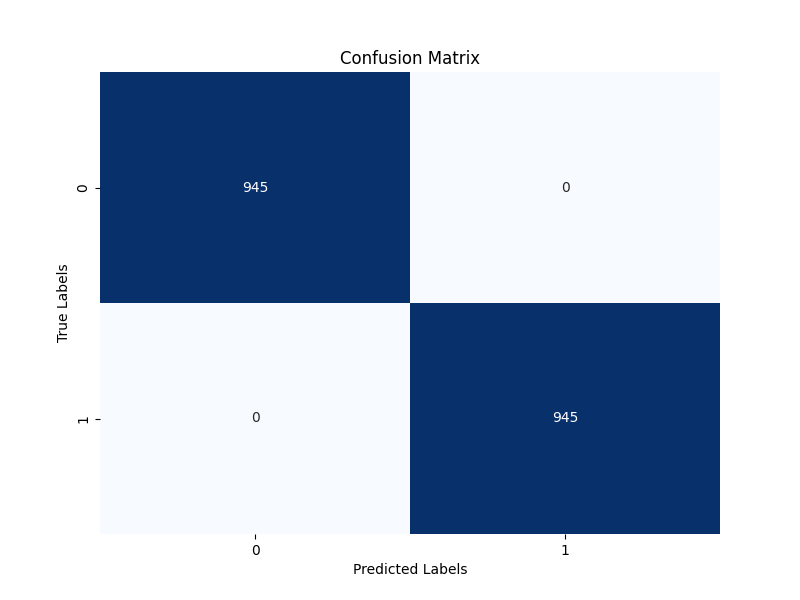}
        \caption{Confusion Matrix for the test set of CEDAR}
        \label{fig:cedar_c}
    \end{subfigure}
    \hfill
    \begin{subfigure}{0.49\textwidth}
        \centering
        \includegraphics[width=\linewidth]{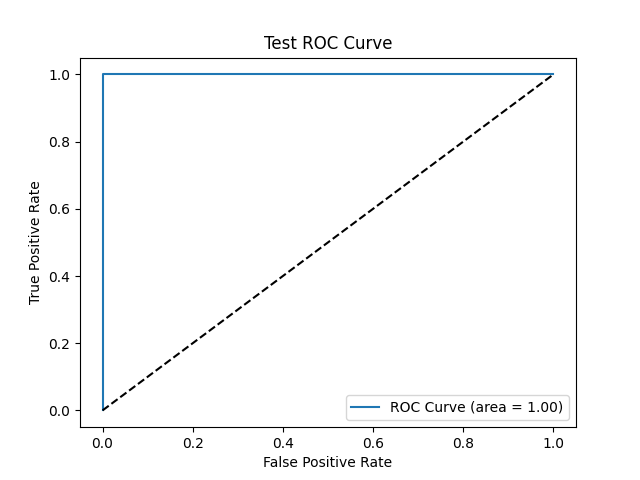}
        \caption{ROC Curve for the test set of CEDAR}
        \label{fig:cedar_roc}
    \end{subfigure}
    \caption{CI - Evaluation metrics for the CEDAR test set}
    \label{fig:cedar_results}
\end{figure}

\begin{figure}[H]
    \centering
    \begin{subfigure}{0.49\textwidth}
        \centering
        \includegraphics[width=\linewidth]{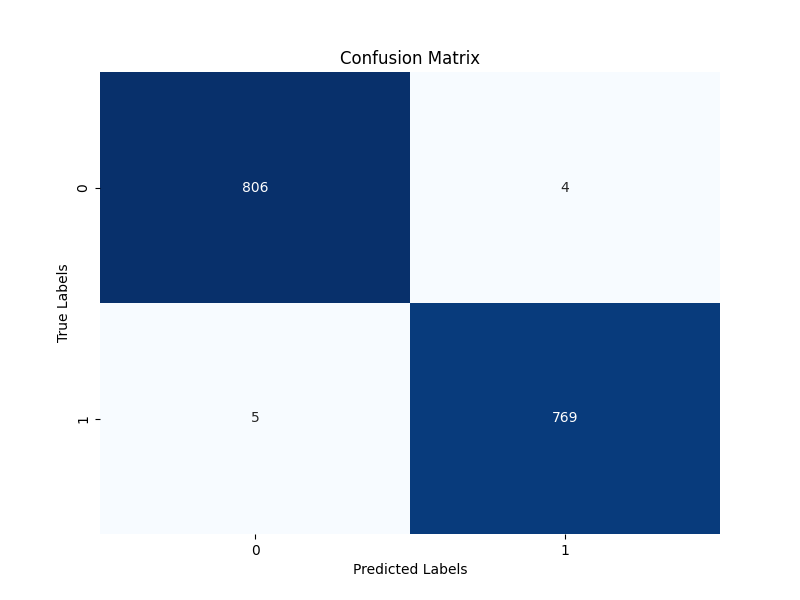}
        \caption{Confusion Matrix for the test set of ICDAR}
        \label{fig:icdar_c}
    \end{subfigure}
    \hfill
    \begin{subfigure}{0.49\textwidth}
        \centering
        \includegraphics[width=\linewidth]{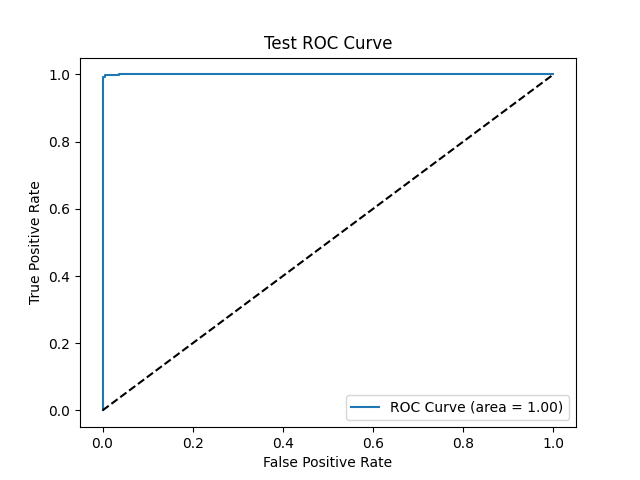}
        \caption{ROC Curve for the test set of ICDAR}
        \label{fig:icdar_roc}
    \end{subfigure}
    \caption{CI - Evaluation metrics for the ICDAR test set}
    \label{fig:icdar_results}
\end{figure}

\begin{figure}[H]
    \centering
    \begin{subfigure}{0.49\textwidth}
        \centering
        \includegraphics[width=\linewidth]{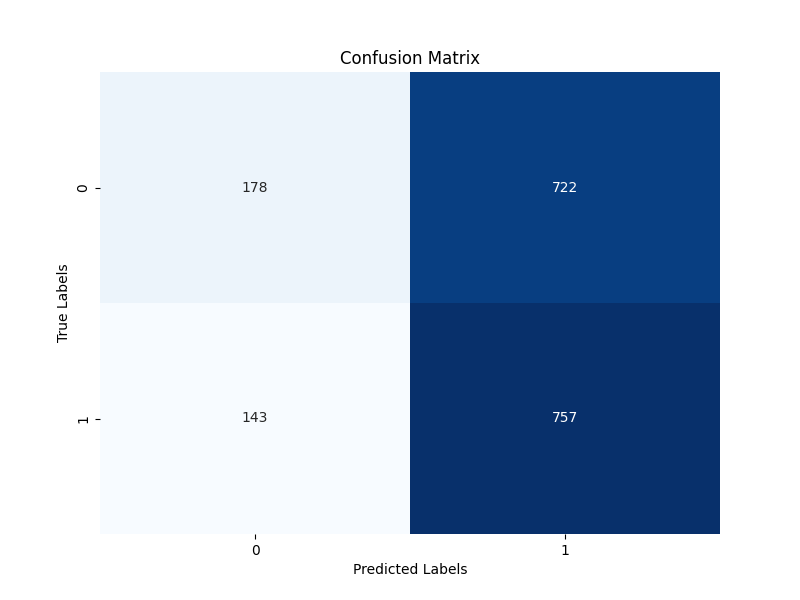}
        \caption{Confusion Matrix for the test set of GPDS}
        \label{fig:gpds_c}
    \end{subfigure}
    \hfill
    \begin{subfigure}{0.49\textwidth}
        \centering
        \includegraphics[width=\linewidth]{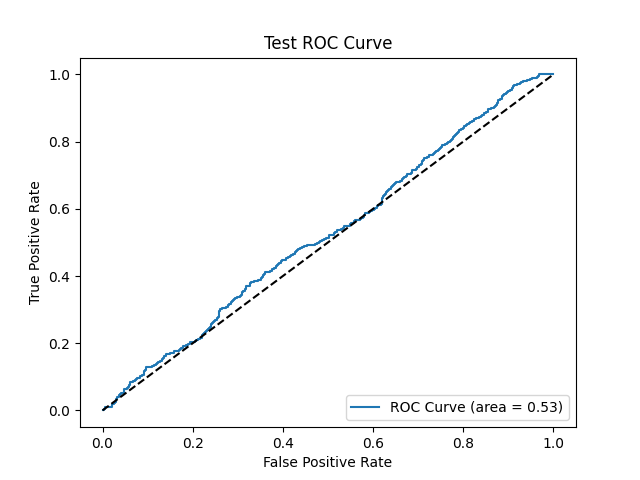}
        \caption{ROC Curve for the test set of GPDS}
        \label{fig:gpds_roc}
    \end{subfigure}
    \caption{CI - Evaluation metrics for the GPDS test set}
    \label{fig:gpds_results}
\end{figure}

\begin{figure}[H]
    \centering
    \begin{subfigure}{0.49\textwidth}
        \centering
        \includegraphics[width=\linewidth]{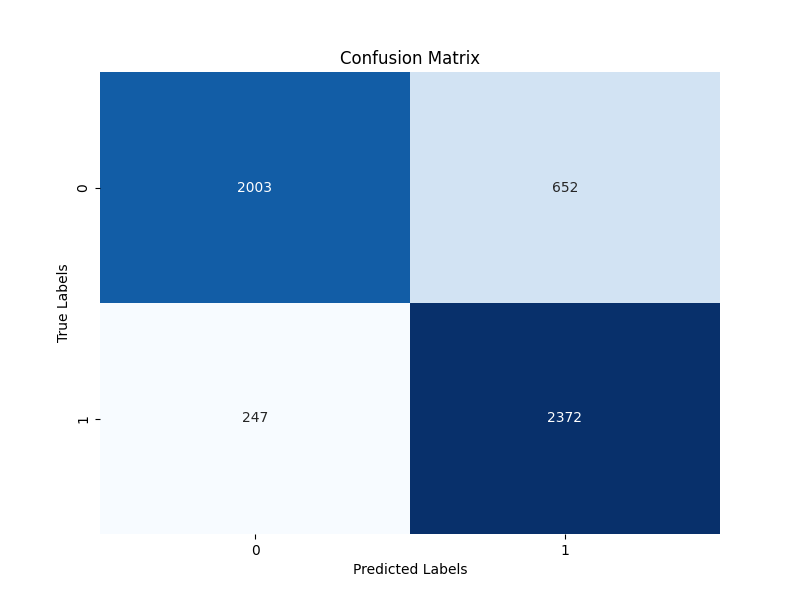}
        \caption{Confusion Matrix for the test set of MERGED}
        \label{fig:merged_c}
    \end{subfigure}
    \hfill
    \begin{subfigure}{0.49\textwidth}
        \centering
        \includegraphics[width=\linewidth]{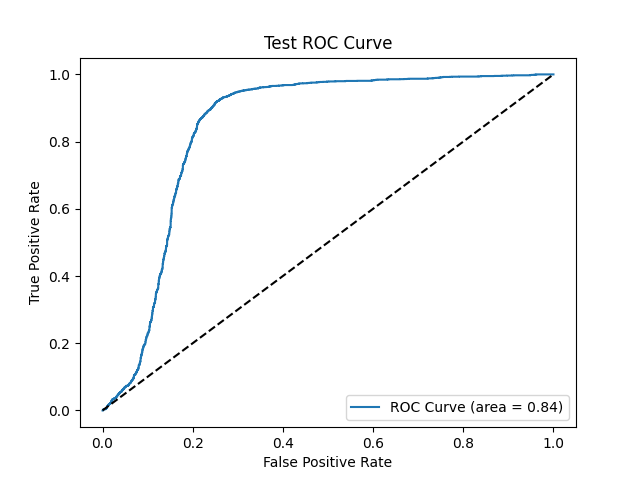}
        \caption{ROC Curve for the test set of MERGED}
        \label{fig:merged_roc}
    \end{subfigure}
    \caption{CI - Evaluation metrics for the MERGED test set}
    \label{fig:merged_results}
\end{figure}

\subsubsection{CEDAR + GPDS Training}
\begin{figure}[H]
    \centering
    \begin{subfigure}{0.49\textwidth}
        \centering
        \includegraphics[width=\linewidth]{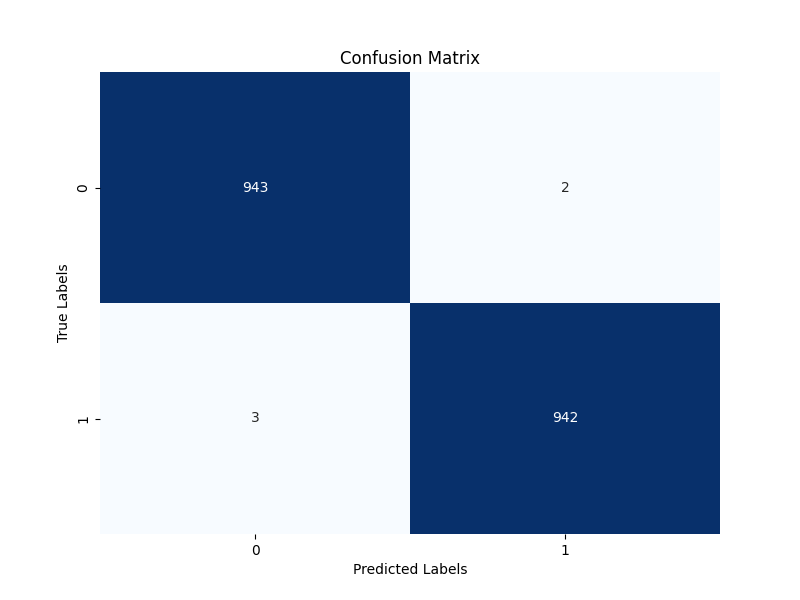}
        \caption{Confusion Matrix for the test set of CEDAR}
        \label{fig:cedar_c}
    \end{subfigure}
    \hfill
    \begin{subfigure}{0.49\textwidth}
        \centering
        \includegraphics[width=\linewidth]{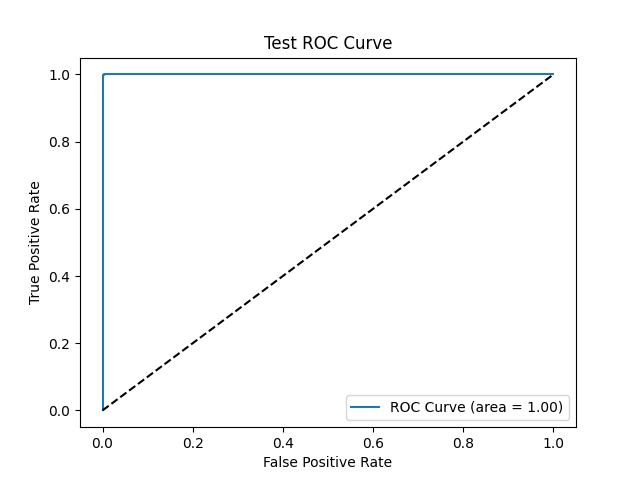}
        \caption{ROC Curve for the test set of CEDAR}
        \label{fig:cedar_roc}
    \end{subfigure}
    \caption{CG - Evaluation metrics for the CEDAR test set}
    \label{fig:cedar_results}
\end{figure}

\begin{figure}[H]
    \centering
    \begin{subfigure}{0.49\textwidth}
        \centering
        \includegraphics[width=\linewidth]{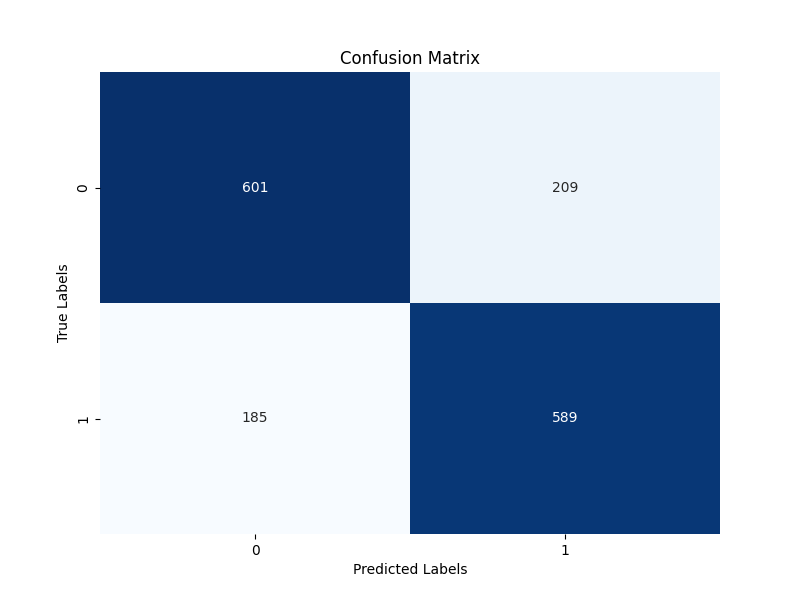}
        \caption{Confusion Matrix for the test set of ICDAR}
        \label{fig:icdar_c}
    \end{subfigure}
    \hfill
    \begin{subfigure}{0.49\textwidth}
        \centering
        \includegraphics[width=\linewidth]{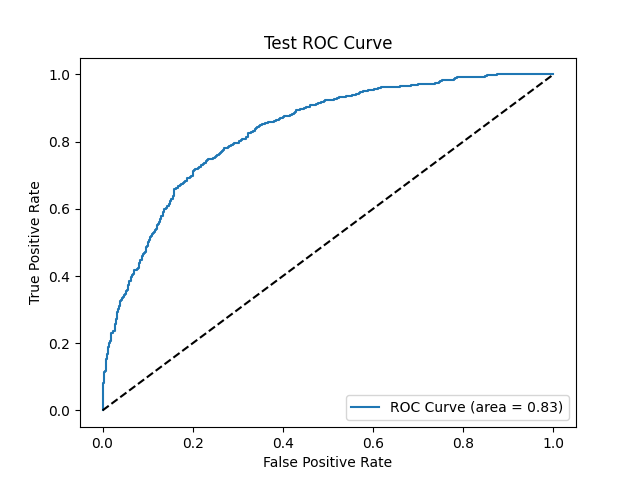}
        \caption{ROC Curve for the test set of ICDAR}
        \label{fig:icdar_roc}
    \end{subfigure}
    \caption{CG - Evaluation metrics for the ICDAR test set}
    \label{fig:icdar_results}
\end{figure}

\begin{figure}[H]
    \centering
    \begin{subfigure}{0.49\textwidth}
        \centering
        \includegraphics[width=\linewidth]{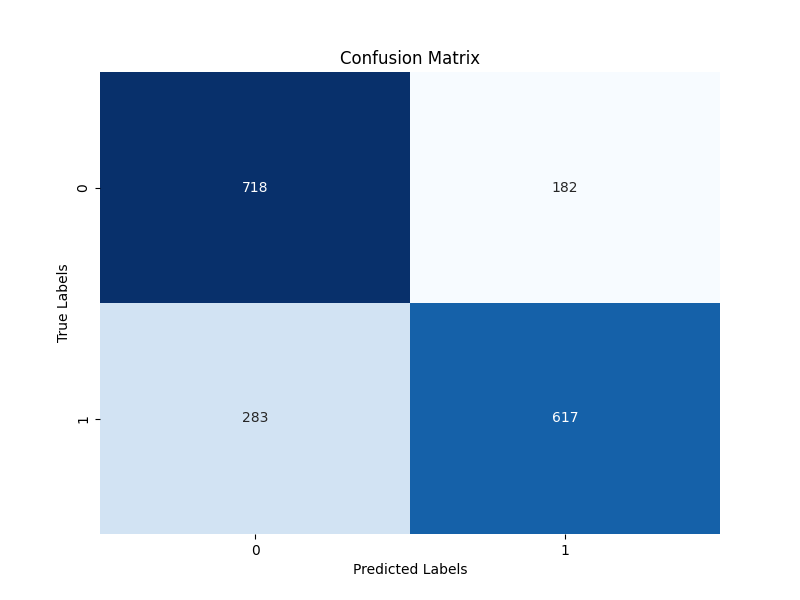}
        \caption{Confusion Matrix for the test set of GPDS}
        \label{fig:gpds_c}
    \end{subfigure}
    \hfill
    \begin{subfigure}{0.49\textwidth}
        \centering
        \includegraphics[width=\linewidth]{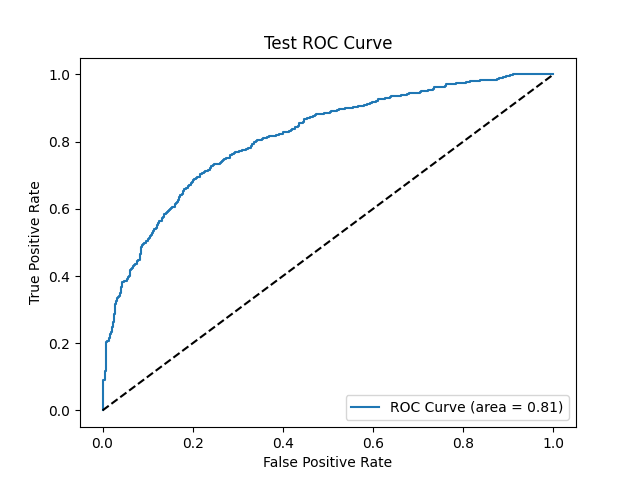}
        \caption{ROC Curve for the test set of GPDS}
        \label{fig:gpds_roc}
    \end{subfigure}
    \caption{CG - Evaluation metrics for the GPDS test set}
    \label{fig:gpds_results}
\end{figure}

\begin{figure}[H]
    \centering
    \begin{subfigure}{0.49\textwidth}
        \centering
        \includegraphics[width=\linewidth]{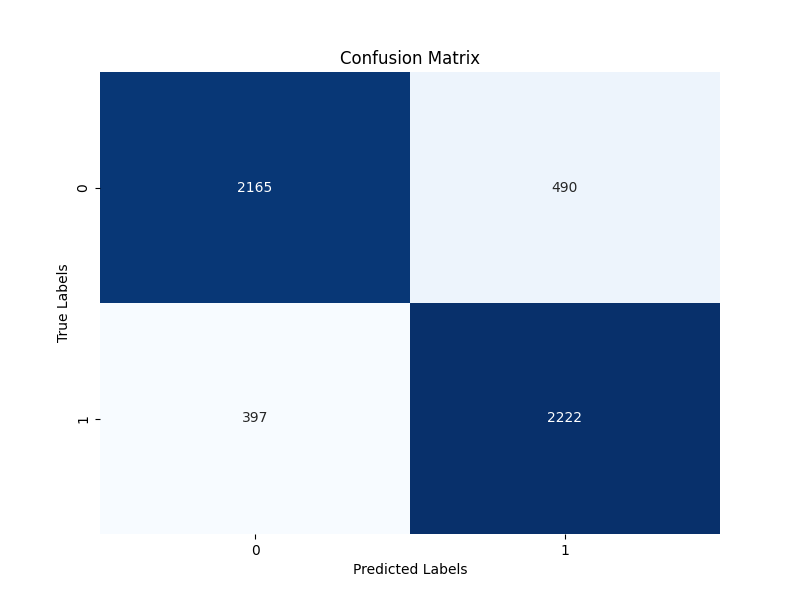}
        \caption{Confusion Matrix for the test set of MERGED}
        \label{fig:merged_c}
    \end{subfigure}
    \hfill
    \begin{subfigure}{0.49\textwidth}
        \centering
        \includegraphics[width=\linewidth]{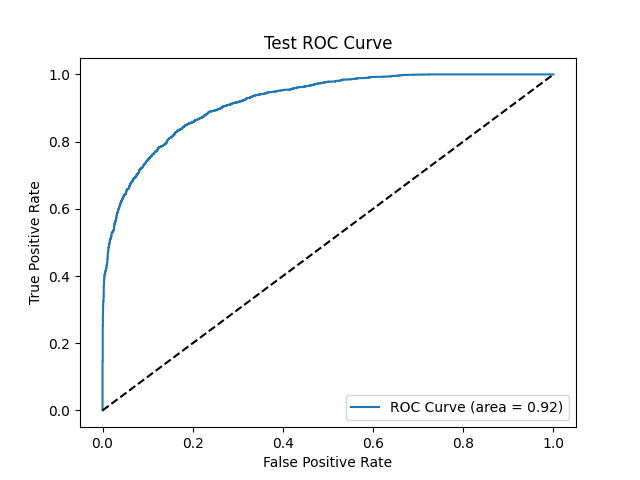}
        \caption{ROC Curve for the test set of MERGED}
        \label{fig:merged_roc}
    \end{subfigure}
    \caption{CG - Evaluation metrics for the MERGED test set}
    \label{fig:merged_results}
\end{figure}

\subsubsection{ICDAR + GPDS Training}
\begin{figure}[H]
    \centering
    \begin{subfigure}{0.49\textwidth}
        \centering
        \includegraphics[width=\linewidth]{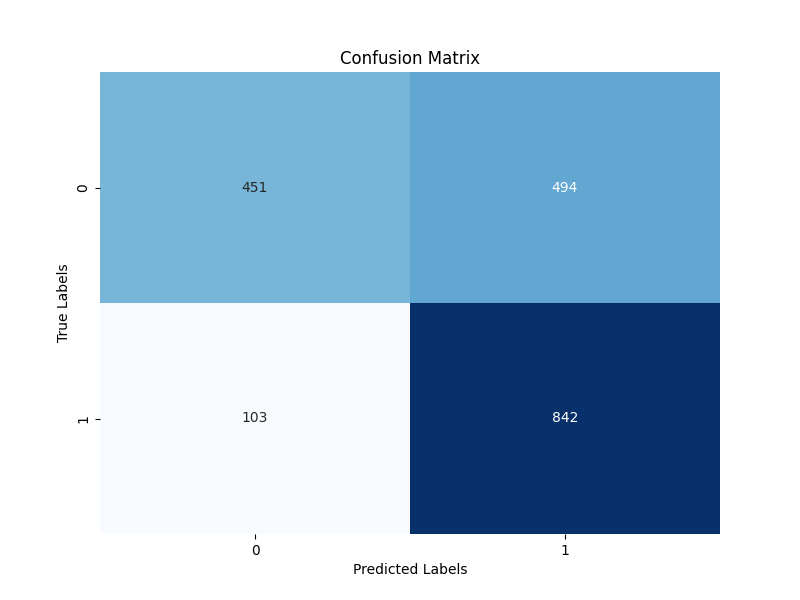}
        \caption{Confusion Matrix for the test set of CEDAR}
        \label{fig:cedar_c}
    \end{subfigure}
    \hfill
    \begin{subfigure}{0.49\textwidth}
        \centering
        \includegraphics[width=\linewidth]{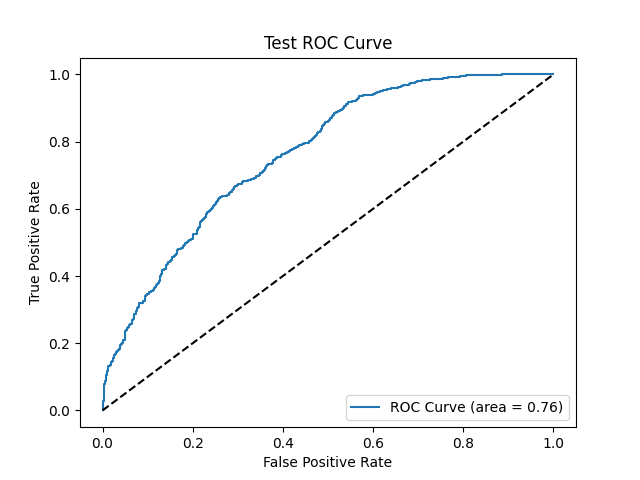}
        \caption{ROC Curve for the test set of CEDAR}
        \label{fig:cedar_roc}
    \end{subfigure}
    \caption{IG - Evaluation metrics for the CEDAR test set}
    \label{fig:cedar_results}
\end{figure}

\begin{figure}[H]
    \centering
    \begin{subfigure}{0.49\textwidth}
        \centering
        \includegraphics[width=\linewidth]{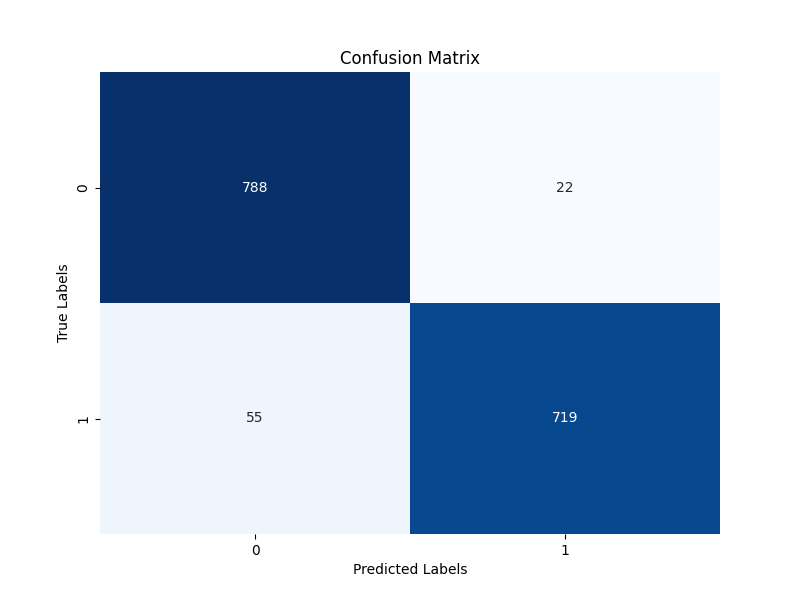}
        \caption{Confusion Matrix for the test set of ICDAR}
        \label{fig:icdar_c}
    \end{subfigure}
    \hfill
    \begin{subfigure}{0.49\textwidth}
        \centering
        \includegraphics[width=\linewidth]{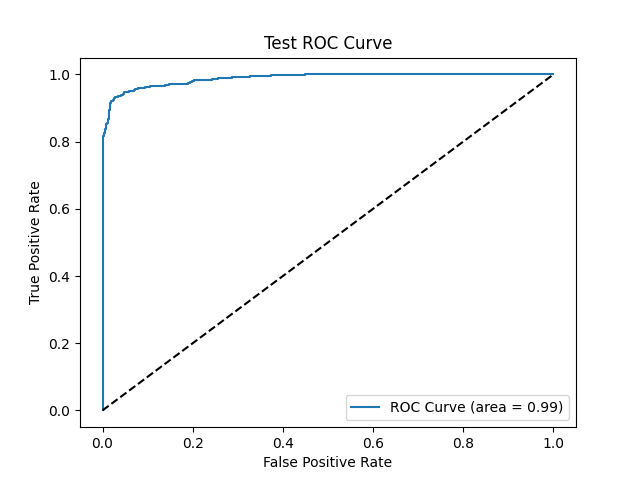}
        \caption{ROC Curve for the test set of ICDAR}
        \label{fig:icdar_roc}
    \end{subfigure}
    \caption{IG - Evaluation metrics for the ICDAR test set}
    \label{fig:icdar_results}
\end{figure}

\begin{figure}[H]
    \centering
    \begin{subfigure}{0.49\textwidth}
        \centering
        \includegraphics[width=\linewidth]{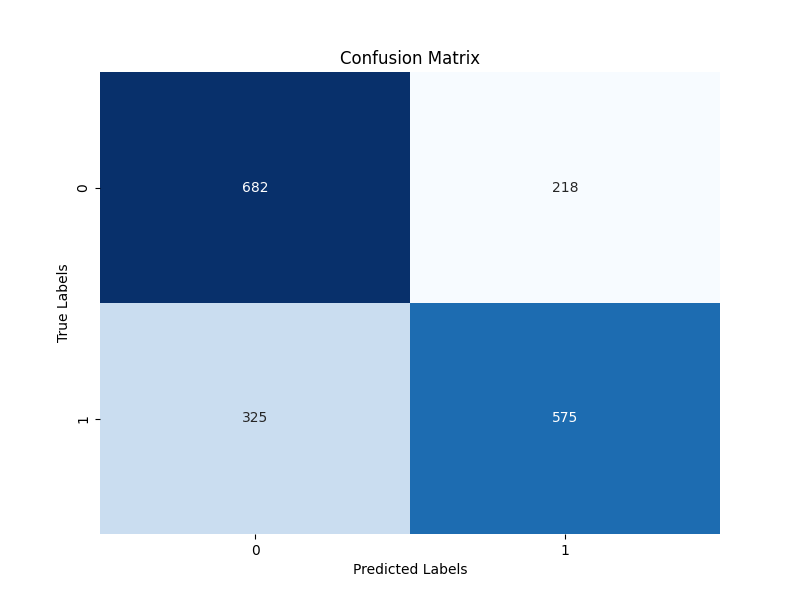}
        \caption{Confusion Matrix for the test set of GPDS}
        \label{fig:gpds_c}
    \end{subfigure}
    \hfill
    \begin{subfigure}{0.49\textwidth}
        \centering
        \includegraphics[width=\linewidth]{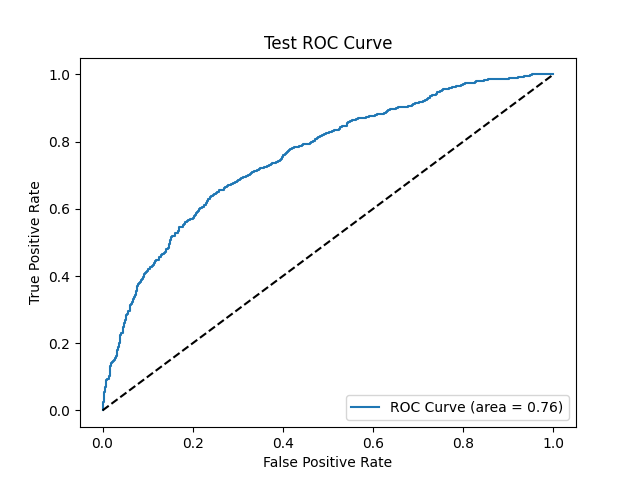}
        \caption{ROC Curve for the test set of GPDS}
        \label{fig:gpds_roc}
    \end{subfigure}
    \caption{IG - Evaluation metrics for the GPDS test set}
    \label{fig:gpds_results}
\end{figure}

\begin{figure}[H]
    \centering
    \begin{subfigure}{0.49\textwidth}
        \centering
        \includegraphics[width=\linewidth]{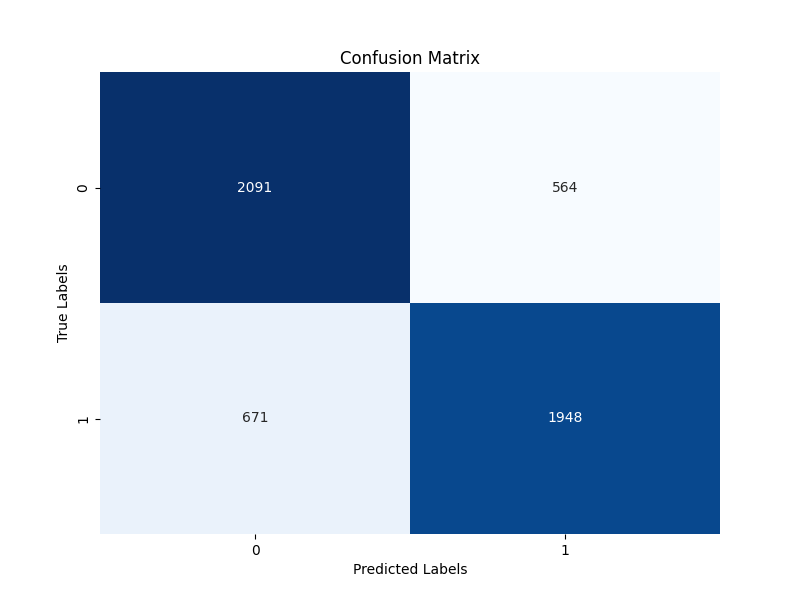}
        \caption{Confusion Matrix for the test set of MERGED}
        \label{fig:merged_c}
    \end{subfigure}
    \hfill
    \begin{subfigure}{0.49\textwidth}
        \centering
        \includegraphics[width=\linewidth]{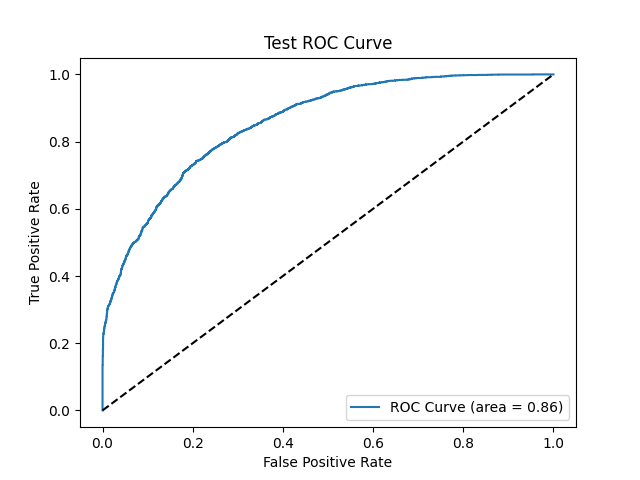}
        \caption{ROC Curve for the test set of MERGED}
        \label{fig:merged_roc}
    \end{subfigure}
    \caption{IG - Evaluation metrics for the MERGED test set}
    \label{fig:merged_results}
\end{figure}

\subsection{Triplet Loss Training}

\subsubsection{CEDAR + ICDAR Training}
\begin{figure}[H]
    \centering
    \begin{subfigure}{0.49\textwidth}
        \centering
        \includegraphics[width=\linewidth]{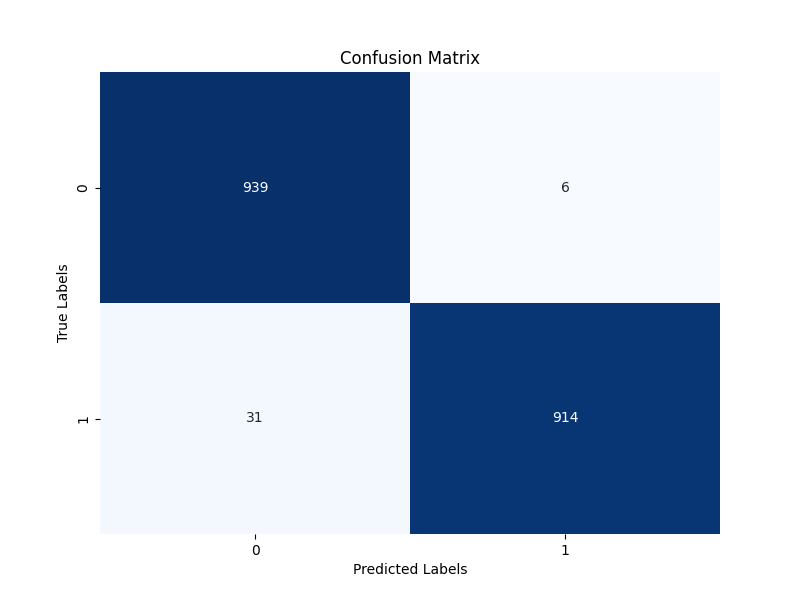}
        \caption{Confusion Matrix for the test set of CEDAR}
        \label{fig:cedar_c}
    \end{subfigure}
    \hfill
    \begin{subfigure}{0.49\textwidth}
        \centering
        \includegraphics[width=\linewidth]{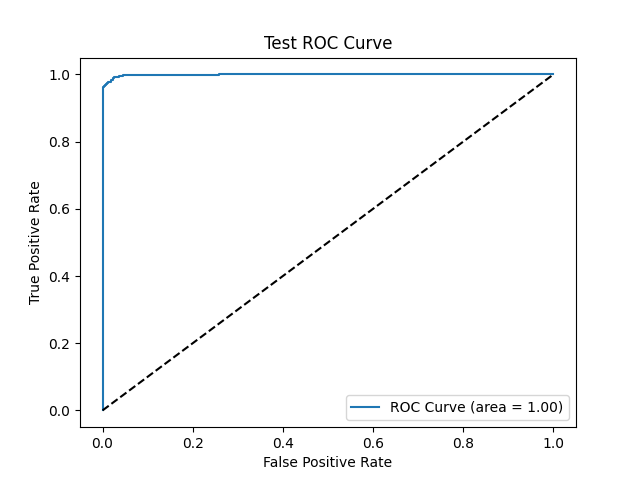}
        \caption{ROC Curve for the test set of CEDAR}
        \label{fig:cedar_roc}
    \end{subfigure}
    \caption{CI - Evaluation metrics for the CEDAR test set}
    \label{fig:cedar_results}
\end{figure}

\begin{figure}[H]
    \centering
    \begin{subfigure}{0.49\textwidth}
        \centering
        \includegraphics[width=\linewidth]{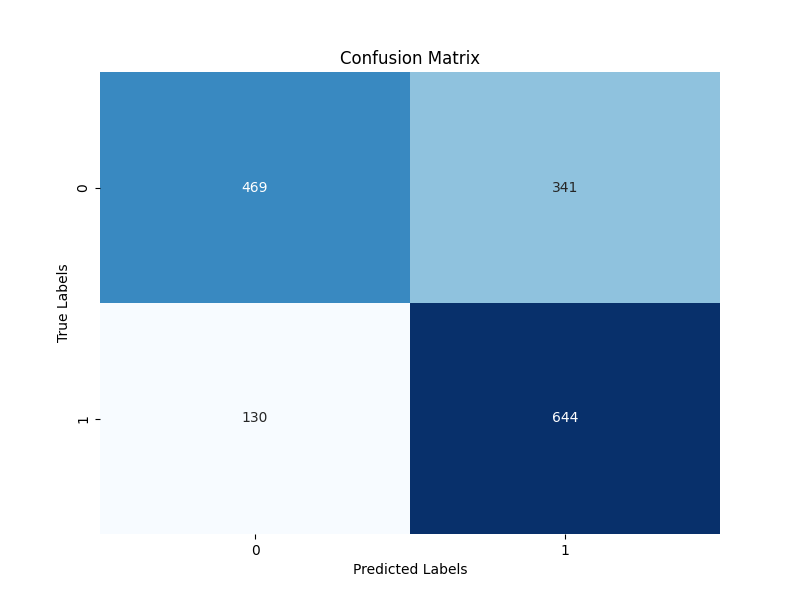}
        \caption{Confusion Matrix for the test set of ICDAR}
        \label{fig:icdar_c}
    \end{subfigure}
    \hfill
    \begin{subfigure}{0.49\textwidth}
        \centering
        \includegraphics[width=\linewidth]{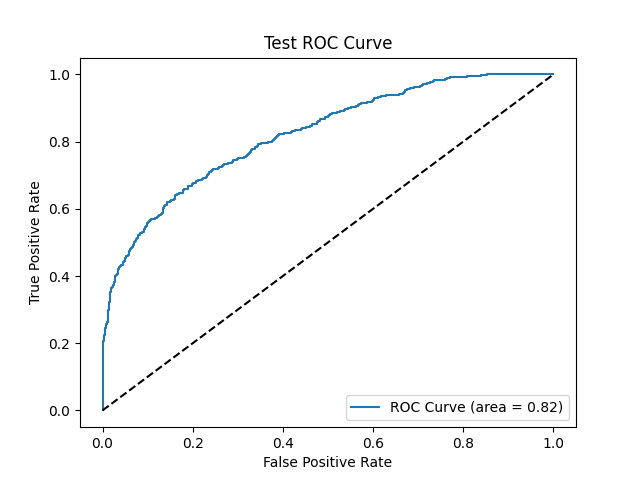}
        \caption{ROC Curve for the test set of ICDAR}
        \label{fig:icdar_roc}
    \end{subfigure}
    \caption{CI - Evaluation metrics for the ICDAR test set}
    \label{fig:icdar_results}
\end{figure}

\begin{figure}[H]
    \centering
    \begin{subfigure}{0.49\textwidth}
        \centering
        \includegraphics[width=\linewidth]{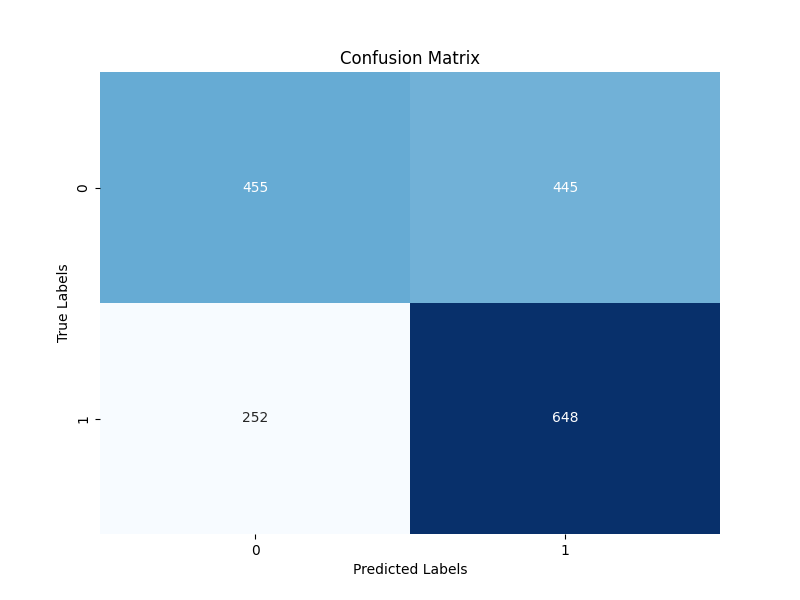}
        \caption{Confusion Matrix for the test set of GPDS}
        \label{fig:gpds_c}
    \end{subfigure}
    \hfill
    \begin{subfigure}{0.49\textwidth}
        \centering
        \includegraphics[width=\linewidth]{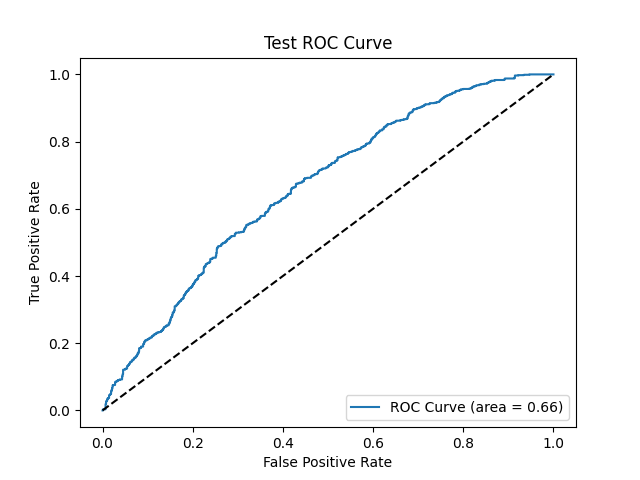}
        \caption{ROC Curve for the test set of GPDS}
        \label{fig:gpds_roc}
    \end{subfigure}
    \caption{CI - Evaluation metrics for the GPDS test set}
    \label{fig:gpds_results}
\end{figure}

\begin{figure}[H]
    \centering
    \begin{subfigure}{0.49\textwidth}
        \centering
        \includegraphics[width=\linewidth]{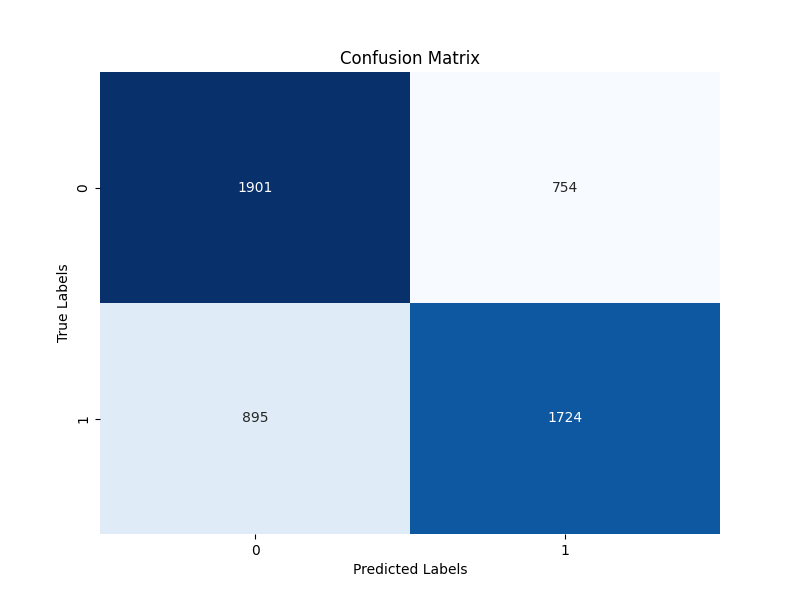}
        \caption{Confusion Matrix for the test set of MERGED}
        \label{fig:merged_c}
    \end{subfigure}
    \hfill
    \begin{subfigure}{0.49\textwidth}
        \centering
        \includegraphics[width=\linewidth]{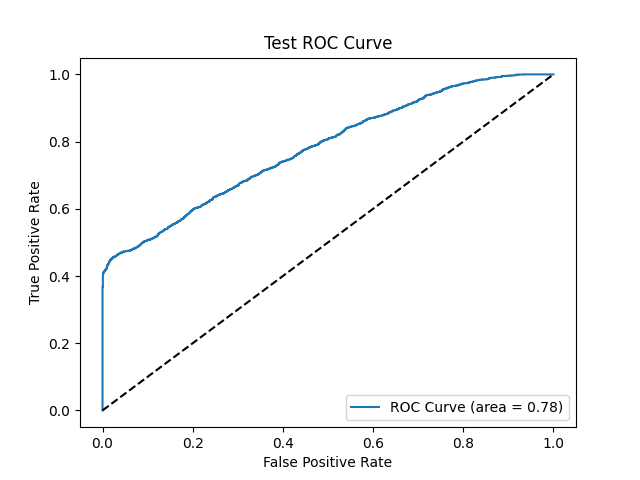}
        \caption{ROC Curve for the test set of MERGED}
        \label{fig:merged_roc}
    \end{subfigure}
    \caption{CI - Evaluation metrics for the MERGED test set}
    \label{fig:merged_results}
\end{figure}

\subsubsection{CEDAR + GPDS Training}
\begin{figure}[H]
    \centering
    \begin{subfigure}{0.49\textwidth}
        \centering
        \includegraphics[width=\linewidth]{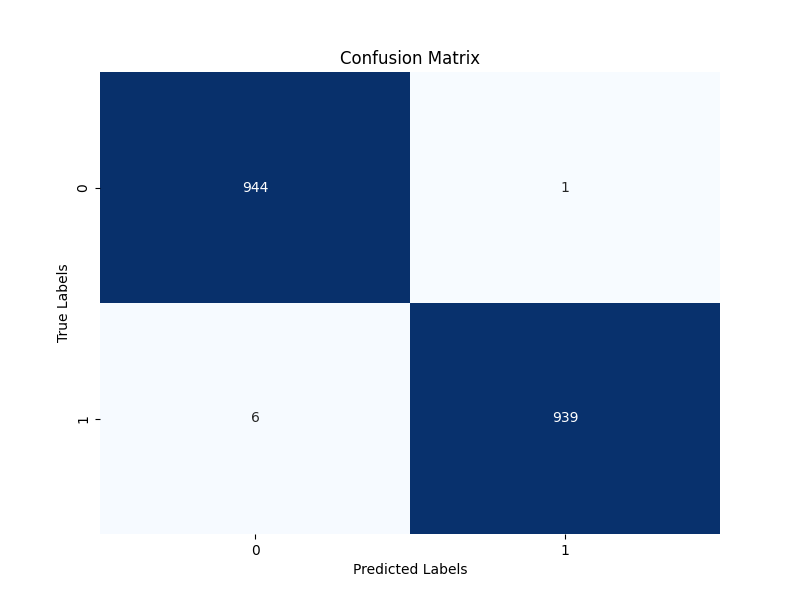}
        \caption{Confusion Matrix for the test set of CEDAR}
        \label{fig:cedar_c}
    \end{subfigure}
    \hfill
    \begin{subfigure}{0.49\textwidth}
        \centering
        \includegraphics[width=\linewidth]{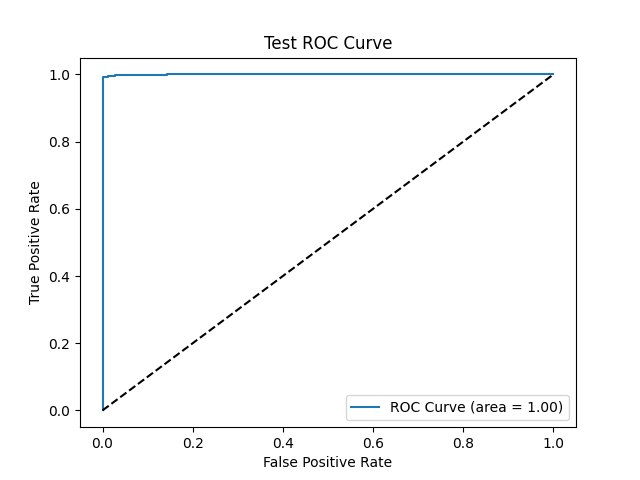}
        \caption{ROC Curve for the test set of CEDAR}
        \label{fig:cedar_roc}
    \end{subfigure}
    \caption{CG - Evaluation metrics for the CEDAR test set}
    \label{fig:cedar_results}
\end{figure}

\begin{figure}[H]
    \centering
    \begin{subfigure}{0.49\textwidth}
        \centering
        \includegraphics[width=\linewidth]{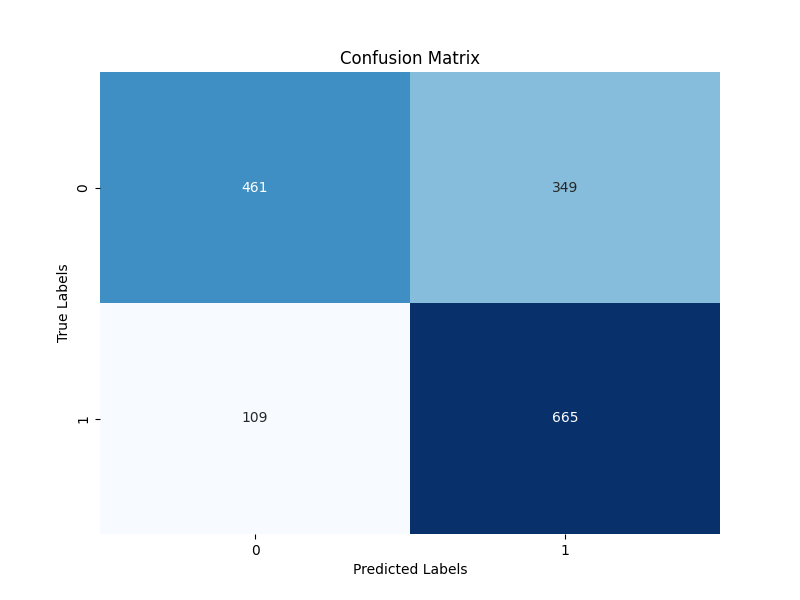}
        \caption{Confusion Matrix for the test set of ICDAR}
        \label{fig:icdar_c}
    \end{subfigure}
    \hfill
    \begin{subfigure}{0.49\textwidth}
        \centering
        \includegraphics[width=\linewidth]{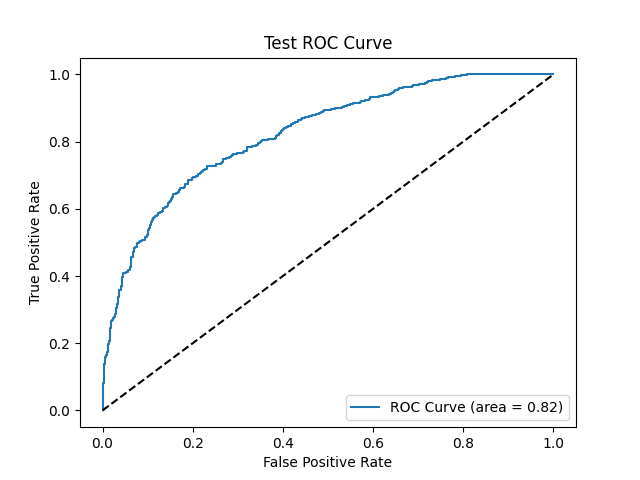}
        \caption{ROC Curve for the test set of ICDAR}
        \label{fig:icdar_roc}
    \end{subfigure}
    \caption{CG - Evaluation metrics for the ICDAR test set}
    \label{fig:icdar_results}
\end{figure}

\begin{figure}[H]
    \centering
    \begin{subfigure}{0.49\textwidth}
        \centering
        \includegraphics[width=\linewidth]{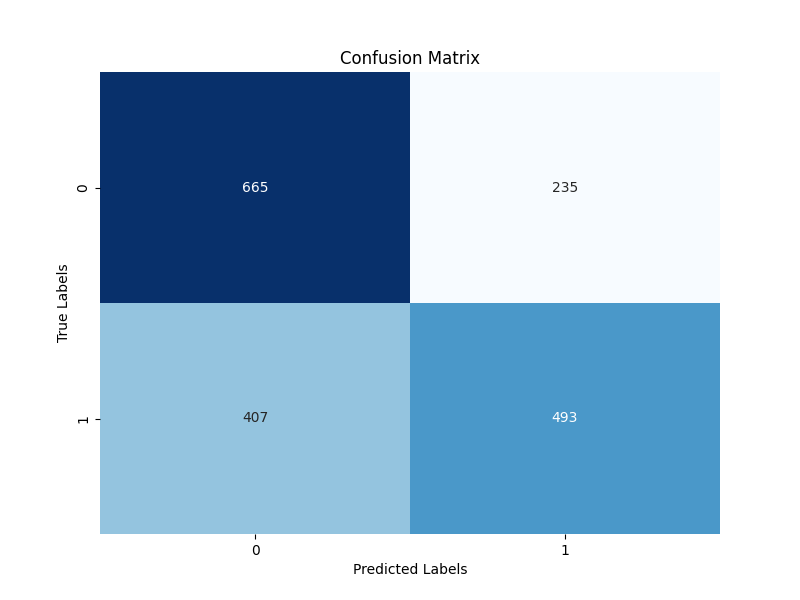}
        \caption{Confusion Matrix for the test set of GPDS}
        \label{fig:gpds_c}
    \end{subfigure}
    \hfill
    \begin{subfigure}{0.49\textwidth}
        \centering
        \includegraphics[width=\linewidth]{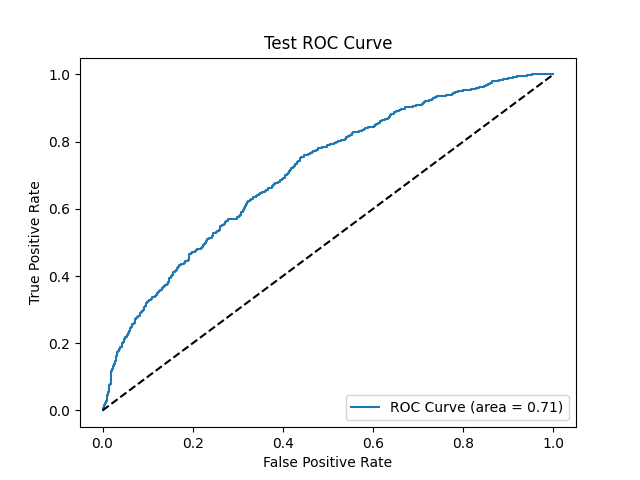}
        \caption{ROC Curve for the test set of GPDS}
        \label{fig:gpds_roc}
    \end{subfigure}
    \caption{CG - Evaluation metrics for the GPDS test set}
    \label{fig:gpds_results}
\end{figure}

\begin{figure}[H]
    \centering
    \begin{subfigure}{0.49\textwidth}
        \centering
        \includegraphics[width=\linewidth]{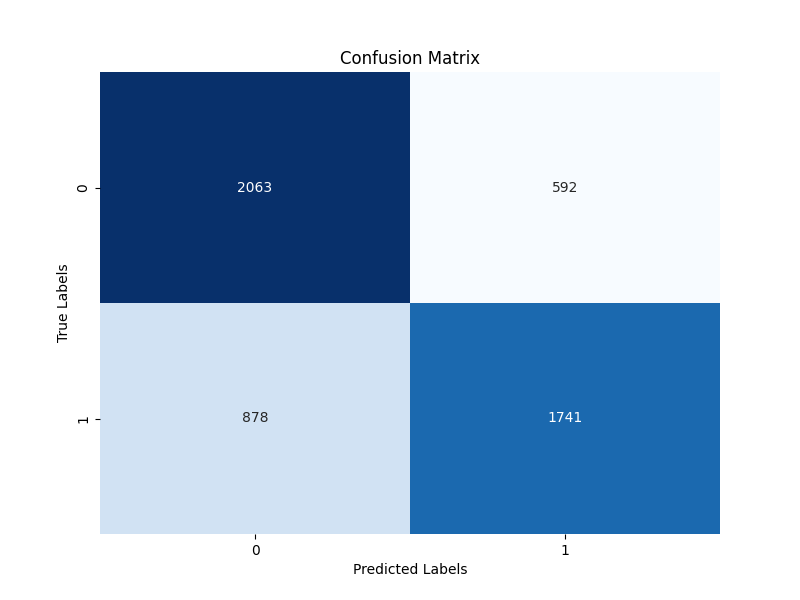}
        \caption{Confusion Matrix for the test set of MERGED}
        \label{fig:merged_c}
    \end{subfigure}
    \hfill
    \begin{subfigure}{0.49\textwidth}
        \centering
        \includegraphics[width=\linewidth]{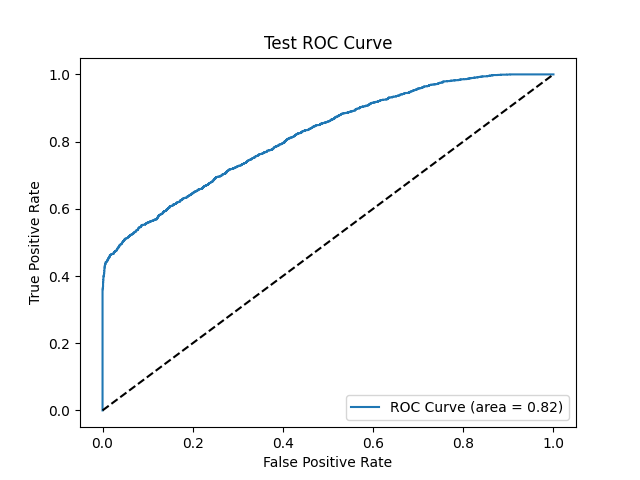}
        \caption{ROC Curve for the test set of MERGED}
        \label{fig:merged_roc}
    \end{subfigure}
    \caption{CG - Evaluation metrics for the MERGED test set}
    \label{fig:merged_results}
\end{figure}

\subsubsection{ICDAR + GPDS Training}
\begin{figure}[H]
    \centering
    \begin{subfigure}{0.49\textwidth}
        \centering
        \includegraphics[width=\linewidth]{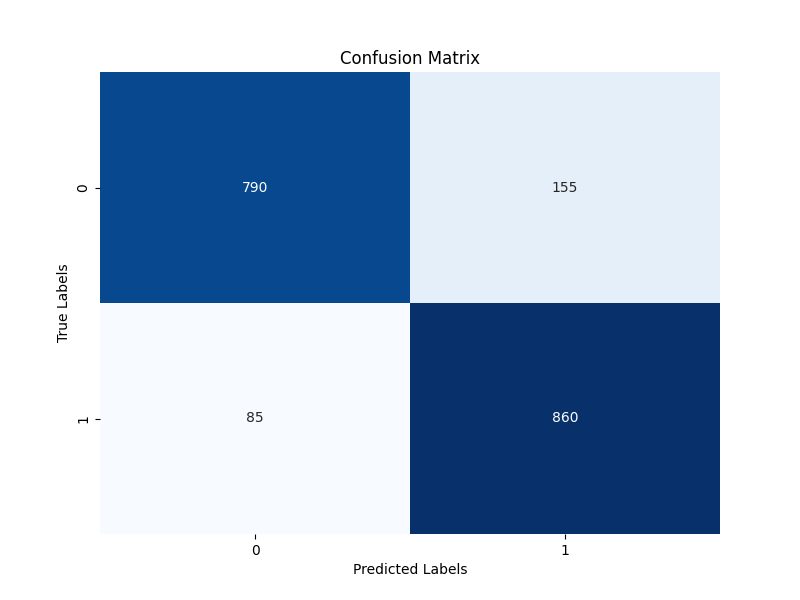}
        \caption{Confusion Matrix for the test set of CEDAR}
        \label{fig:cedar_c}
    \end{subfigure}
    \hfill
    \begin{subfigure}{0.49\textwidth}
        \centering
        \includegraphics[width=\linewidth]{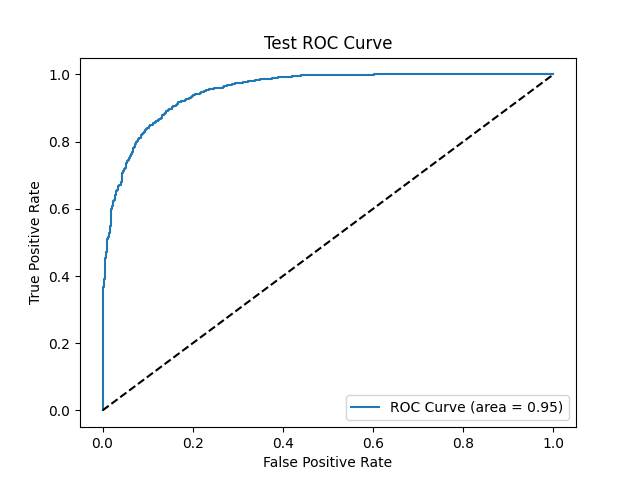}
        \caption{ROC Curve for the test set of CEDAR}
        \label{fig:cedar_roc}
    \end{subfigure}
    \caption{IG - Evaluation metrics for the CEDAR test set}
    \label{fig:cedar_results}
\end{figure}

\begin{figure}[H]
    \centering
    \begin{subfigure}{0.49\textwidth}
        \centering
        \includegraphics[width=\linewidth]{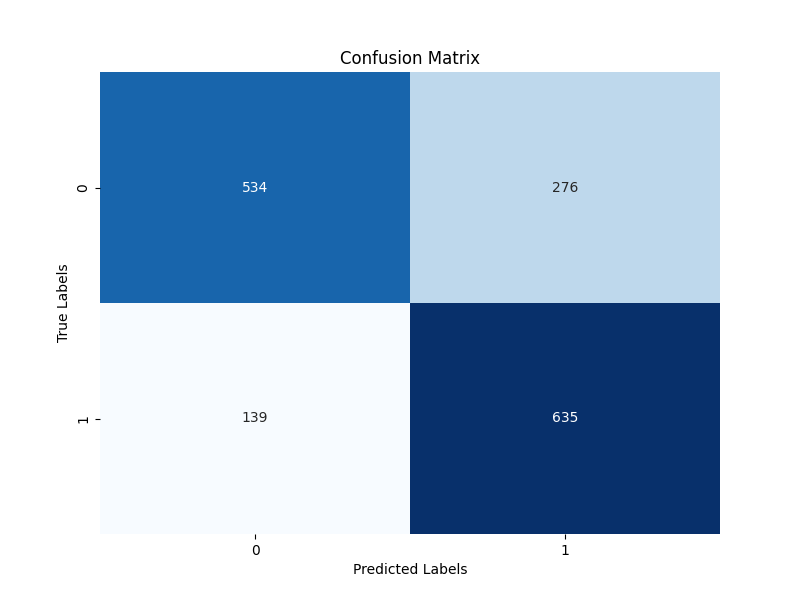}
        \caption{Confusion Matrix for the test set of ICDAR}
        \label{fig:icdar_c}
    \end{subfigure}
    \hfill
    \begin{subfigure}{0.49\textwidth}
        \centering
        \includegraphics[width=\linewidth]{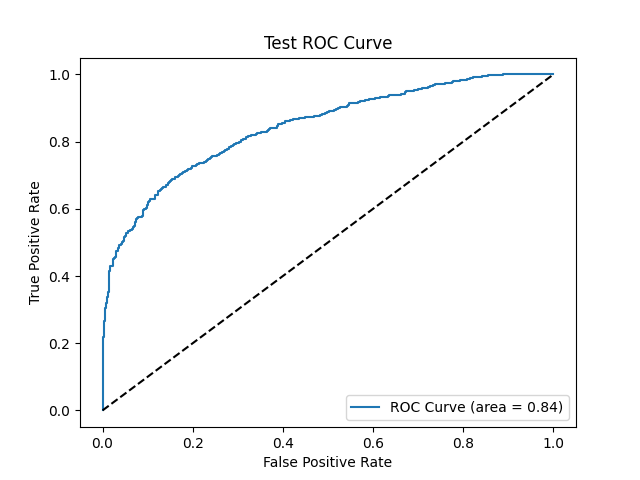}
        \caption{ROC Curve for the test set of ICDAR}
        \label{fig:icdar_roc}
    \end{subfigure}
    \caption{IG - Evaluation metrics for the ICDAR test set}
    \label{fig:icdar_results}
\end{figure}

\begin{figure}[H]
    \centering
    \begin{subfigure}{0.49\textwidth}
        \centering
        \includegraphics[width=\linewidth]{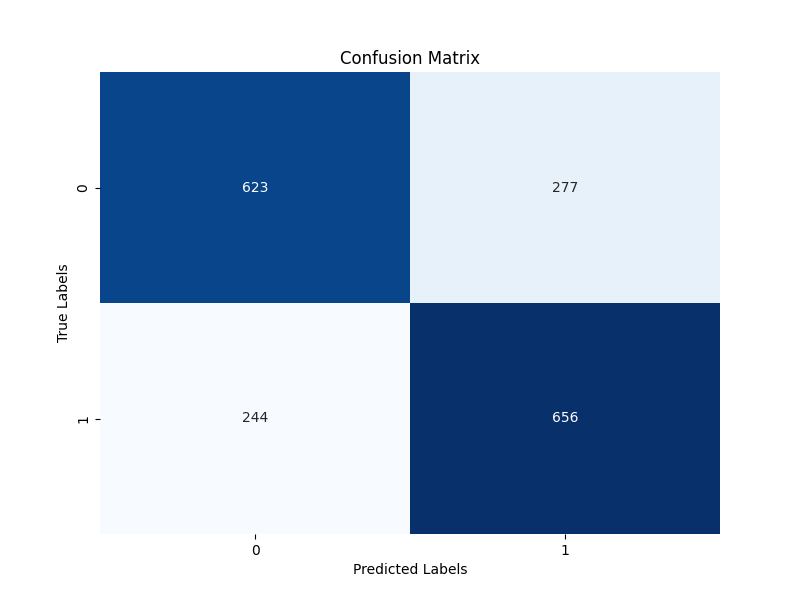}
        \caption{Confusion Matrix for the test set of GPDS}
        \label{fig:gpds_c}
    \end{subfigure}
    \hfill
    \begin{subfigure}{0.49\textwidth}
        \centering
        \includegraphics[width=\linewidth]{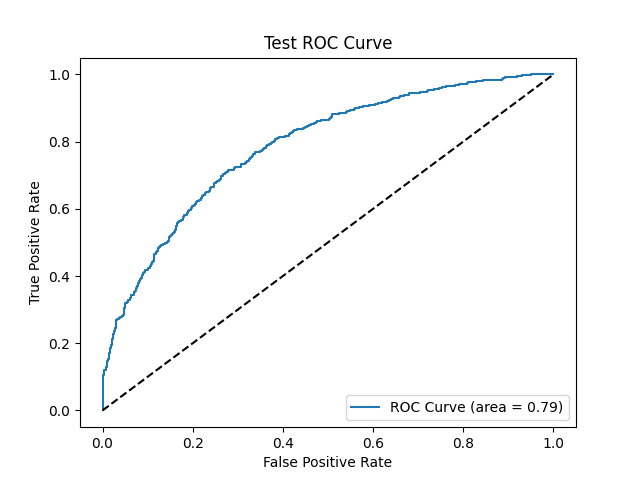}
        \caption{ROC Curve for the test set of GPDS}
        \label{fig:gpds_roc}
    \end{subfigure}
    \caption{IG - Evaluation metrics for the GPDS test set}
    \label{fig:gpds_results}
\end{figure}

\begin{figure}[H]
    \centering
    \begin{subfigure}{0.49\textwidth}
        \centering
        \includegraphics[width=\linewidth]{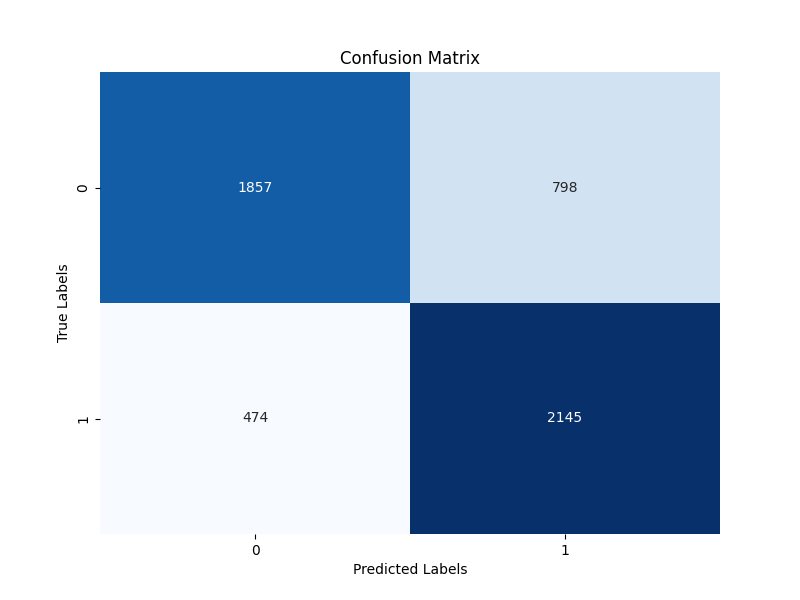}
        \caption{Confusion Matrix for the test set of MERGED}
        \label{fig:merged_c}
    \end{subfigure}
    \hfill
    \begin{subfigure}{0.49\textwidth}
        \centering
        \includegraphics[width=\linewidth]{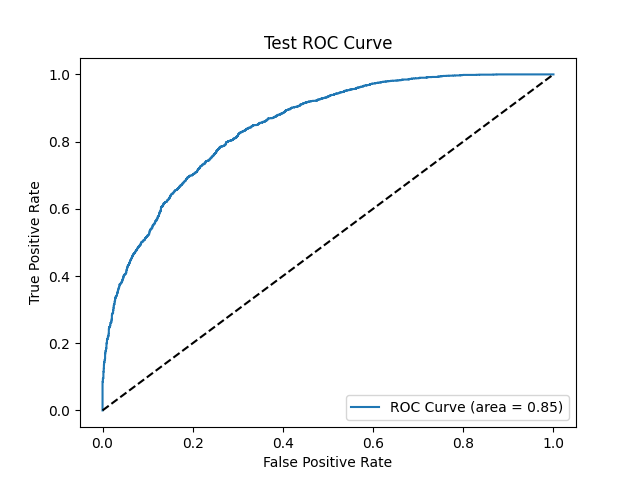}
        \caption{ROC Curve for the test set of MERGED}
        \label{fig:merged_roc}
    \end{subfigure}
    \caption{IG - Evaluation metrics for the MERGED test set}
    \label{fig:merged_results}
\end{figure}

\section{PART 2}

\subsection{Contrastive Loss Training}

\subsubsection{CEDAR + ICDAR Training}
\begin{figure}[H]
    \centering
    \begin{subfigure}{0.49\textwidth}
        \centering
        \includegraphics[width=\linewidth]{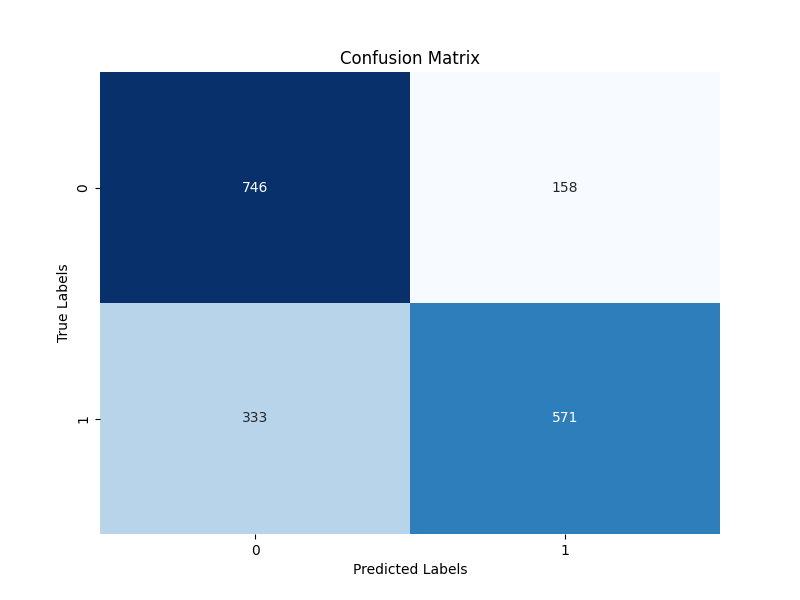}
        \caption{Confusion Matrix for the test set of CEDAR}
        \label{fig:cedar_c}
    \end{subfigure}
    \hfill
    \begin{subfigure}{0.49\textwidth}
        \centering
        \includegraphics[width=\linewidth]{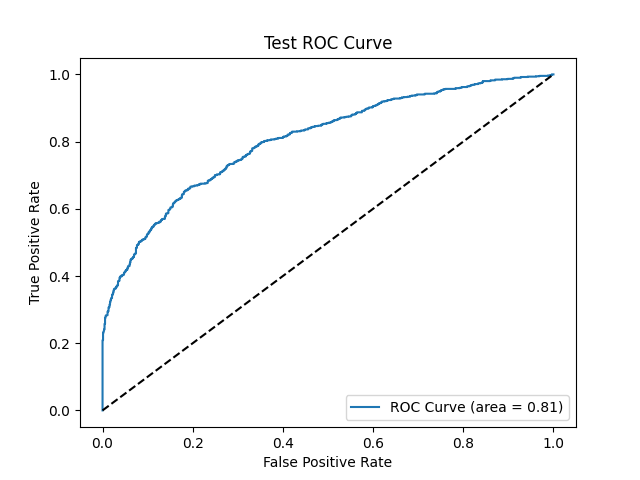}
        \caption{ROC Curve for the test set of CEDAR}
        \label{fig:cedar_roc}
    \end{subfigure}
    \caption{CI - Evaluation metrics for the CEDAR test set}
    \label{fig:cedar_results}
\end{figure}

\begin{figure}[H]
    \centering
    \begin{subfigure}{0.49\textwidth}
        \centering
        \includegraphics[width=\linewidth]{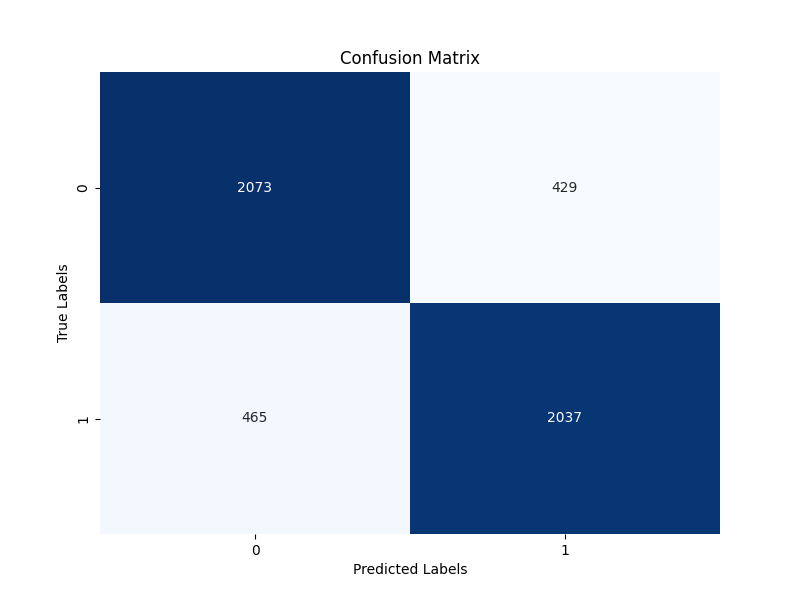}
        \caption{Confusion Matrix for the test set of ICDAR}
        \label{fig:icdar_c}
    \end{subfigure}
    \hfill
    \begin{subfigure}{0.49\textwidth}
        \centering
        \includegraphics[width=\linewidth]{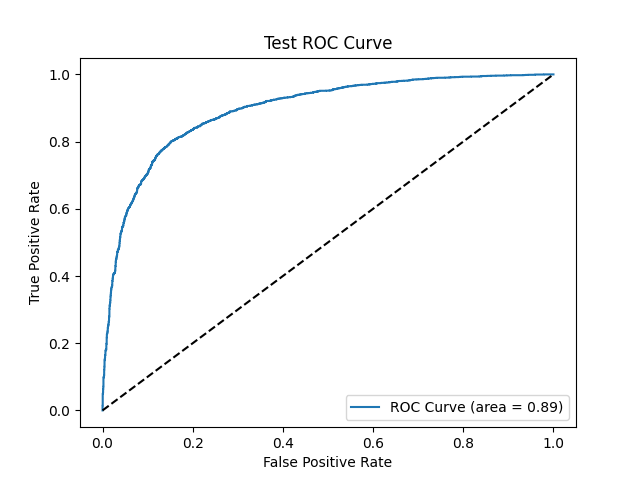}
        \caption{ROC Curve for the test set of ICDAR}
        \label{fig:icdar_roc}
    \end{subfigure}
    \caption{CI - Evaluation metrics for the ICDAR test set}
    \label{fig:icdar_results}
\end{figure}

\begin{figure}[H]
    \centering
    \begin{subfigure}{0.49\textwidth}
        \centering
        \includegraphics[width=\linewidth]{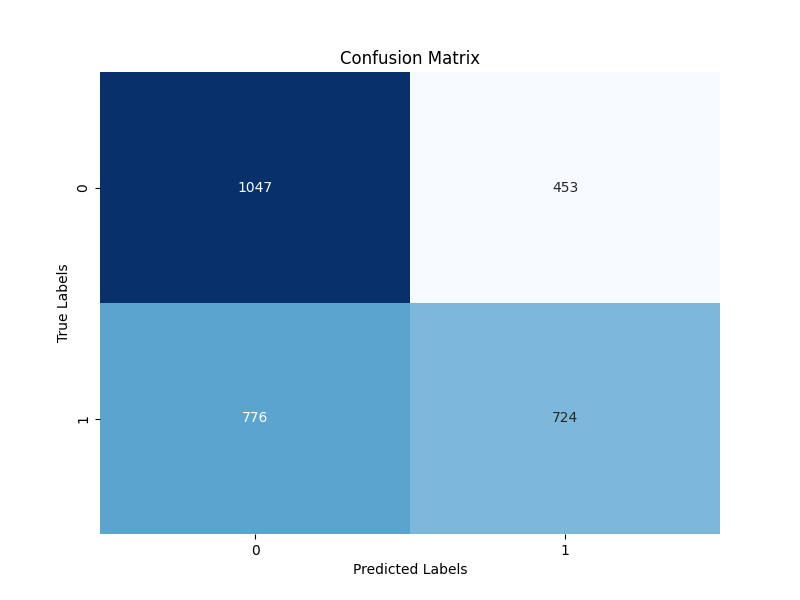}
        \caption{Confusion Matrix for the test set of GPDS}
        \label{fig:gpds_c}
    \end{subfigure}
    \hfill
    \begin{subfigure}{0.49\textwidth}
        \centering
        \includegraphics[width=\linewidth]{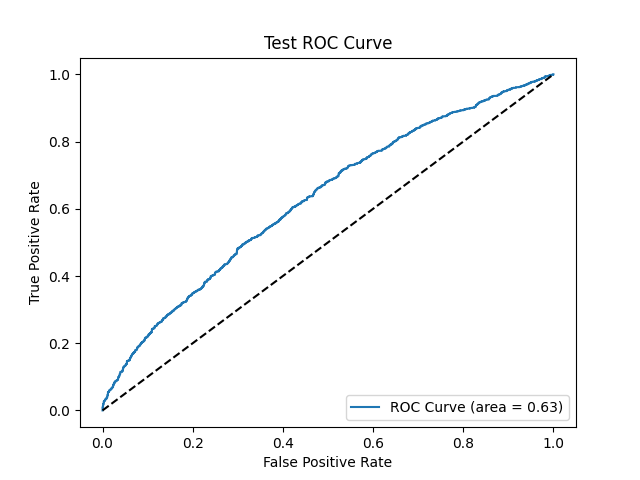}
        \caption{ROC Curve for the test set of GPDS}
        \label{fig:gpds_roc}
    \end{subfigure}
    \caption{CI - Evaluation metrics for the GPDS test set}
    \label{fig:gpds_results}
\end{figure}

\begin{figure}[H]
    \centering
    \begin{subfigure}{0.49\textwidth}
        \centering
        \includegraphics[width=\linewidth]{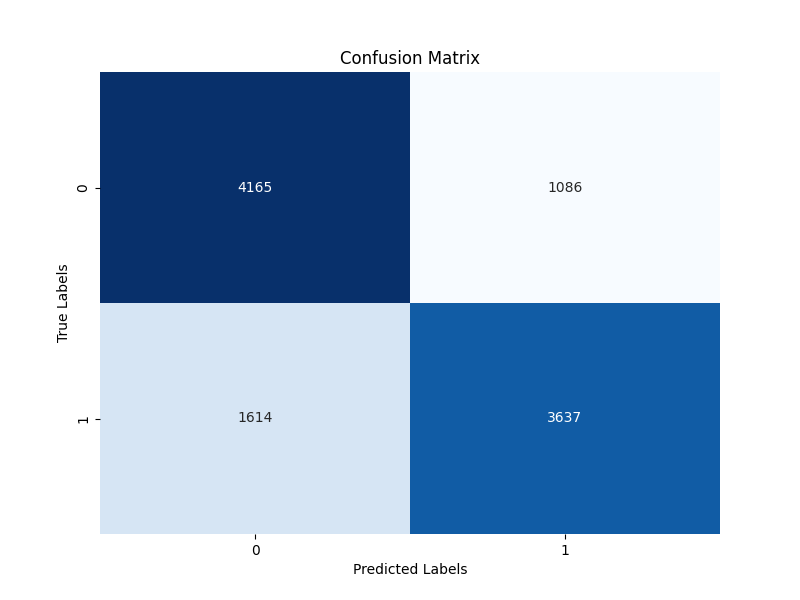}
        \caption{Confusion Matrix for the test set of MERGED}
        \label{fig:merged_c}
    \end{subfigure}
    \hfill
    \begin{subfigure}{0.49\textwidth}
        \centering
        \includegraphics[width=\linewidth]{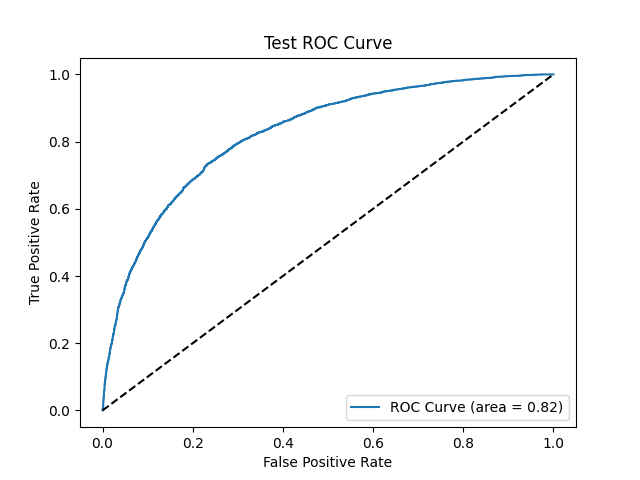}
        \caption{ROC Curve for the test set of MERGED}
        \label{fig:merged_roc}
    \end{subfigure}
    \caption{CI - Evaluation metrics for the MERGED test set}
    \label{fig:merged_results}
\end{figure}

\subsubsection{CEDAR + GPDS Training}
\begin{figure}[H]
    \centering
    \begin{subfigure}{0.49\textwidth}
        \centering
        \includegraphics[width=\linewidth]{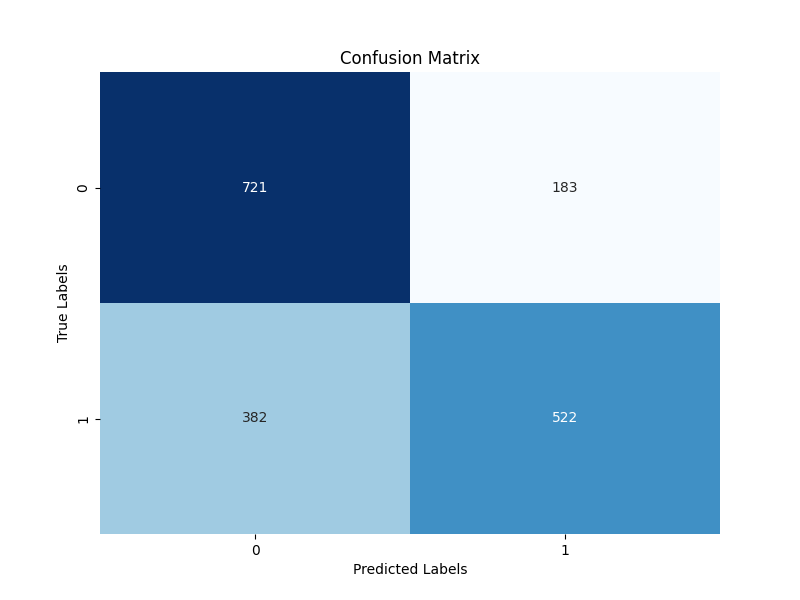}
        \caption{Confusion Matrix for the test set of CEDAR}
        \label{fig:cedar_c}
    \end{subfigure}
    \hfill
    \begin{subfigure}{0.49\textwidth}
        \centering
        \includegraphics[width=\linewidth]{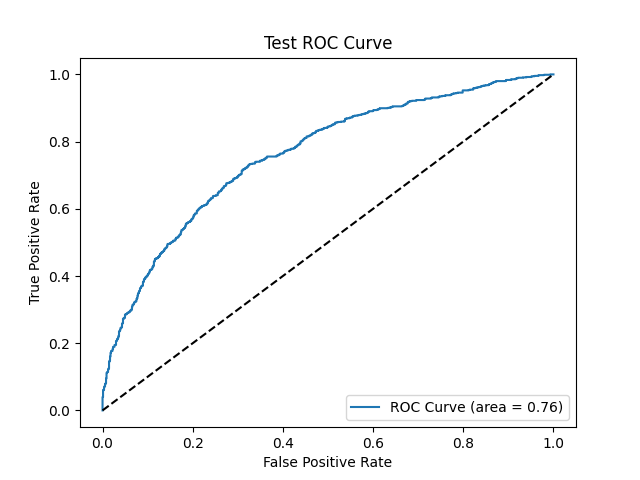}
        \caption{ROC Curve for the test set of CEDAR}
        \label{fig:cedar_roc}
    \end{subfigure}
    \caption{CI - Evaluation metrics for the CEDAR test set}
    \label{fig:cedar_results}
\end{figure}

\begin{figure}[H]
    \centering
    \begin{subfigure}{0.49\textwidth}
        \centering
        \includegraphics[width=\linewidth]{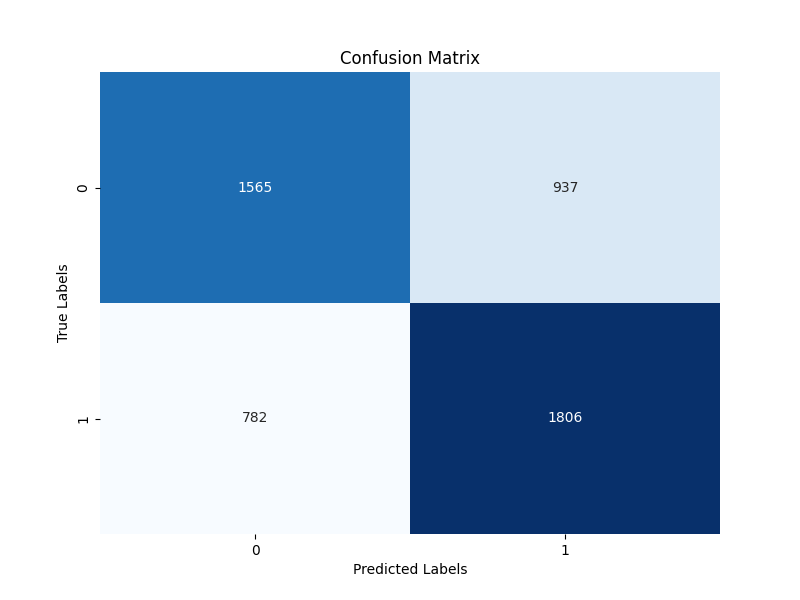}
        \caption{Confusion Matrix for the test set of ICDAR}
        \label{fig:icdar_c}
    \end{subfigure}
    \hfill
    \begin{subfigure}{0.49\textwidth}
        \centering
        \includegraphics[width=\linewidth]{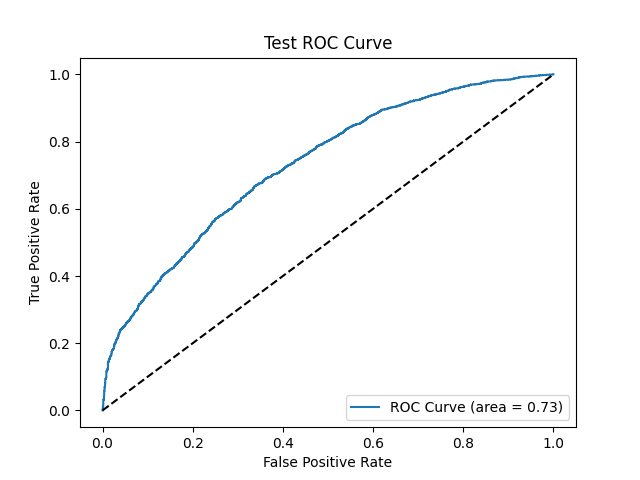}
        \caption{ROC Curve for the test set of ICDAR}
        \label{fig:icdar_roc}
    \end{subfigure}
    \caption{CI - Evaluation metrics for the ICDAR test set}
    \label{fig:icdar_results}
\end{figure}

\begin{figure}[H]
    \centering
    \begin{subfigure}{0.49\textwidth}
        \centering
        \includegraphics[width=\linewidth]{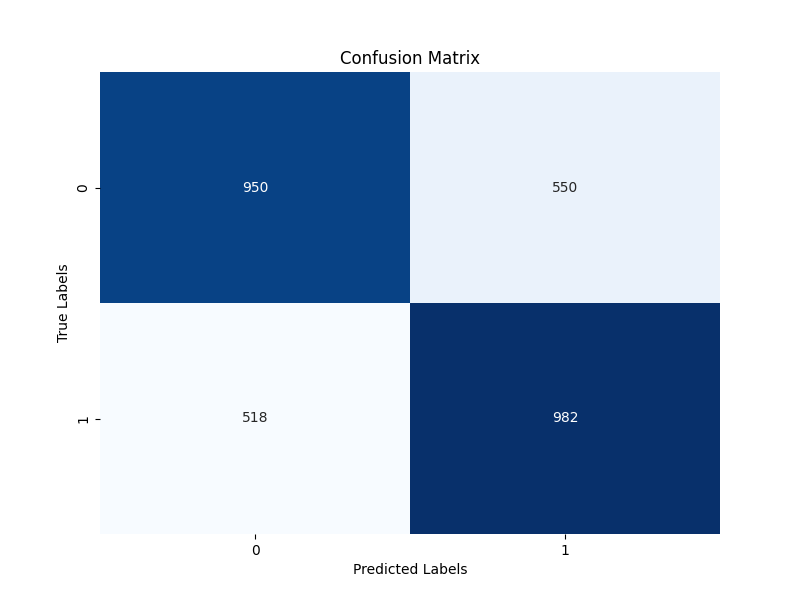}
        \caption{Confusion Matrix for the test set of GPDS}
        \label{fig:gpds_c}
    \end{subfigure}
    \hfill
    \begin{subfigure}{0.49\textwidth}
        \centering
        \includegraphics[width=\linewidth]{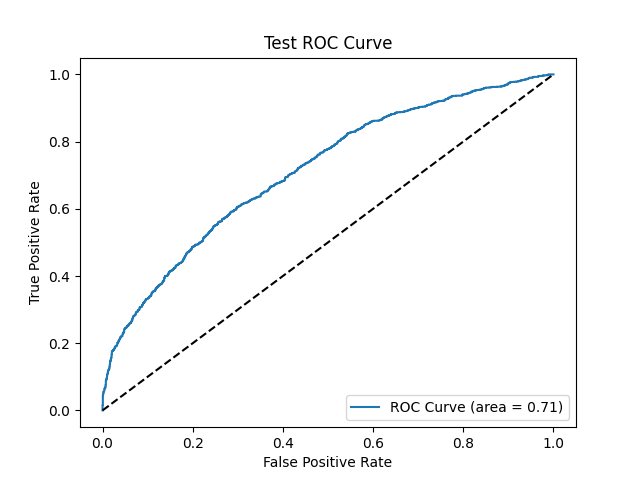}
        \caption{ROC Curve for the test set of GPDS}
        \label{fig:gpds_roc}
    \end{subfigure}
    \caption{CI - Evaluation metrics for the GPDS test set}
    \label{fig:gpds_results}
\end{figure}

\begin{figure}[H]
    \centering
    \begin{subfigure}{0.49\textwidth}
        \centering
        \includegraphics[width=\linewidth]{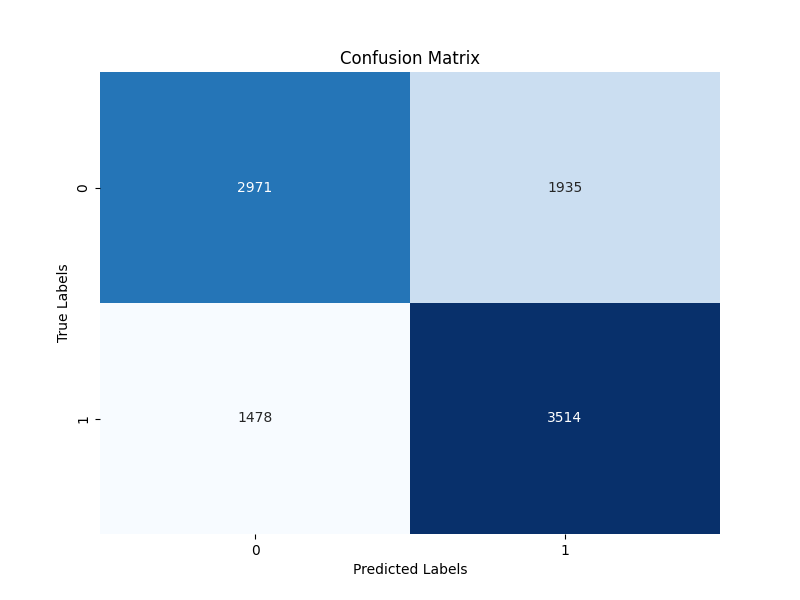}
        \caption{Confusion Matrix for the test set of MERGED}
        \label{fig:merged_c}
    \end{subfigure}
    \hfill
    \begin{subfigure}{0.49\textwidth}
        \centering
        \includegraphics[width=\linewidth]{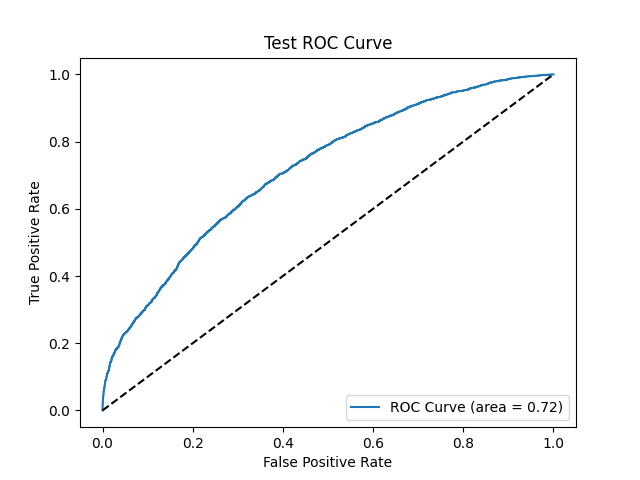}
        \caption{ROC Curve for the test set of MERGED}
        \label{fig:merged_roc}
    \end{subfigure}
    \caption{CI - Evaluation metrics for the MERGED test set}
    \label{fig:merged_results}
\end{figure}

\subsubsection{ICDAR + GPDS Training}
\begin{figure}[H]
    \centering
    \begin{subfigure}{0.49\textwidth}
        \centering
        \includegraphics[width=\linewidth]{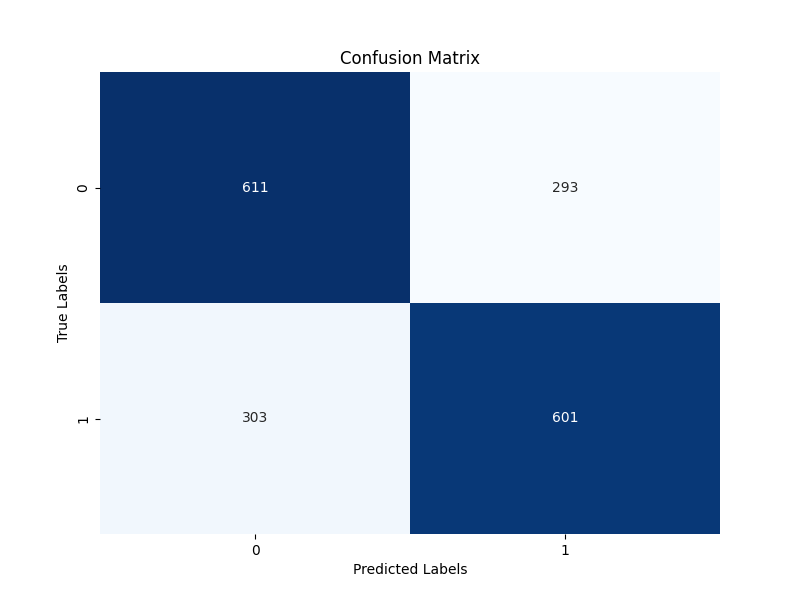}
        \caption{Confusion Matrix for the test set of CEDAR}
        \label{fig:cedar_c}
    \end{subfigure}
    \hfill
    \begin{subfigure}{0.49\textwidth}
        \centering
        \includegraphics[width=\linewidth]{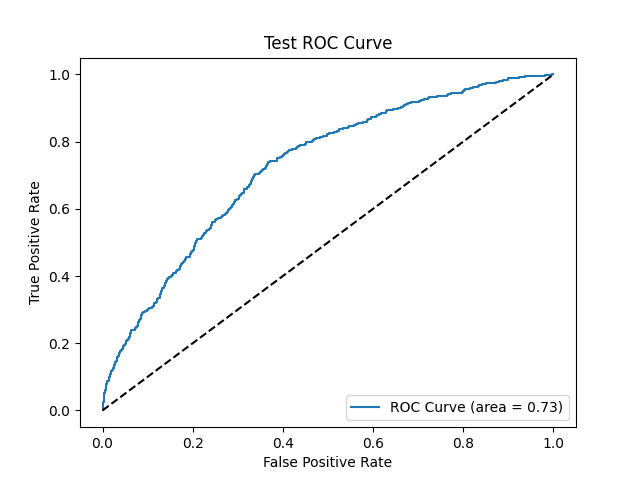}
        \caption{ROC Curve for the test set of CEDAR}
        \label{fig:cedar_roc}
    \end{subfigure}
    \caption{CI - Evaluation metrics for the CEDAR test set}
    \label{fig:cedar_results}
\end{figure}

\begin{figure}[H]
    \centering
    \begin{subfigure}{0.49\textwidth}
        \centering
        \includegraphics[width=\linewidth]{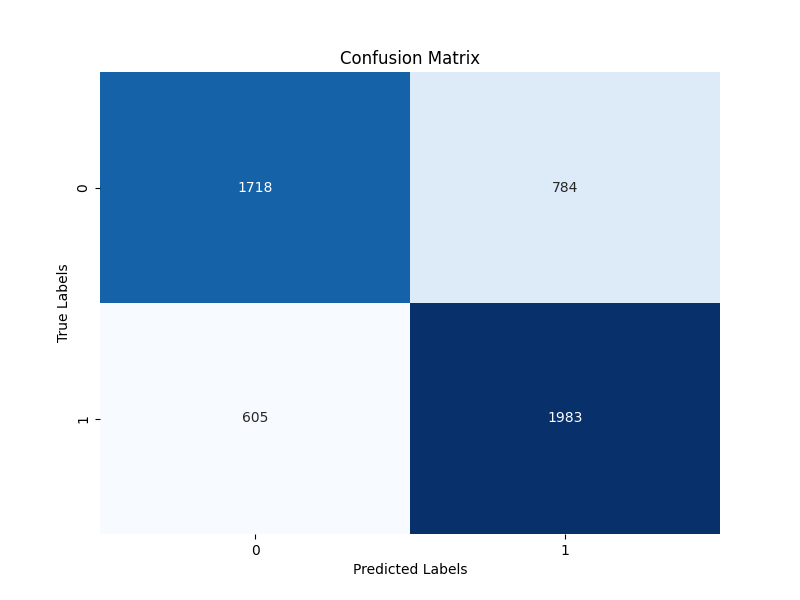}
        \caption{Confusion Matrix for the test set of ICDAR}
        \label{fig:icdar_c}
    \end{subfigure}
    \hfill
    \begin{subfigure}{0.49\textwidth}
        \centering
        \includegraphics[width=\linewidth]{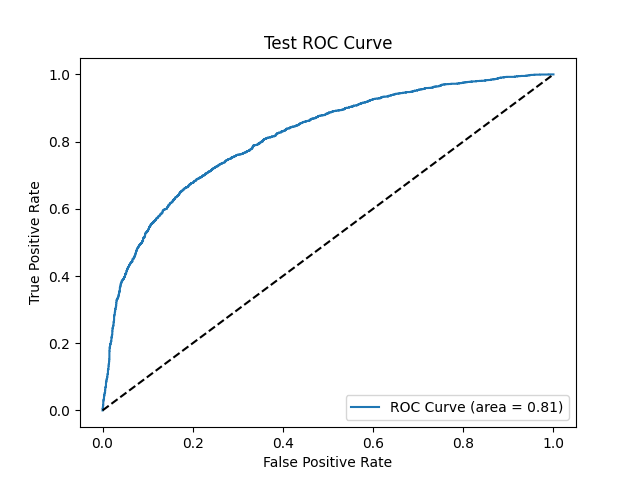}
        \caption{ROC Curve for the test set of ICDAR}
        \label{fig:icdar_roc}
    \end{subfigure}
    \caption{CI - Evaluation metrics for the ICDAR test set}
    \label{fig:icdar_results}
\end{figure}

\begin{figure}[H]
    \centering
    \begin{subfigure}{0.49\textwidth}
        \centering
        \includegraphics[width=\linewidth]{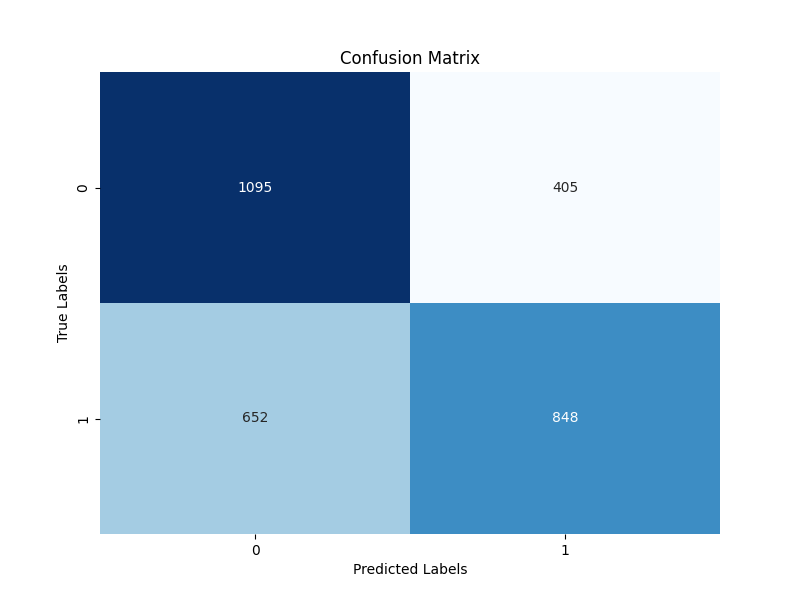}
        \caption{Confusion Matrix for the test set of GPDS}
        \label{fig:gpds_c}
    \end{subfigure}
    \hfill
    \begin{subfigure}{0.49\textwidth}
        \centering
        \includegraphics[width=\linewidth]{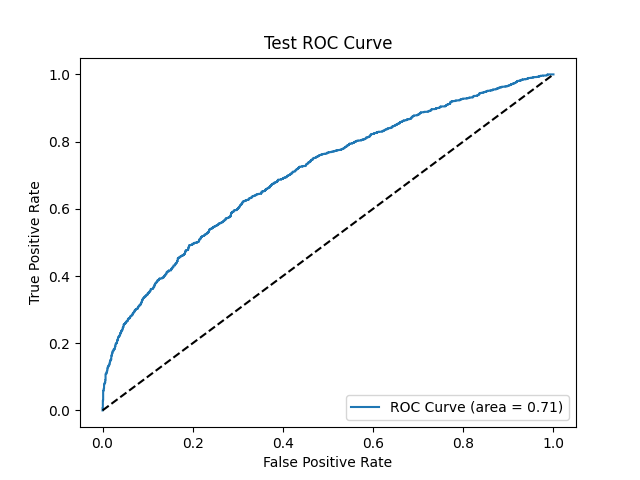}
        \caption{ROC Curve for the test set of GPDS}
        \label{fig:gpds_roc}
    \end{subfigure}
    \caption{CI - Evaluation metrics for the GPDS test set}
    \label{fig:gpds_results}
\end{figure}

\begin{figure}[H]
    \centering
    \begin{subfigure}{0.49\textwidth}
        \centering
        \includegraphics[width=\linewidth]{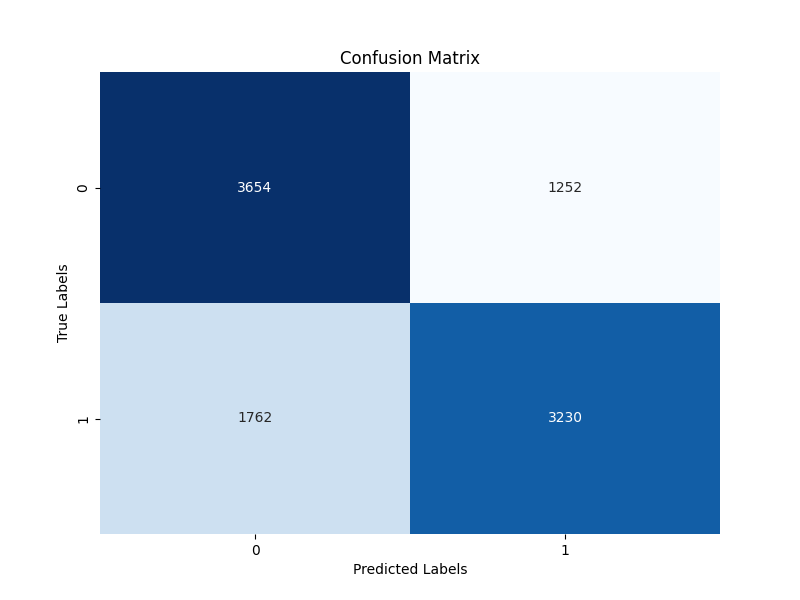}
        \caption{Confusion Matrix for the test set of MERGED}
        \label{fig:merged_c}
    \end{subfigure}
    \hfill
    \begin{subfigure}{0.49\textwidth}
        \centering
        \includegraphics[width=\linewidth]{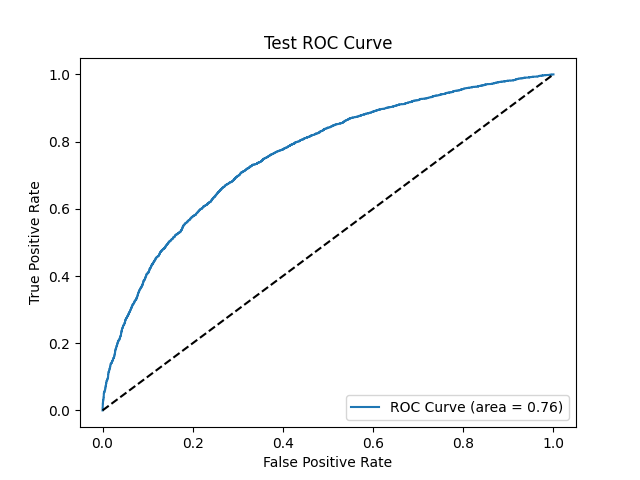}
        \caption{ROC Curve for the test set of MERGED}
        \label{fig:merged_roc}
    \end{subfigure}
    \caption{CI - Evaluation metrics for the MERGED test set}
    \label{fig:merged_results}
\end{figure}

\subsection{Triplet Loss Training}

\subsubsection{CEDAR + ICDAR Training}
\begin{figure}[H]
    \centering
    \begin{subfigure}{0.49\textwidth}
        \centering
        \includegraphics[width=\linewidth]{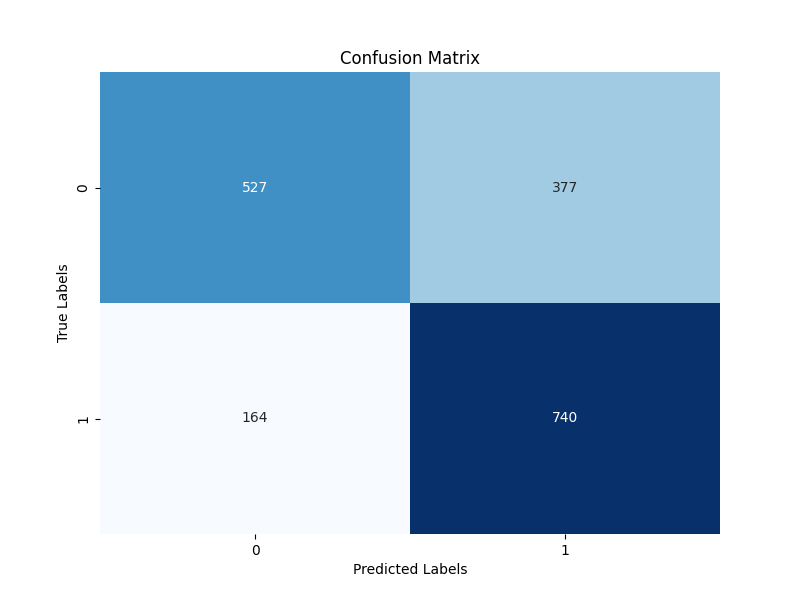}
        \caption{Confusion Matrix for the test set of CEDAR}
        \label{fig:cedar_c}
    \end{subfigure}
    \hfill
    \begin{subfigure}{0.49\textwidth}
        \centering
        \includegraphics[width=\linewidth]{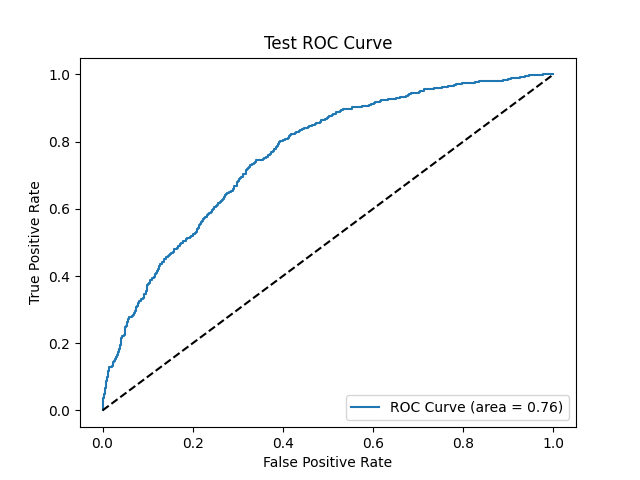}
        \caption{ROC Curve for the test set of CEDAR}
        \label{fig:cedar_roc}
    \end{subfigure}
    \caption{CI - Evaluation metrics for the CEDAR test set}
    \label{fig:cedar_results}
\end{figure}

\begin{figure}[H]
    \centering
    \begin{subfigure}{0.49\textwidth}
        \centering
        \includegraphics[width=\linewidth]{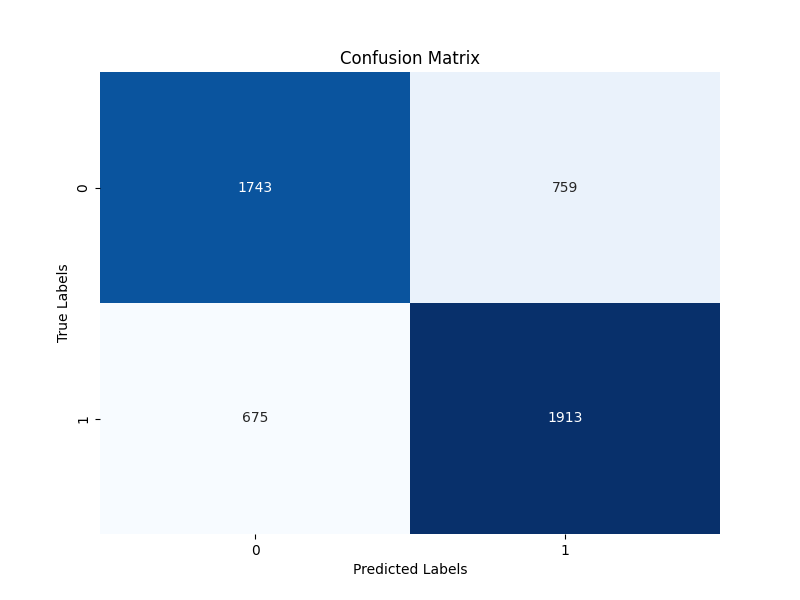}
        \caption{Confusion Matrix for the test set of ICDAR}
        \label{fig:icdar_c}
    \end{subfigure}
    \hfill
    \begin{subfigure}{0.49\textwidth}
        \centering
        \includegraphics[width=\linewidth]{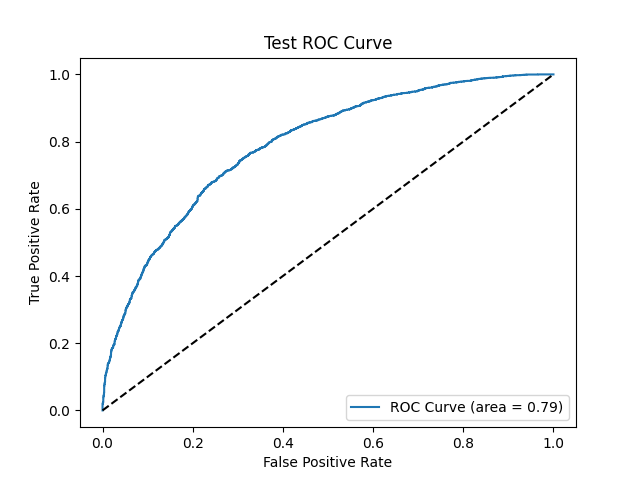}
        \caption{ROC Curve for the test set of ICDAR}
        \label{fig:icdar_roc}
    \end{subfigure}
    \caption{CI - Evaluation metrics for the ICDAR test set}
    \label{fig:icdar_results}
\end{figure}

\begin{figure}[H]
    \centering
    \begin{subfigure}{0.49\textwidth}
        \centering
        \includegraphics[width=\linewidth]{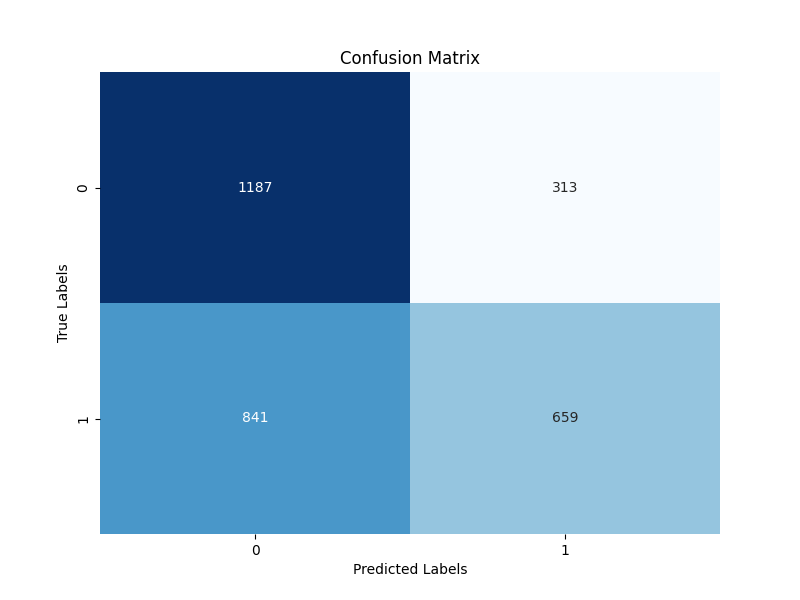}
        \caption{Confusion Matrix for the test set of GPDS}
        \label{fig:gpds_c}
    \end{subfigure}
    \hfill
    \begin{subfigure}{0.49\textwidth}
        \centering
        \includegraphics[width=\linewidth]{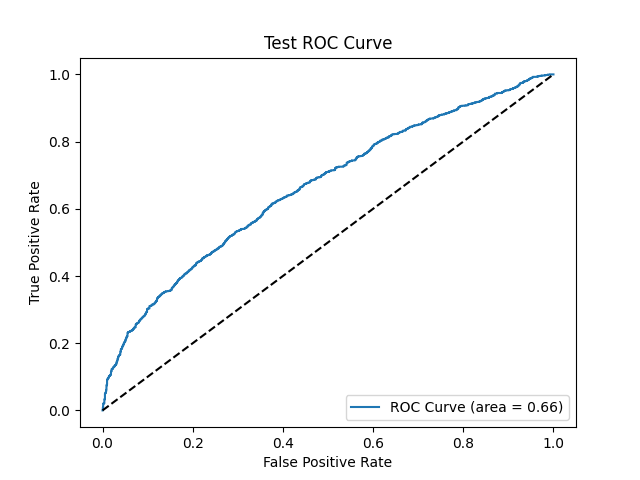}
        \caption{ROC Curve for the test set of GPDS}
        \label{fig:gpds_roc}
    \end{subfigure}
    \caption{CI - Evaluation metrics for the GPDS test set}
    \label{fig:gpds_results}
\end{figure}

\subsubsection{CEDAR + GPDS Training}
\begin{figure}[H]
    \centering
    \begin{subfigure}{0.49\textwidth}
        \centering
        \includegraphics[width=\linewidth]{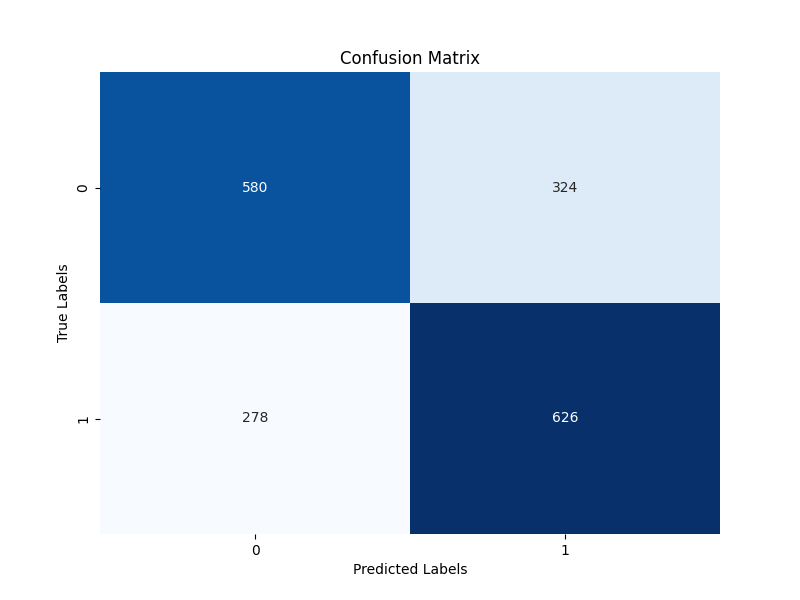}
        \caption{Confusion Matrix for the test set of CEDAR}
        \label{fig:cedar_c}
    \end{subfigure}
    \hfill
    \begin{subfigure}{0.49\textwidth}
        \centering
        \includegraphics[width=\linewidth]{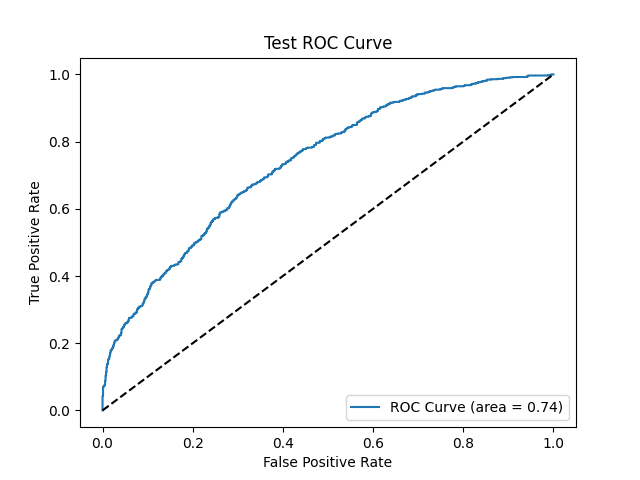}
        \caption{ROC Curve for the test set of CEDAR}
        \label{fig:cedar_roc}
    \end{subfigure}
    \caption{CG - Evaluation metrics for the CEDAR test set}
    \label{fig:cedar_results}
\end{figure}

\begin{figure}[H]
    \centering
    \begin{subfigure}{0.49\textwidth}
        \centering
        \includegraphics[width=\linewidth]{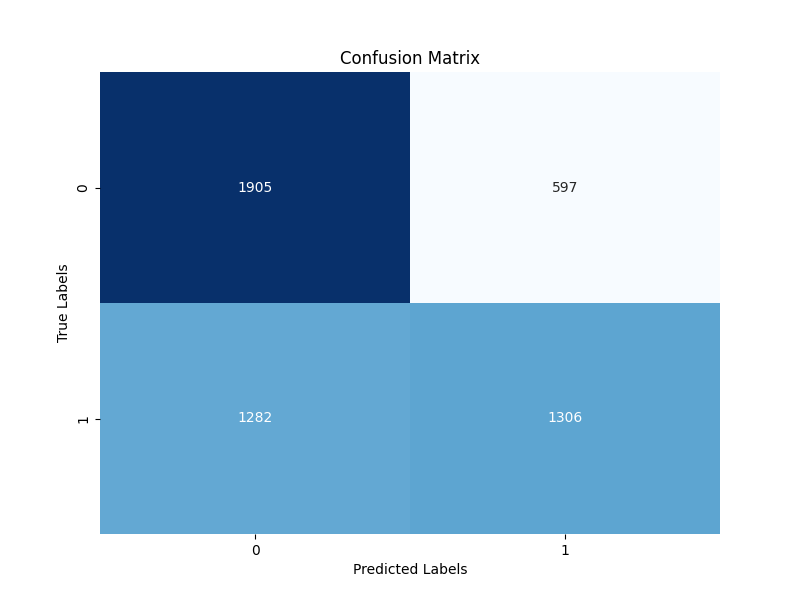}
        \caption{Confusion Matrix for the test set of ICDAR}
        \label{fig:icdar_c}
    \end{subfigure}
    \hfill
    \begin{subfigure}{0.49\textwidth}
        \centering
        \includegraphics[width=\linewidth]{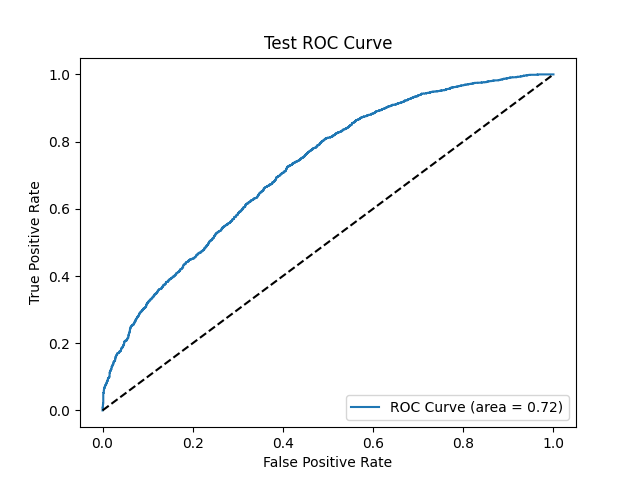}
        \caption{ROC Curve for the test set of ICDAR}
        \label{fig:icdar_roc}
    \end{subfigure}
    \caption{CG - Evaluation metrics for the ICDAR test set}
    \label{fig:icdar_results}
\end{figure}

\begin{figure}[H]
    \centering
    \begin{subfigure}{0.49\textwidth}
        \centering
        \includegraphics[width=\linewidth]{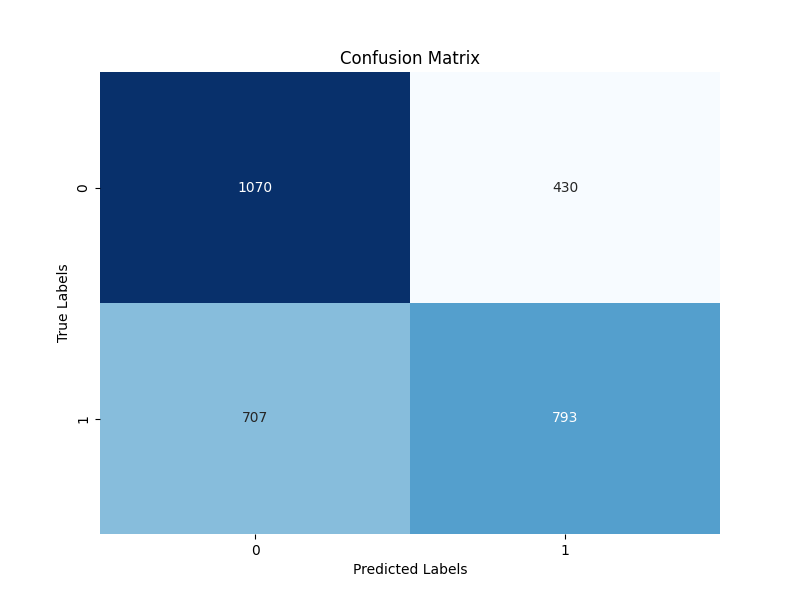}
        \caption{Confusion Matrix for the test set of GPDS}
        \label{fig:gpds_c}
    \end{subfigure}
    \hfill
    \begin{subfigure}{0.49\textwidth}
        \centering
        \includegraphics[width=\linewidth]{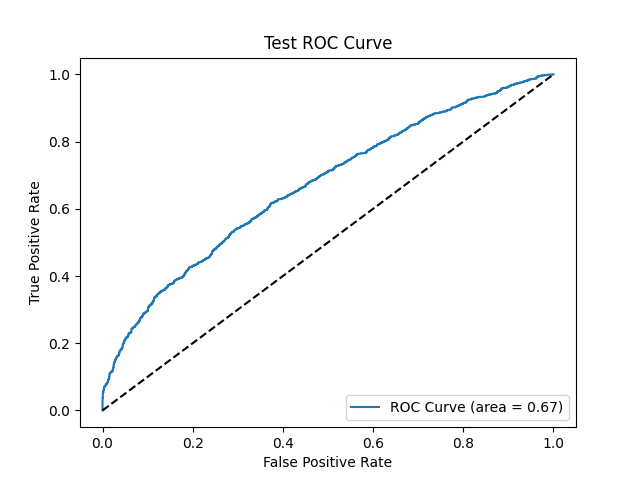}
        \caption{ROC Curve for the test set of GPDS}
        \label{fig:gpds_roc}
    \end{subfigure}
    \caption{CI - Evaluation metrics for the GPDS test set}
    \label{fig:gpds_results}
\end{figure}

\subsubsection{ICDAR + GPDS Training}
\begin{figure}[H]
    \centering
    \begin{subfigure}{0.49\textwidth}
        \centering
        \includegraphics[width=\linewidth]{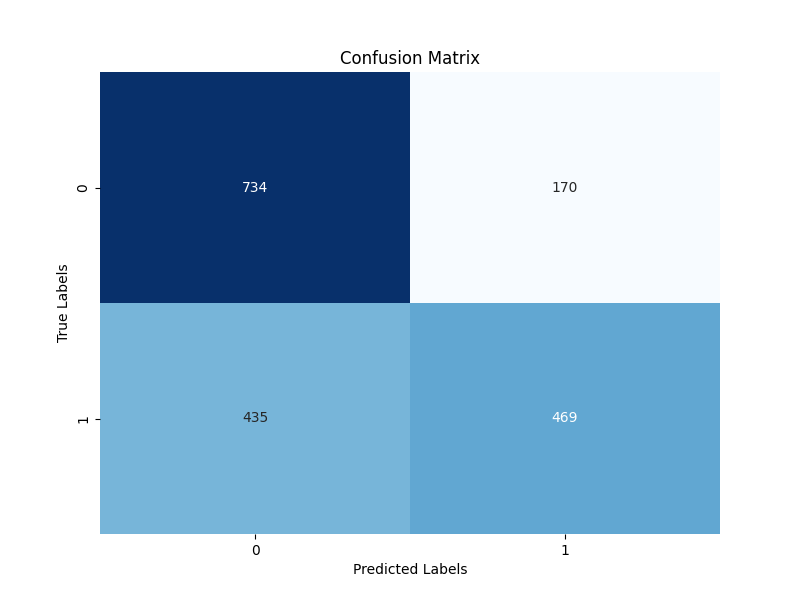}
        \caption{Confusion Matrix for the test set of CEDAR}
        \label{fig:cedar_c}
    \end{subfigure}
    \hfill
    \begin{subfigure}{0.49\textwidth}
        \centering
        \includegraphics[width=\linewidth]{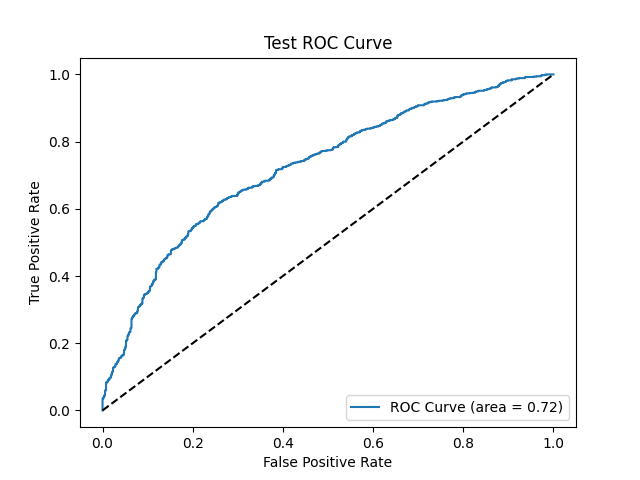}
        \caption{ROC Curve for the test set of CEDAR}
        \label{fig:cedar_roc}
    \end{subfigure}
    \caption{IG - Evaluation metrics for the CEDAR test set}
    \label{fig:cedar_results}
\end{figure}

\begin{figure}[H]
    \centering
    \begin{subfigure}{0.49\textwidth}
        \centering
        \includegraphics[width=\linewidth]{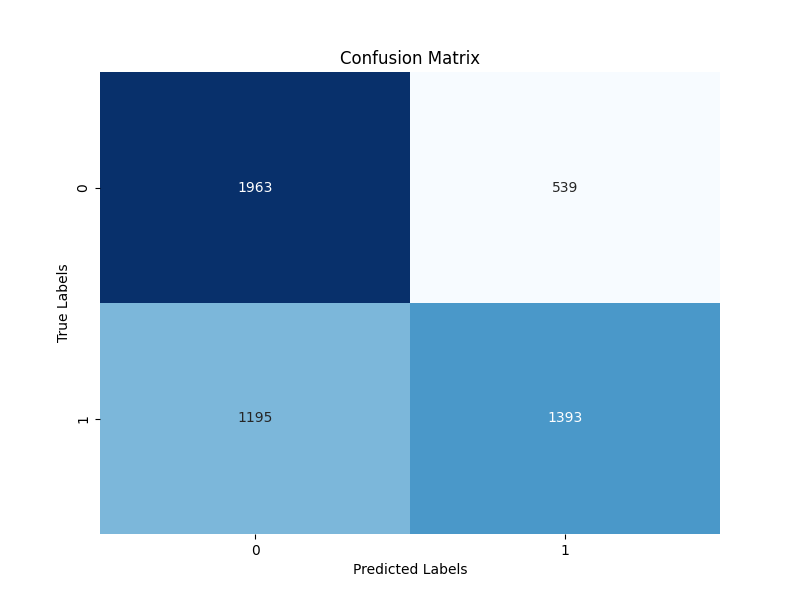}
        \caption{Confusion Matrix for the test set of ICDAR}
        \label{fig:icdar_c}
    \end{subfigure}
    \hfill
    \begin{subfigure}{0.49\textwidth}
        \centering
        \includegraphics[width=\linewidth]{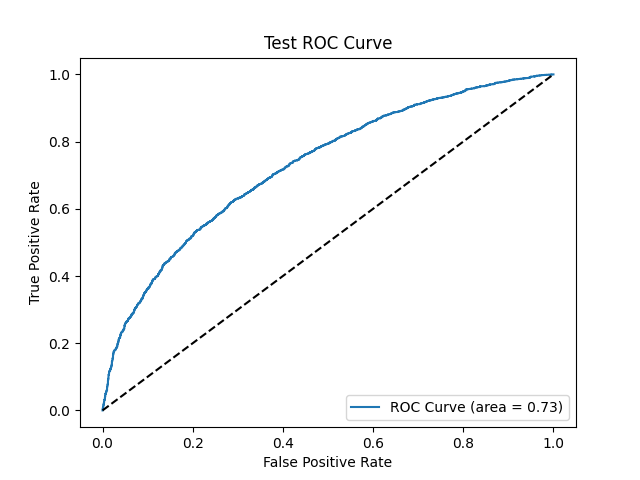}
        \caption{ROC Curve for the test set of ICDAR}
        \label{fig:icdar_roc}
    \end{subfigure}
    \caption{IG - Evaluation metrics for the ICDAR test set}
    \label{fig:icdar_results}
\end{figure}

\begin{figure}[H]
    \centering
    \begin{subfigure}{0.49\textwidth}
        \centering
        \includegraphics[width=\linewidth]{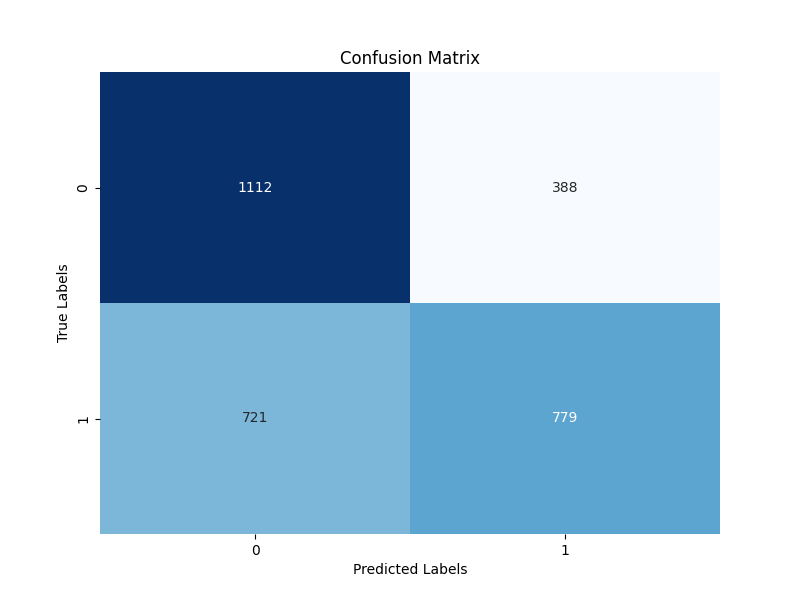}
        \caption{Confusion Matrix for the test set of GPDS}
        \label{fig:gpds_c}
    \end{subfigure}
    \hfill
    \begin{subfigure}{0.49\textwidth}
        \centering
        \includegraphics[width=\linewidth]{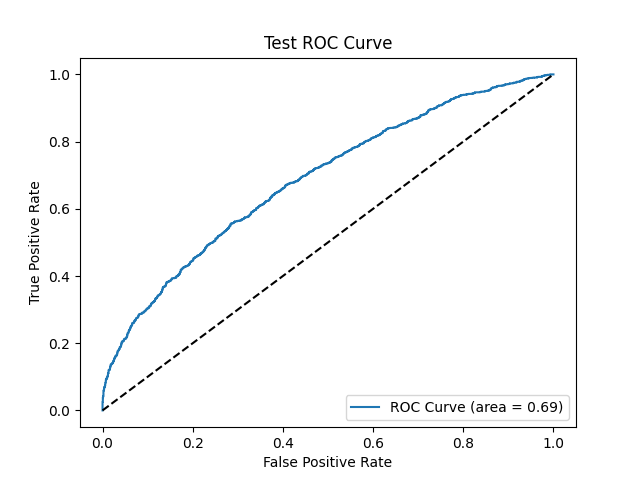}
        \caption{ROC Curve for the test set of GPDS}
        \label{fig:gpds_roc}
    \end{subfigure}
    \caption{IG - Evaluation metrics for the GPDS test set}
    \label{fig:gpds_results}
\end{figure}


\begin{landscape}
\begin{figure}[ht]
    \centering
    \includegraphics[width=\paperwidth, height=\paperheight, keepaspectratio]{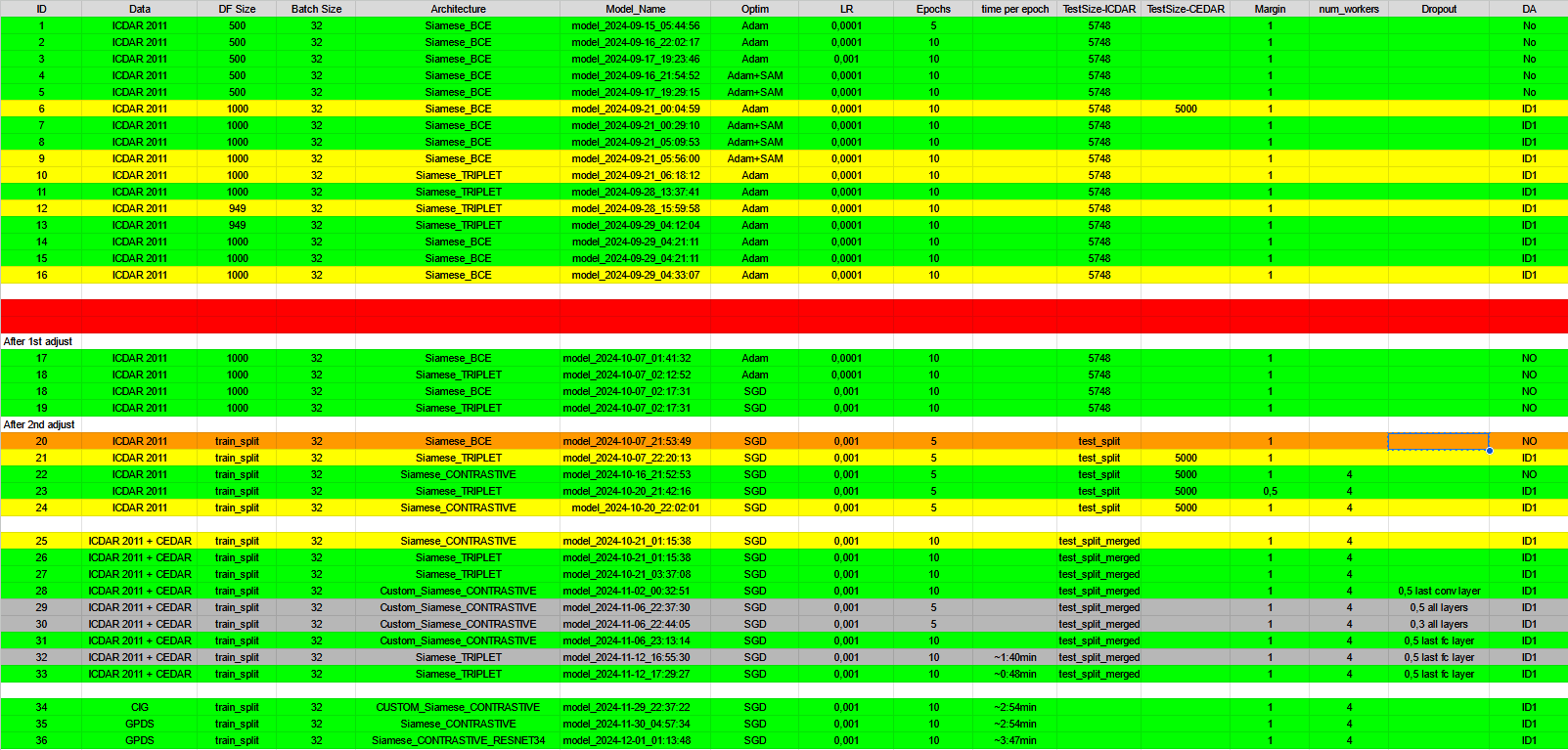} 
    \caption{Training log structure. Each row corresponds to an individual experiment, and each column indicates the respective hyperparameter configuration, specific observations, and data augmentation techniques applied.}
    \label{fig:training_log}
\end{figure}
\end{landscape}

\end{document}